\RequirePackage{fix-cm}
\documentclass[11pt]{article}

\usepackage[utf8]{inputenc}
\usepackage{xurl}
\usepackage{acl}
\usepackage{times}
\usepackage{latexsym}
\usepackage{algpseudocode}
\usepackage{algorithm}
\usepackage{amsmath}
\usepackage[T1]{fontenc}
\usepackage{titlesec}
\titlespacing*{\section}{0pt}{*1}{*1}
\usepackage[table]{xcolor} 
\usepackage{svg}
\definecolor{highlightgreen}{rgb}{0.85,1,0.85} 
\usepackage[utf8]{inputenc}
\usepackage{needspace}
\usepackage{microtype}

\usepackage{inconsolata}

\usepackage{graphicx}
\usepackage{amsmath}

\usepackage{caption}    
\title{Reason, Reward, Refine: Step-Level Errors Corrections with Structured \\ Feedback for Physics Reasoning in Small Language Models \vspace{1.5ex}}

\author{
  Raj Jaiswal$^{1*}$, 
  Dhruv Jain$^{2*}$, 
  Rishabh Dhawan$^{1*}$, 
  Sree Krishna Uppalapati$^{1*}$, \\[0.5ex]
  \textbf{Shin’ichi Satoh}$^4$, 
  \textbf{Tanuja Ganu}$^5$, 
  \textbf{Rajiv Ratn Shah}$^3$ \\[0.3ex]
  $^1$IIIT Delhi \quad $^2$IIT BHU  \quad $^3$IIT Kanpur \quad $^4$NII Tokyo \quad $^5$Microsoft Research India \\[0.5ex]
  \small \texttt{\{jaiswalp, rishabh23002, sree23533\}@iiitd.ac.in} \\
  \small \texttt{dhruv.jain.ece21@itbhu.ac.in, satoh@nii.ac.jp, rajivratn@iitk.ac.in, tanuja.ganu@microsoft.com} \\[0.3ex]
  \small $^*$Equal contribution
}

\usepackage{tcolorbox}
\tcbuselibrary{breakable, skins}

  \newtcolorbox{samplebox}{
    colback=white,
    colframe=black!70,
    boxrule=0.4pt,
    arc=2pt,
    left=8pt, right=8pt, top=6pt, bottom=6pt,
    fontupper=\ttfamily\footnotesize,
    breakable,
  }

  \newtcolorbox{promptbox}{
    colback=gray!5,
    colframe=black,
    boxrule=0.5pt,
    arc=2pt,
    left=8pt, right=8pt, top=6pt, bottom=6pt,
    fontupper=\ttfamily\footnotesize,
    breakable,
  }

\definecolor{headerModel}{RGB}{240,244,250}
\definecolor{headerGroup}{RGB}{225,232,245}
\definecolor{r1}{RGB}{255,252,252}   
\definecolor{r2}{RGB}{255,228,228}   
\definecolor{r3}{RGB}{255,195,195}   
\definecolor{r4}{RGB}{255,155,155}   
\definecolor{r5}{RGB}{235,80,80}     

\definecolor{compGreen}{RGB}{220,245,220}
\definecolor{compRed}{RGB}{255,220,220}

\usepackage{booktabs}
\usepackage{xcolor}
\usepackage{colortbl}
\usepackage{multirow}
\usepackage{array}
\definecolor{headerblue}{RGB}{31, 78, 121}
\definecolor{rowgray}{RGB}{242, 242, 242}

\usepackage{booktabs}
\usepackage{multirow}
\usepackage{makecell}
\usepackage{array}
\usepackage{colortbl}
\usepackage{xcolor}
\usepackage{graphicx}
\usepackage{float}
\usepackage{microtype}
\usepackage{inconsolata}
\usepackage{amsmath}

\usepackage{tcolorbox}
\tcbuselibrary{breakable, skins}

\usepackage{listings}
\definecolor{resBlue}{RGB}{31,78,121}
\definecolor{resPurple}{RGB}{88,44,131}
\definecolor{resRed}{RGB}{180,30,30}
\definecolor{resGreen}{RGB}{30,120,60}
\definecolor{resOrange}{RGB}{180,90,20}
\definecolor{resTeal}{RGB}{20,110,110}

\definecolor{pastelBlue}{RGB}{242,247,254}
\definecolor{pastelBlueDeep}{RGB}{235,243,252}
\definecolor{pastelBlueFrame}{RGB}{31,78,121}

\definecolor{pastelTeal}{RGB}{242,252,252}
\definecolor{pastelTealDeep}{RGB}{235,250,250}
\definecolor{pastelTealFrame}{RGB}{20,110,110}

\definecolor{pastelPurple}{RGB}{247,244,253}
\definecolor{pastelPurpleDeep}{RGB}{242,238,252}
\definecolor{pastelPurpleFrame}{RGB}{88,44,131}

\definecolor{pastelOrange}{RGB}{254,249,242}
\definecolor{pastelOrangeDeep}{RGB}{252,244,233}
\definecolor{pastelOrangeFrame}{RGB}{160,80,20}

\definecolor{pastelGreen}{RGB}{242,252,245}
\definecolor{pastelGreenFrame}{RGB}{30,120,60}
\definecolor{pastelRed}{RGB}{254,244,243}
\definecolor{pastelRedFrame}{RGB}{180,30,30}

\definecolor{heatLow}{RGB}{254,226,216}
\definecolor{heatMid}{RGB}{252,174,145}
\definecolor{heatHigh}{RGB}{251,106,74}

\definecolor{gg1}{RGB}{65,171,93}
\definecolor{gg2}{RGB}{116,196,118}
\definecolor{gg3}{RGB}{186,228,179}
\definecolor{gg4}{RGB}{229,245,224}
\definecolor{gg5}{RGB}{247,252,245}   

\definecolor{g1}{RGB}{108,198,120}
\definecolor{g2}{RGB}{162,220,170}
\definecolor{g3}{RGB}{198,235,202}
\definecolor{g4}{RGB}{226,245,228}

\newtcolorbox{sectionbox}[2]{
  enhanced,
  colback=#1,
  colframe=#2,
  colbacktitle=#2,
  coltitle=white,
  fonttitle=\normalfont\bfseries\normalsize,
  title={#1},
  boxrule=0.7pt, arc=4pt,
  left=8pt, right=8pt, top=5pt, bottom=5pt,
}

\newtcolorbox{contentbox}[3]{
  enhanced, breakable,
  colback=#1,
  colframe=#2,
  colbacktitle=#2,
  coltitle=white,
  fonttitle=\normalfont\bfseries\small,
  title={#3},
  boxrule=0.7pt, arc=4pt,
  left=8pt, right=8pt, top=6pt, bottom=6pt,
}

\newtcolorbox{analysisbox}[2]{
  enhanced,
  colback=#1,
  colframe=#2,
  boxrule=0.5pt, arc=3pt,
  left=6pt, right=6pt, top=4pt, bottom=4pt,
}

\newtcolorbox{configbox}[3]{
  enhanced, breakable,
  colback=#1,
  colframe=#2,
  colbacktitle=#2,
  coltitle=white,
  fonttitle=\normalfont\bfseries\small,
  title={#3},
  boxrule=0.6pt, arc=4pt,
  left=10pt, right=10pt, top=8pt, bottom=8pt,
}

\usepackage{etoolbox}
\AtBeginEnvironment{algorithm}{\small}

\usepackage{algorithm}
\usepackage{algpseudocode}
\usepackage{xcolor}
\usepackage{tikz}
\usetikzlibrary{backgrounds}

\definecolor{algBlue}{RGB}{235,244,255}
\definecolor{algYellow}{RGB}{255,251,230}
\definecolor{algGreen}{RGB}{235,248,238}
\definecolor{algComment}{RGB}{140,140,140}

\algrenewcomment[1]{%
  \hfill\textcolor{algComment}{\(\triangleright\) \textit{#1}}}

\usepackage{pifont}
\newcommand{\cmark}{\textcolor{resGreen}{\ding{51}}}
\newcommand{\xmark}{\textcolor{resRed}{\ding{55}}}

\begin{document}
\maketitle

\begin{abstract}
Physics reasoning fails structurally in small language models:
an error at any step propagates forward, corrupting every
inference that follows.\ Limited domain knowledge, hallucination
under multi-step derivation, and distributional sensitivity
compound this failure. We propose a step-level reward framework
that identifies the first reasoning error, generates targeted
structured feedback, and trains the model to revise its solution
via policy gradient with KL regularization, without exposing it
to ground truth solutions as generation targets. Unlike
annotation-dependent step-level methods, no preference data
construction is required and the external verifier operates
exclusively at training time.\ Across five physics
benchmarks, our framework delivers accuracy gains of 17--20\% over
CoT prompting and 10--16\% over the strongest baseline, reduces
calculation errors from 56.9\% to 23.5\%, and reduces
miscomprehension errors from 22.3\% to 12.0\% in the best observed
cases. Conceptual errors reduce from 89.7\% to 68.7\%, yet persist
as the hardest failure mode across all conditions.\\ \\
\textit{Code, prompts, and experimental details are provided in the
Appendix section (\S\ref{sec:appendixmain}}).

\end{abstract}


\section{Introduction}

Physics reasoning is inherently sequential: each solution step 
depends on the correctness of all preceding steps, and an error 
at any point invalidates every inference that 
follows~\cite{jaiswal2024improvingphysicsreasoninglarge, 
anand2024enhancingllmsphysicsproblemsolving, ding2023using}. Language models have made substantial progress on reasoning 
tasks~\cite{wei2022chain, fu2022complexity, zhang2022automatic}, 
yet physics reasoning remains an open challenge for small language 
models (SLMs)~\cite{srivastava2025reasoningabilitysmalllanguage, 
boye2025large}. Larger models benefit from knowledge compression 
across broad training distributions; smaller models do not, and 
the reasoning failures that emerge are structurally distinct from 
those observed at scale~\cite{zhang2024small, 
srivastava2025reasoningabilitysmalllanguage}.

\begin{table}[t]
\centering
\caption{\textbf{Our Framework Achieves the Highest Average Accuracy
Across All Conditions.} Per-model average (\%) across five benchmarks;
Cell shading indicates per-row rank across conditions. Discussed in \S\ref{sec:results}.}
\fontsize{8.5pt}{9pt}\selectfont
\setlength{\tabcolsep}{5pt}
\renewcommand{\arraystretch}{1.3}
\begin{tabular}{l | ccccc}
\toprule
\textbf{Model} &
\textbf{CoT} & \textbf{RAG} & \textbf{SFT} &
\textbf{DPO} & \textbf{Ours} \\
\midrule
Qwen 2.5 1.5B
& \cellcolor{gg3}46.1
& \cellcolor{gg2}53.3
& \cellcolor{gg5}41.0
& \cellcolor{gg5}40.7
& \cellcolor{gg1}58.5 \\
LLaMA 3.2 1B
& \cellcolor{gg4}32.5
& \cellcolor{gg3}41.1
& \cellcolor{gg5}37.5
& \cellcolor{gg5}37.5
& \cellcolor{gg1}53.9 \\
LLaMA 3.2 3B
& \cellcolor{gg3}44.8
& \cellcolor{gg2}51.2
& \cellcolor{gg4}44.0
& \cellcolor{gg4}44.1
& \cellcolor{gg1}64.4 \\
Phi 3.5 Mini 3.8B
& \cellcolor{gg3}50.5
& \cellcolor{gg2}54.3
& \cellcolor{gg4}51.0
& \cellcolor{gg3}52.1
& \cellcolor{gg1}61.5 \\
\bottomrule
\end{tabular}
\label{tab:avg_results}
\end{table}

Prior work has progressively improved physics reasoning in 
language models, with each approach addressing a distinct 
limitation~\cite{ding2023using, pang2024physics,anand2024enhancingllmsphysicsproblemsolving}. Chain-of-Thought prompting~\cite{wei2022chain} improves reasoning 
transparency by externalizing intermediate steps, though 
step-level error propagation across dependent steps remains 
unaddressed. Retrieval-Augmented 
Generation~\cite{lewis2020retrieval} provides access to relevant 
domain knowledge at inference time, though correct retrieval does 
not guarantee correct application of retrieved 
content~\cite{anand2024enhancingllmsphysicsproblemsolving}. Supervised fine-tuning on 
expert reasoning trajectories~\cite{ho2022large, luo2023wizardmath} 
improves structural response quality, though models trained under 
token prediction objectives have been shown to reproduce surface 
reasoning patterns without internalizing the underlying 
logic~\cite{lobo2024impact, mccloskey1989catastrophic}. Direct 
Preference Optimization~\cite{rafailov2023direct} introduces 
comparative signal between correct and incorrect solution pairs, 
though preference is computed over complete responses and does not 
identify which reasoning step produced the incorrect 
outcome~\cite{lai2024stepdpo, chen2024svpo}.

\begin{table*}[ht]
\centering
\caption{\textbf{Comparison with Prior Work.}
Prior work covers at most three of the five dimensions;
this work is the first to address all five jointly}
\label{tab:related_work}
\fontsize{8pt}{10pt}\selectfont
\setlength{\tabcolsep}{10pt}
\renewcommand{\arraystretch}{1.4}
\begin{tabular}{l | ccccc}
\toprule

\textbf{Work} &
\textbf{\shortstack{Physics\\Domain}} &
\textbf{\shortstack{Step-Level\\Localization}} &
\textbf{\shortstack{Structured\\Feedback}} &
\textbf{\shortstack{No Preference\\Data}} &
\textbf{\shortstack{Position\\Reward}} \\

\midrule

Step-DPO \cite{lai2024stepdpo}
& \xmark & \cmark & \xmark & \xmark & \xmark \\

Full-Step-DPO \cite{xu2025full}
& \xmark & \cmark & \xmark & \cmark & \cmark \\

SVPO \cite{chen2024svpo}
& \xmark & \cmark & \xmark & \cmark & \cmark \\

Step-KTO \cite{lin2025stepktooptimizingmathematicalreasoning}
& \xmark & \cmark & \xmark & \cmark & \xmark \\

Self-Refine \cite{madaan2024self}
& \xmark & \cmark & \cmark & \cmark & \xmark \\

\midrule

\textbf{This Work}
& \cmark & \cmark & \cmark & \cmark & \cmark \\

\bottomrule
\end{tabular}
\end{table*}

Step-level preference optimization~\cite{xu2025full, lai2024stepdpo} 
assigns rewards at individual reasoning steps, but depends on 
externally annotated step-level data and a reliable verifier. 
In scientific domains, both are prohibitively 
costly, and small language models have 
been shown to be unreliable self-verifiers~\cite{li2024evaluating, 
zhang2024small}.

Our contributions are as follows:
\begin{itemize}

\item \textbf{Step-Level Reward Mechanism:}
We propose $r = e_{\text{first}} / (n+1)$, penalizing
earlier reasoning failures more heavily and enabling
correction at the precise point of failure rather than
over the complete response, without requiring annotated
step-level preference data.
(\S\ref{sec:reward})

\item \textbf{Error-Type Reduction via Structured Feedback:}
Error-type-conditioned feedback at training time reduces
Calculation errors from 56.9\% to 23.5\%, Problem
Miscomprehension from 22.3\% to 12.0\%, and Conceptual
Misapplication from 89.7\% to 68.7\% in the best
observed case per error type; Conceptual Misapplication
persists across all conditions.
(\S\ref{sec:feedback_gen}, Table~\ref{tab:component})

\item \textbf{Empirical Gains Across Models and Benchmarks:}
Our framework delivers accuracy gains of 17--20\% over
CoT prompting and 10--16\% over all evaluated baselines
across four open-source language models and five physics
benchmarks, with a peak gain of 27.1\% on JEEBench for
LLaMA 3.2 3B.
(Table~\ref{tab:main-results-part1},
Table~\ref{tab:main-results-part2})

\end{itemize}

\section{Related Work}
\label{sec:relatedwork}
Reinforcement learning from human feedback~\cite{ouyang2022training, 
christiano2017deep} introduced reward-based policy optimization 
as an alternative to imitation, training models against a scalar 
reward signal rather than fixed output targets. Direct Preference 
Optimization~\cite{rafailov2023direct} reformulates this as a 
closed-form preference objective over chosen and rejected response 
pairs, removing the need for an explicit reward model. Both 
frameworks treat the complete response as a single unit; the 
reward or preference signal is computed over the output as a 
whole and does not identify which reasoning step produced an 
incorrect outcome~\cite{lai2024step}.

Step-DPO~\cite{lai2024step} and Full-Step-DPO~\cite{xu2025full} 
extend preference optimization to individual reasoning steps, 
assigning step-wise rewards across the reasoning chain. 
Step-level supervision has been shown to improve over 
response-level signal on mathematical reasoning 
benchmarks~\cite{xu2025full}, though both methods require 
separately annotated step-level preference data and an external 
verifier to evaluate intermediate steps~\cite{xu2025full, 
cobbe2021training}.

Self-Refine~\cite{madaan2024self} and 
Reflexion~\cite{shinn2024reflexion} implement iterative 
refinement through self-generated feedback, while 
SCoRe~\cite{kumar2024training} trains models to self-correct 
via reinforcement learning. Small language models have been 
shown to be unreliable verifiers of their own reasoning in 
sub-4B parameter regimes~\cite{zhang2024small, 
tyen-etal-2024-llms, li2024evaluating, huang2023large}, 
making self-correction an insufficient mechanism at this scale.

\begin{figure*}[!t]
    \centering
    \includegraphics[width=0.90\textwidth]{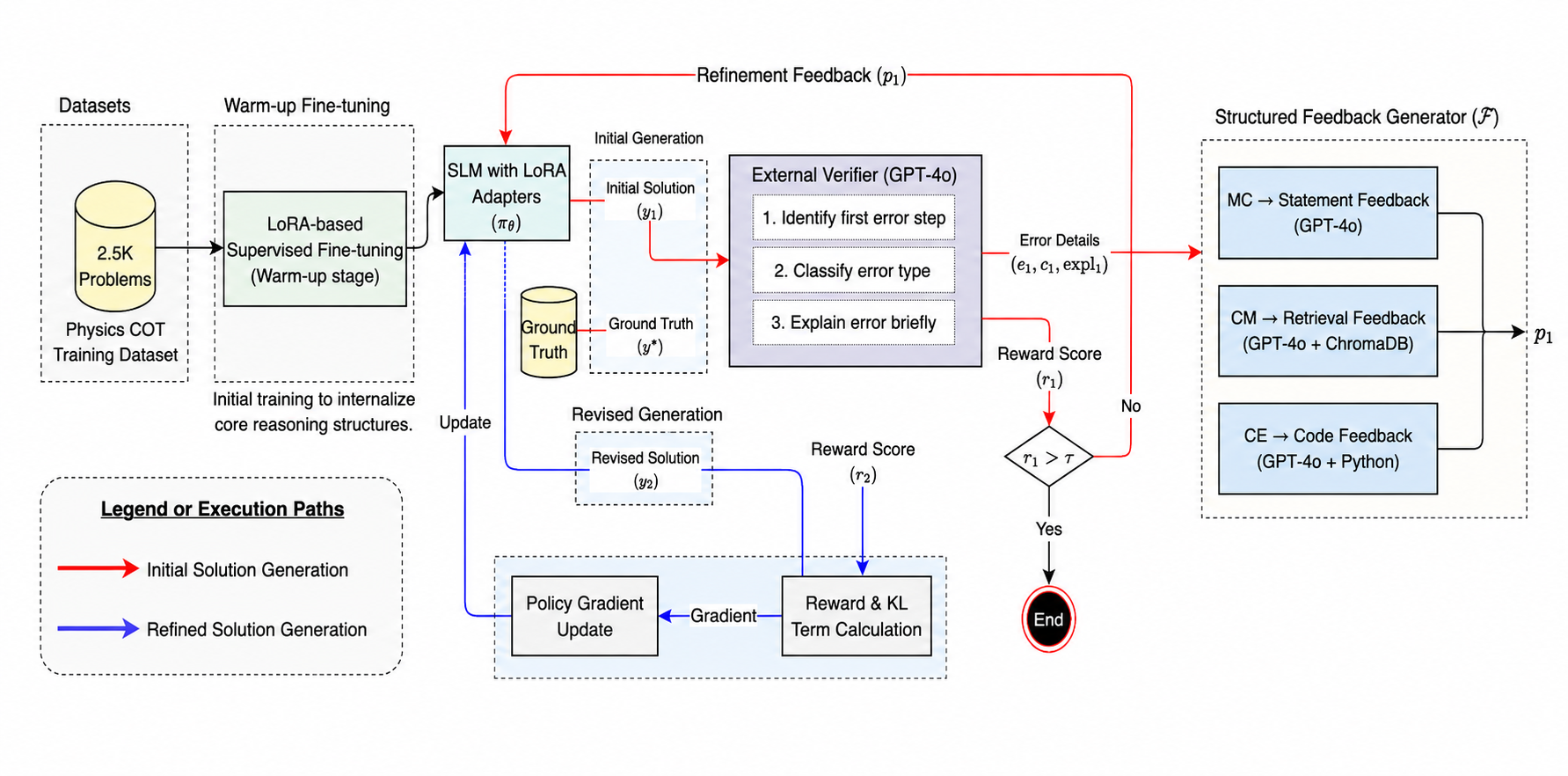}
    \caption{\textbf{Step-Level Reward Penalizes Earlier Failures  More Heavily; Feedback Targets the Precise Point of Failure.} $\pi_\theta$ generates $y_1$, receives error-type-conditioned feedback $p_1$ from GPT-4o, and revises to $y_2$; reward  $r = e_{\text{first}}/(n+1)$ drives policy gradient update. Discussed \S\ref{sec:methodology}.}
    \label{fig:flow_diagram}
\end{figure*}

\section{Methodology}
\label{sec:methodology}

We propose a step-level reward training framework for physics
reasoning in small language models, illustrated in
Figure~\ref{fig:flow_diagram}. The framework operates in three
stages. Stage 1 (\S\ref{sec:warmup}) initializes the policy
$\pi_\theta$ through supervised fine-tuning on numbered
step-by-step chain-of-thought solutions. Stage 2
(\S\ref{sec:reward}) enters an iterative training loop: the
model generates an initial solution $y_1$, an external verifier
(GPT-4o) identifies the first reasoning error against ground
truth $y^*$ and computes a step-level reward $r =
e_{\text{first}} / (n+1)$. Stage 3 (\S\ref{sec:feedback_gen})
generates error-conditioned feedback $p_1$; the model produces
a revised solution $y_2$ from which the policy is updated via
policy gradient with KL regularization
(\S\ref{sec:policy_update}). The model is never exposed to
correct solutions as generation targets — it receives only
structured feedback identifying the first error step and its
type.
\subsection{Stage 1: Supervised Fine-Tuning Warm-up}
\label{sec:warmup}

The fine-tuning corpus comprises 2,494 physics problems sourced
from JEE preparation
materials~\cite{Pandey:2020,Tipler:1999,Pinsky:1989,Halliday:2014},
distinct from all evaluation benchmarks. Each problem is paired
with a numbered chain-of-thought solution across five domains:
Mechanics, Electromagnetism, Thermodynamics, Waves \& Optics,
and Modern Physics.

\begin{figure*}[t]
\centering
\includegraphics[width=1.0\textwidth]{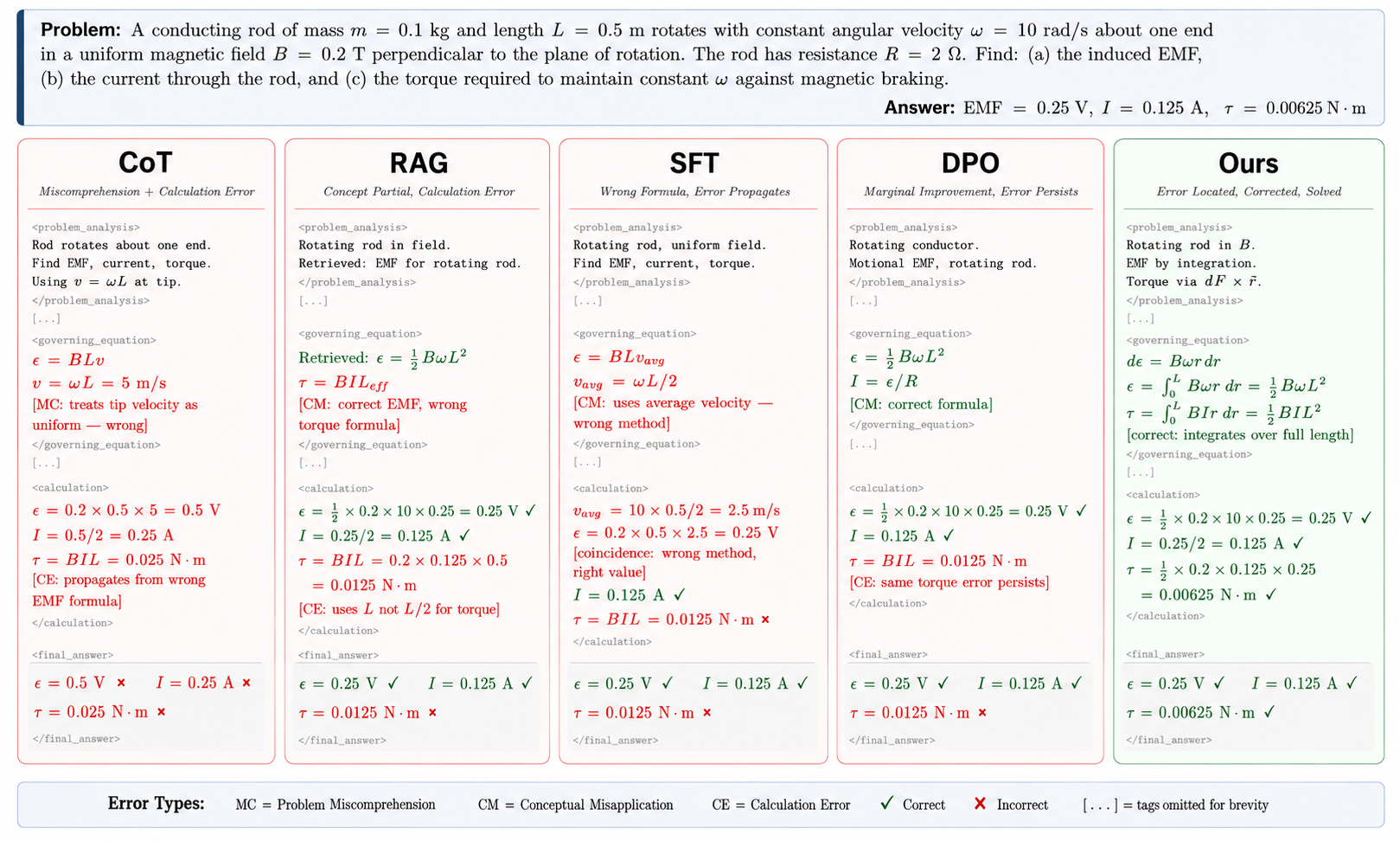}
\caption{\textbf{Targeted Feedback at the First Error Step Enables the Model to Reason Toward a Correct Solution.} Each column shows one Baseline responses on the same problem;
red annotations mark the first error step and type, green checkmarks in \texttt{<final\_answer>} indicate correct quantities. Structured feedback on the identified error step progressively corrects reasoning across revision passes. Discussed in \S\ref{sec:errortypereduciton} \& \S\ref{sec:stpeslevelworksandfail}}
\label{fig:qualitative}
\label{fig:qualitative2}
\end{figure*}

The policy $\pi_\theta$ is initialized via LoRA
adapters~\cite{hu2022lora} on this corpus. This stage
internalizes the step-indexed format required for the verifier
to reliably identify individual reasoning steps; without it,
reward computation lacks the structural consistency it depends
on. Upon completion, the adapter weights are frozen as the
reference policy $\pi_{\text{ref}}$, serving as the KL
regularization anchor throughout Stage 2, and $\pi_\theta$ is
reinitialized from these same weights as the starting point
for reward-guided training. Hyperparameter details are in
Appendix~\ref{sec:appendix_config}.

\subsection{Stage 2: Step-Level Reward Mechanism}
\label{sec:reward}

Physics reasoning failures exhibit three recurring
patterns~\cite{jaiswal2024improvingphysicsreasoninglarge}:
Problem Miscomprehension (MC), misidentified
objective or misread quantities; Conceptual
Misapplication (CM), wrong governing law or principle
applied outside validity conditions; and Calculation
Error (CE), correct setup but arithmetic or algebraic
error in execution. Each error type is interdependent; classification follows
MC, CM, CE priority order to route feedback accordingly
(\S\ref{sec:feedback_gen}).

\paragraph{Verifier.}
Let $x$ denote a physics problem and $y^*$ its ground truth
solution. An external verifier receives a model-generated
solution $y_1$ and identifies the first step at which reasoning
deviates from $y^*$, returning a structured error tuple
$(e_1, c_1, \text{expl}_1)$: the index of the first error step
$e_1$, the error category $c_1 \in \{\text{MC}, \text{CM},
\text{CE}\}$, and a natural language explanation
$\text{expl}_1$. The verifier is GPT-4o, queried via the
OpenAI API. The verifier operates exclusively at training
time; no external model is queried at inference.

\paragraph{Reward.}
The step-level reward is defined as:
\begin{equation}
    r = \frac{e_{\text{first}}}{n + 1}
    \label{eq:reward}
\end{equation}
where $e_{\text{first}}$ is the step index of the first
reasoning error and $n$ is the total number of reasoning steps.
This assigns higher reward to solutions that sustain correct
reasoning further into the chain, penalizing earlier failures
more heavily regardless of error type. If the verifier returns
no error ($e_1 = 0$), reward is set to 1.0 and the sample is
skipped in the training loop.

\subsection{Stage 3: Feedback Generation}
\label{sec:feedback_gen}

The feedback generator $\mathcal{F}$ receives the error tuple
$(e_1, c_1, \text{expl}_1)$ and produces structured feedback
$p_1$ conditioned on error type $c_1$ via one of three
channels, all implemented using GPT-4o; feedback is generated
only when $r_1 \leq \tau$, skipping solutions that already
meet the reward threshold. For Problem Miscomprehension
errors, a structured prompt instructs the model to re-read
the problem statement and correct its understanding of the
given conditions before regenerating. For Conceptual
Misapplication errors, a retrieval query is issued against a
vector store built from a physics formula
corpus~\cite{neet2023} using sentence-transformer
embeddings~\cite{reimers2019sentencebert} to supply
grounded governing principles and formulas rather than
relying on GPT-4o generation alone; the retrieved content
is combined with the error explanation into feedback
identifying the correct physical law. For Calculation errors,
Python code is generated to correctly perform the computation
at the identified error step; the execution output is
incorporated into feedback clarifying the correct calculation
and the original arithmetic error.

\subsection{Policy Update}
\label{sec:policy_update}

The policy $\pi_\theta$ generates a revised solution
$y_2 \sim \pi_\theta(\cdot \mid x, y_1, p_1)$, conditioned
on the original problem $x$, the first attempt $y_1$, and
structured feedback $p_1$. The revised solution is evaluated
by the verifier to produce a second error tuple
$(e_2, c_2, \text{expl}_2)$ and reward $r_2 = e_2 / (n+1)$,
where $e_2$ is the first error step index in $y_2$ evaluated
against the ground truth solution $y^*$; if the verifier
returns no error, $r_2 = 1.0$.

The policy is updated via a policy gradient objective with
KL regularization against the frozen reference policy
$\pi_{\text{ref}}$ — the Stage 1 SFT checkpoint:

\begingroup
\small
\begin{equation}
\begin{split}
\mathcal{L}(\theta) = -\Bigl[
    &\bigl(\log \pi_\theta(y_2 \mid x, y_1, p_1) -
    \log \pi_\theta(y_1 \mid x)\bigr) \cdot r_2 \\
    &- \beta \, D_{\mathrm{KL}}\bigl(
        \pi_{\theta}(\cdot \mid x) \,\|\,
        \pi_{\text{ref}}(\cdot \mid x)
    \bigr)
\Bigr]
\end{split}
\label{eq:objective}
\end{equation}
\endgroup

\noindent where $\beta = 0.1$ controls KL regularization
strength and $\tau = 0.9$ determines early stopping;
full sensitivity analysis is in
Appendix~\ref{sec:appendix_config}.
Gradients are computed with respect to $y_2$ only; the
log-probability of $y_1$ serves as a fixed baseline and is
detached from the computation graph. The objective
incentivizes improvement from $y_1$ to $y_2$ by weighting
the log-probability gain by the step-level reward $r_2$.
The KL term prevents the policy from deviating excessively
from the SFT initialization. The KL term is computed on
$\pi_\theta(\cdot \mid x)$ rather than on the full
conditioning context $(x, y_1, p_1)$, anchoring the policy
to the base reasoning distribution established during
warm-up independent of feedback quality. Full hyperparameter
and code details are provided in
Appendix~\ref{sec:appendix_config}.

\begin{algorithm}[t]
\fontsize{8.5pt}{10.5pt}\selectfont
\caption{: Step-Level Reward Training for Physics Reasoning in Small LMs}
\label{alg:training}
\begin{algorithmic}[1]
\Require Training dataset $\mathcal{D}$, policy $\pi_{\theta}$
         (warm-started from Stage 1 SFT), reference policy
         $\pi_{\text{ref}}$ (Stage 1 checkpoint, frozen),
         external verifier $\mathcal{V}$ (GPT-4o),
         feedback generator $\mathcal{F}$, threshold $\tau = 0.9$
\For{\textbf{each} training step}
    \State Sample $(x, y^{*}) \sim \mathcal{D}$
    \State Generate $y_1 \sim \pi_\theta(\cdot \mid x)$
    \State $(e_1, c_1, \text{expl}_1) \leftarrow
           \mathcal{V}(y_1, y^{*})$
           \Comment{Verify against ground truth}
    \State $r_1 \leftarrow e_1\,/\,(n + 1)$
           \Comment{Step-level reward}
    \If{$r_1 > \tau$}
        \State \textbf{continue}
        \Comment{Initial solution correct; skip}
    \EndIf
    \State $p_1 \leftarrow \mathcal{F}(x, y_1, e_1, c_1,
           \text{expl}_1)$
           \Comment{Structured feedback}
    \State Generate $y_2 \sim \pi_\theta(\cdot \mid x, y_1, p_1)$
           \Comment{Revised solution}
    \State $(e_2, c_2, \text{expl}_2) \leftarrow
           \mathcal{V}(y_2, y^{*})$
    \State $r_2 \leftarrow e_2\,/\,(n + 1)$
    \State Update $\theta$ via Eq.~\ref{eq:objective}
           \Comment{Policy gradient + KL}
\EndFor
\end{algorithmic}
\end{algorithm}


\begin{table*}[ht]
\centering
\caption{\textbf{Our Framework Leads on Easy Benchmarks;
Knowledge Grounding via RAG Outperforms Fine-Tuning Approaches.}
Final answer accuracy (\%) confirms RAG improves over CoT across all models while SFT and DPO fail to consistently improve over CoT. Cell shading indicates per-row rank across conditions. Discussed in \S\ref{sec:baslineperformance}.}
\fontsize{7.5pt}{9pt}\selectfont
\setlength{\tabcolsep}{4pt}
\renewcommand{\arraystretch}{1.3}
\begin{tabular}{l | c c c c c | c c c c c | c c c c c }
\toprule
\textbf{Model} &
\multicolumn{5}{c|}{\textbf{SciEval-Static}} &
\multicolumn{5}{c|}{\textbf{MMLU-High}} &
\multicolumn{5}{c}{\textbf{MMLU-College}} \\
\cmidrule(lr){2-6} \cmidrule(lr){7-11} \cmidrule(lr){12-16}
& CoT & RAG & FT & DPO & Ours
& CoT & RAG & FT & DPO & Ours
& CoT & RAG & FT & DPO & Ours \\
\midrule
Qwen 2.5 1.5B
 & \cellcolor{g3}62.73 & \cellcolor{g2}68.12 & \cellcolor{g4}57.93 & \cellcolor{g4}59.32 & \cellcolor{g1}79.29
 & \cellcolor{g3}47.05 & \cellcolor{g1}59.41 & \cellcolor{g4}41.18 & \cellcolor{g4}46.47 & \cellcolor{g2}55.03
 & \cellcolor{g3}52.62 & \cellcolor{g1}62.27 & \cellcolor{g4}43.22 & \cellcolor{g4}43.22 & \cellcolor{g2}53.64 \\
LLaMA 3.2 1B
 & \cellcolor{g4}44.51 & \cellcolor{g2}57.37 & \cellcolor{g4}55.49 & \cellcolor{g3}56.93 & \cellcolor{g1}68.09
 & \cellcolor{g4}30.58 & \cellcolor{g4}38.02 & \cellcolor{g2}42.35 & \cellcolor{g3}39.41 & \cellcolor{g1}55.29
 & \cellcolor{g4}33.63 & \cellcolor{g2}45.82 & \cellcolor{g3}38.14 & \cellcolor{g4}36.44 & \cellcolor{g1}57.89 \\
LLaMA 3.2 3B
 & \cellcolor{g3}62.26 & \cellcolor{g2}66.87 & \cellcolor{g4}61.59 & \cellcolor{g4}57.93 & \cellcolor{g1}81.52
 & \cellcolor{g3}50.00 & \cellcolor{g2}55.29 & \cellcolor{g4}48.29 & \cellcolor{g4}47.65 & \cellcolor{g1}67.20
 & \cellcolor{g3}53.63 & \cellcolor{g2}61.82 & \cellcolor{g4}45.76 & \cellcolor{g4}50.85 & \cellcolor{g1}69.09 \\
Phi 3.5 Mini 3.8B
 & \cellcolor{g2}62.43 & \cellcolor{g3}61.01 & \cellcolor{g4}59.17 & \cellcolor{g4}56.87 & \cellcolor{g1}65.19
 & \cellcolor{g4}55.17 & \cellcolor{g2}60.53 & \cellcolor{g4}55.97 & \cellcolor{g3}56.12 & \cellcolor{g1}65.83
 & \cellcolor{g3}65.63 & \cellcolor{g2}67.27 & \cellcolor{g4}62.77 & \cellcolor{g4}64.34 & \cellcolor{g1}73.12 \\
\bottomrule
\end{tabular}
\label{tab:main-results-part1}
\end{table*}

\begin{table*}[ht]
\centering
\caption{\textbf{Our Framework Consistently Outperforms on Hard
Benchmarks Where All Baselines Remain Below 42\%.}
Final answer accuracy (\%) shows CoT falling below 40\%, SFT
and DPO failing to surpass CoT, and RAG gains narrowing
relative to Table~\ref{tab:main-results-part1}. Cell shading
indicates per-row rank across conditions. Discussed
in \S\ref{sec:baslineperformance}.}
\fontsize{7.5pt}{9pt}\selectfont
\setlength{\tabcolsep}{4pt}
\renewcommand{\arraystretch}{1.3}
\begin{tabular}{l | c c c c c | c c c c c }
\toprule
\textbf{Model} &
\multicolumn{5}{c|}{\textbf{JEEBench}} &
\multicolumn{5}{c}{\textbf{PhysicsQA}} \\
\cmidrule(lr){2-6} \cmidrule(lr){7-11}
& CoT & RAG & FT & DPO & Ours
& CoT & RAG & FT & DPO & Ours \\
\midrule
Qwen 2.5 1.5B
 & \cellcolor{g4}37.53 & \cellcolor{g2}40.00 & \cellcolor{g3}39.02 & \cellcolor{g4}30.08 & \cellcolor{g1}54.86
 & \cellcolor{g3}30.64 & \cellcolor{g2}36.91 & \cellcolor{g4}23.51 & \cellcolor{g4}24.32 & \cellcolor{g1}49.62 \\
LLaMA 3.2 1B
 & \cellcolor{g4}32.52 & \cellcolor{g3}35.50 & \cellcolor{g4}28.46 & \cellcolor{g2}35.77 & \cellcolor{g1}49.17
 & \cellcolor{g4}21.35 & \cellcolor{g2}28.64 & \cellcolor{g3}23.24 & \cellcolor{g4}18.92 & \cellcolor{g1}39.02 \\
LLaMA 3.2 3B
 & \cellcolor{g4}30.21 & \cellcolor{g2}40.83 & \cellcolor{g4}36.59 & \cellcolor{g3}38.21 & \cellcolor{g1}57.34
 & \cellcolor{g4}27.67 & \cellcolor{g2}31.21 & \cellcolor{g3}27.84 & \cellcolor{g4}25.95 & \cellcolor{g1}46.73 \\
Phi 3.5 Mini 3.8B
 & \cellcolor{g4}35.90 & \cellcolor{g3}41.25 & \cellcolor{g4}38.04 & \cellcolor{g2}41.65 & \cellcolor{g1}50.23
 & \cellcolor{g4}33.35 & \cellcolor{g2}41.59 & \cellcolor{g4}39.19 & \cellcolor{g3}41.49 & \cellcolor{g1}53.22 \\
\bottomrule
\end{tabular}
\label{tab:main-results-part2}
\end{table*}

\section{Experiments}

\subsection{Benchmark Datasets}
We evaluate on five physics benchmarks spanning foundational
to advanced reasoning difficulty: SciEval
Static~\cite{sun2024scieval} (164 MCQ, introductory to
intermediate), MMLU High School and College
Physics~\cite{hendrycks2020measuring} (170 and 118 MCQ,
foundational to undergraduate), JEEBench~\cite{arora2023have}
(123 questions, multi-concept synthesis and mathematical
derivation), and PhysicsQA~\cite{jaiswal2024improvingphysicsreasoninglarge}
(370 problems with verified chain-of-thought solutions
enabling step-level evaluation, introduced in prior work).
All benchmarks are used without filtering or subset selection;
representative samples are provided in
Appendix~\ref{sec:benchmarkss}.

\begin{figure}[t]
\centering
\includegraphics[width=\columnwidth]{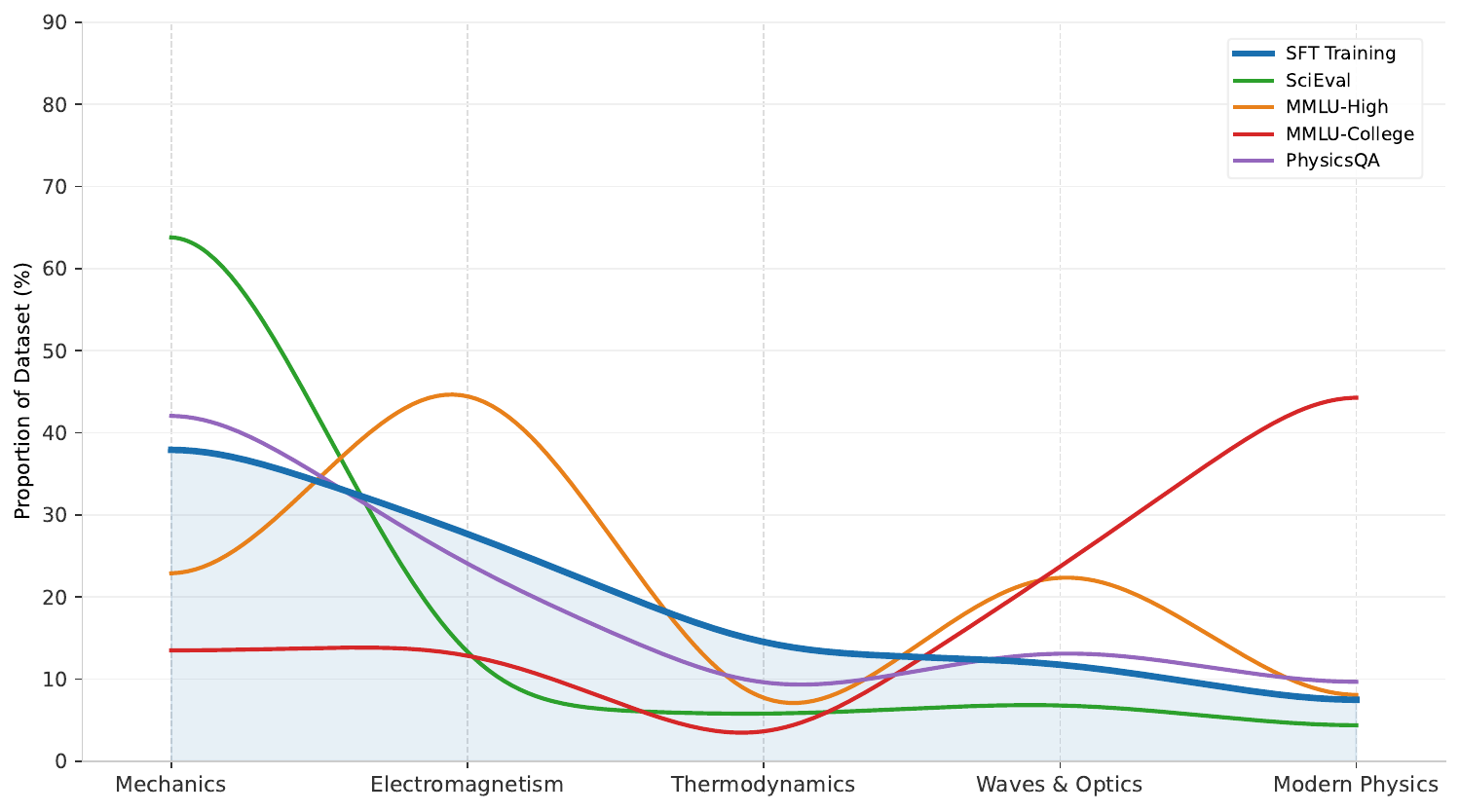}
\caption{\textbf{Common Topic Distributions Confirm Data similarity Across Training and Benchmarks.} Topic distribution (\%) across the SFT training corpus and evaluation benchmarks; the same five domains appear in both but at substantially different proportions.}
\label{fig:domain_dist}
\end{figure}

\subsection{Models}
\label{sec:models}
We evaluate four open-source language models: Qwen 2.5 1.5B
Instruct~\cite{team2024qwen2}, LLaMA 3.2 (1B and 3.2
3B) Instruct~\cite{grattafiori2024llama}, and Phi 3.5 Mini
3.8B Instruct~\cite{microsoft2024phi3}. Each model is
evaluated across five conditions: CoT, RAG, SFT, DPO, and
our framework. All fine-tuned conditions (SFT, DPO, and our
framework) are trained from identical pre-trained base
weights, ensuring performance differences are attributable
to the training objective rather than initialization. CoT
and RAG baselines are evaluated via the Nebius API in
default configuration; RAG augments the same inference with
retrieved context from the physics formula corpus described
in \S\ref{sec:feedback_gen}.


\begin{table*}[ht]
\centering
\caption{\textbf{Calculation Error Reduces Most Consistently;
Conceptual Misapplication Dominates Incorrect Predictions
Across All Conditions.} Reasoning error distribution (\%) among incorrect predictions on Benchmarks; cell shading indicates per-row error severity.
LLaMA 3.2 1B shows simultaneous reduction across all three error types under our framework. Discussed in \S\ref{sec:errortypereduciton}.}
\fontsize{7.5pt}{9.5pt}\selectfont
\setlength{\tabcolsep}{5pt}
\renewcommand{\arraystretch}{1.35}
\begin{tabular}{l | cccc!{\color{black!30}\vrule}c | cccc!{\color{black!30}\vrule}c | cccc!{\color{black!30}\vrule}c}
\toprule

\textbf{Model} &
\multicolumn{5}{c|}{\textbf{MC (\%) — Problem Miscomprehension}} &
\multicolumn{5}{c|}{\textbf{CM (\%) — Conceptual Misapplication}} &
\multicolumn{5}{c}{\textbf{CE (\%) — Calculation Error}} \\

\cmidrule(lr){2-6}\cmidrule(lr){7-11}\cmidrule(lr){12-16}

& CoT & RAG & SFT & DPO & \textbf{Ours}
& CoT & RAG & SFT & DPO & \textbf{Ours}
& CoT & RAG & SFT & DPO & \textbf{Ours} \\

\midrule

Qwen 2.5 1.5B
& \cellcolor{r3}9.3
& \cellcolor{r4}15.0
& \cellcolor{r2}4.6
& \cellcolor{r5}16.1
& \cellcolor{r1}4.1
& \cellcolor{r1}38.6
& \cellcolor{r3}61.4
& \cellcolor{r2}49.5
& \cellcolor{r5}83.9
& \cellcolor{r4}66.1
& \cellcolor{r5}52.1
& \cellcolor{r1}24.5
& \cellcolor{r3}35.0
& \cellcolor{r4}38.9
& \cellcolor{r2}31.4 \\

LLaMA 3.2 1B
& \cellcolor{r2}22.3
& \cellcolor{r4}34.1
& \cellcolor{r5}38.7
& \cellcolor{r3}25.3
& \cellcolor{r1}12.0
& \cellcolor{r3}89.7
& \cellcolor{r2}79.5
& \cellcolor{r4}93.0
& \cellcolor{r5}97.3
& \cellcolor{r1}68.7
& \cellcolor{r5}55.0
& \cellcolor{r3}45.5
& \cellcolor{r4}49.6
& \cellcolor{r2}30.0
& \cellcolor{r1}27.5 \\

LLaMA 3.2 3B
& \cellcolor{r3}7.9
& \cellcolor{r1}5.9
& \cellcolor{r2}6.4
& \cellcolor{r4}9.1
& \cellcolor{r5}12.0
& \cellcolor{r4}73.8
& \cellcolor{r2}47.2
& \cellcolor{r3}55.8
& \cellcolor{r5}80.7
& \cellcolor{r1}42.7
& \cellcolor{r1}37.8
& \cellcolor{r5}44.9
& \cellcolor{r4}40.4
& \cellcolor{r3}38.7
& \cellcolor{r2}38.5 \\

Phi 3.5 Mini 3.8B
& \cellcolor{r4}8.1
& \cellcolor{r1}6.0
& \cellcolor{r3}6.7
& \cellcolor{r5}8.3
& \cellcolor{r2}6.4
& \cellcolor{r2}50.8
& \cellcolor{r5}57.9
& \cellcolor{r1}44.6
& \cellcolor{r3}55.6
& \cellcolor{r4}56.7
& \cellcolor{r4}56.9
& \cellcolor{r2}26.4
& \cellcolor{r5}62.5
& \cellcolor{r3}27.8
& \cellcolor{r1}23.5 \\

\bottomrule
\end{tabular}

\label{tab:error_distribution}
\end{table*}

\subsection{Baseline Setup}

\paragraph{Chain-of-Thought (CoT).}
CoT prompting elicits step-by-step reasoning without any
parameter update. Each model is prompted zero-shot with:
\textit{``You are an expert physics assistant. Given a
question, generate the final solution. Let's think step
by step.''} This reflects each model's intrinsic reasoning
capability under standard inference conditions.

\begin{table}[ht]
\centering
\caption{\textbf{Human Evaluation Confirms Reliability of
Step-Level Assessment Across All Benchmarks.} Cohen's Kappa
between four human evaluators for step-level scoring across
all evaluation samples. Discussed in \S\ref{sec:evaluation}.}
\label{tab:human_agreement}
\fontsize{7.5pt}{9.5pt}\selectfont
\setlength{\tabcolsep}{5pt}
\renewcommand{\arraystretch}{1.35}
\begin{tabular}{l | c | c | c | c}
\toprule
\textbf{Model} &
\textbf{SciEval} & \textbf{MMLU} &
\textbf{PhysicsQA} & \textbf{JEEBench} \\
\midrule
Qwen 2.5 1.5B      & 0.84 & 0.82 & 0.81 & 0.77 \\
LLaMA 3.2 1B       & 0.81 & 0.79 & 0.78 & 0.75 \\
LLaMA 3.2 3B       & 0.87 & 0.85 & 0.84 & 0.82 \\
Phi 3.5 Mini 3.8B  & 0.89 & 0.88 & 0.86 & 0.84 \\
\bottomrule
\end{tabular}
\end{table}

\begingroup
\setlength{\emergencystretch}{3em}

\paragraph{Retrieval-Augmented Generation (RAG).}
RAG augments inference with externally retrieved physics knowledge.
The retrieval corpus contains formulas and physical constants across
five domains, with no worked examples or narrative explanations.
The corpus is divided into 500-character segments with a 50-character
overlap and embedded offline using
\path{text-embedding-ada-002}.
The resulting embeddings are stored in a local FAISS vector store.
At inference time, the three most similar chunks are retrieved using
cosine similarity and inserted into the prompt. No external API is
queried during inference. Full configuration details are provided in
Appendix~\ref{sec:ragconfig}.

\endgroup

\paragraph{Supervised Fine-Tuning (SFT).}
SFT trains each model on the 2,494-problem corpus using
LoRA adapters~\cite{hu2022lora} via the Hugging Face PEFT
framework, pairing each problem with a structured
chain-of-thought solution enforcing explicit reasoning
steps. Training uses AdamW with learning rate
\texttt{5e-5}, batch size 4, gradient accumulation 8, for
3 epochs on a single H100 GPU. Full configuration details
are in Appendix~\ref{sec:sftconfig}.

\paragraph{Direct Preference Optimization (DPO).}
DPO is applied on top of the SFT checkpoint using
preference pairs from the same 2,494-problem corpus.
Chosen solutions are verified for physical correctness;
rejected solutions have errors injected at reasoning
execution steps, directing the preference signal toward
execution rather than problem interpretation. IPO loss is used in place of standard sigmoid DPO as it does not
assume a consistent preference gap across
pairs~\cite{azar2023generaltheoreticalparadigmunderstand}. Training uses AdamW with
learning rate \texttt{2e-6}, batch size 2, gradient
accumulation 16, for 1 epoch on a single H100 GPU. Full
configuration details are in Appendix~\ref{sec:dpoconfig}.

\subsection{Evaluation}
\label{sec:evaluation}

Final answer accuracy is the primary metric across all five
benchmarks. We employ a three-step shadow protocol: Step 1
verifies answer correctness via string matching; Step 2
checks structural compliance across seven required XML tags;
Step 3, applied only to structurally compliant solutions,
attributes errors across the three error types using four
human evaluators with qualifying scores in physics,
chemistry, and mathematics at the national engineering
entrance level. Cohen's Kappa ranges from 0.75 to 0.89
across all models and benchmarks. Full protocol details
and agreement scores are in Table~\ref{tab:human_agreement} and  Appendix~\ref{sec:evalpipeline}.

\begin{figure}[t]
\centering
\includegraphics[width=\columnwidth]{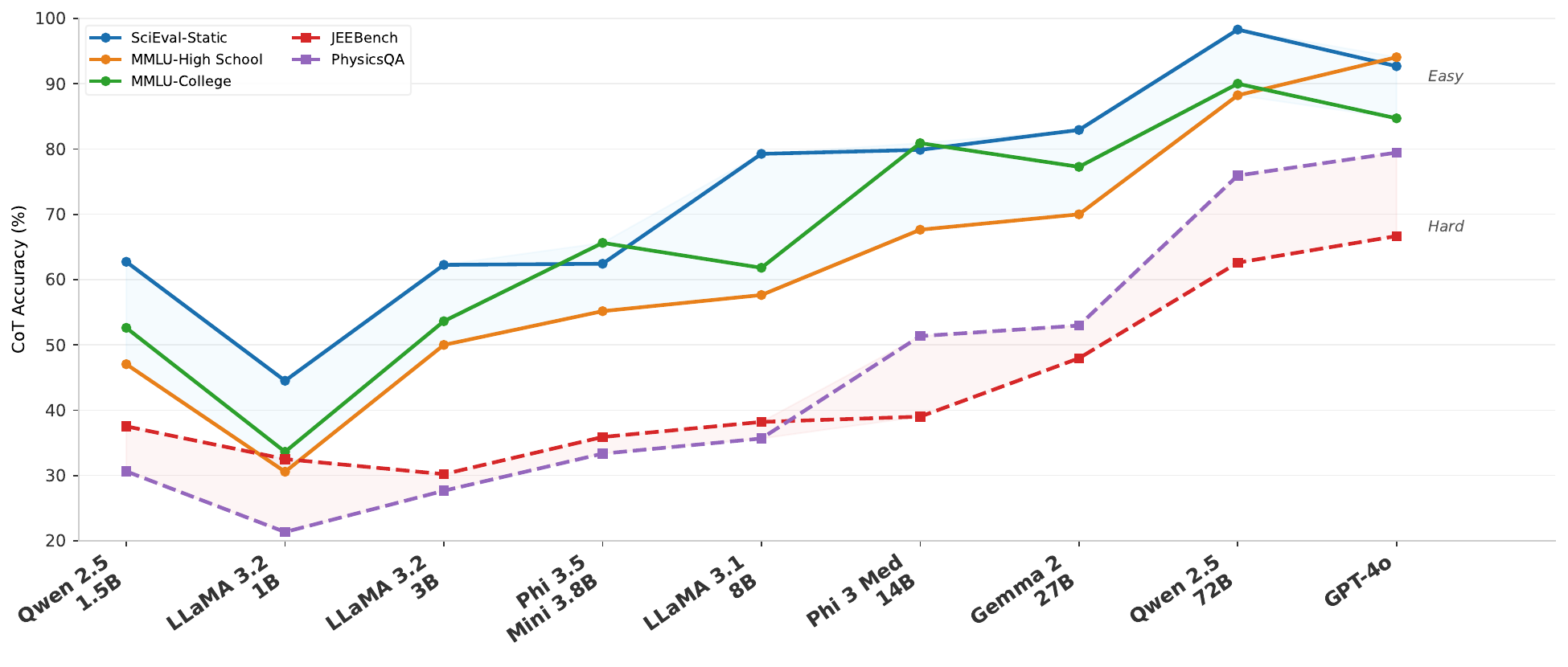}
\caption{\textbf{Accuracy Improves With Model Size; Gains Are Larger on Easy Than Hard Benchmarks.} CoT accuracy (\%) across models (1B--72B and GPT-4o); across SciEval, MMLU-High, MMLU-College (easy) and JEEBench, PhysicsQA (hard). Discussed in
\S\ref{sec:existingmethodsfails}.}
\label{fig:scaling}
\end{figure}

\section{Results}
\label{sec:results}

\subsection{Accuracy Increases With Model Scale.}
\label{sec:existingmethodsfails}

We evaluate CoT accuracy across nine models spanning 1B to
72B parameters across easy benchmarks (SciEval, MMLU-High,
MMLU-College) and hard benchmarks (JEEBench, PhysicsQA),
shown in Figure~\ref{fig:scaling}. Accuracy improves
consistently as model scale increases on both benchmark
types. On easy benchmarks, gains are steep — Qwen 2.5 72B
reaches 98.3\% on SciEval. On hard benchmarks, the same
scaling trend holds but at lower absolute values — no model
exceeds 80\% on PhysicsQA regardless of scale, leaving
step-level physics reasoning as an open challenge across
all model sizes evaluated.

\subsection{Accuracy Across Conditions.}
\label{sec:baslineperformance}

We report final answer accuracy across five conditions
on all five benchmarks across four models
(Tables~\ref{tab:main-results-part1},
\ref{tab:main-results-part2}). Our framework achieves
the highest accuracy across all models and benchmarks;
on easy benchmarks, LLaMA 3.2 3B reaches 81.52\% on
SciEval under our framework against 62.26\% under CoT.
RAG ranks second consistently on both easy and hard
benchmarks, with gains over CoT higher on easy
benchmarks and narrowing on hard benchmarks. SFT
produces accuracy at or below CoT across the majority
of model-benchmark combinations; DPO shows the same
pattern, with the most pronounced degradation observed
for Qwen 2.5 1.5B on MMLU-College and PhysicsQA.
Our framework does not rank first on MMLU-High and
MMLU-College for Qwen 2.5 1.5B, where RAG ranks above
it — the only two such conditions across both tables.

\definecolor{cg1}{RGB}{198,235,202}
\definecolor{cg2}{RGB}{162,220,170}
\definecolor{cg3}{RGB}{108,198,120}
\definecolor{cr1}{RGB}{255,228,228}
\definecolor{cr2}{RGB}{255,195,195}
\definecolor{cr3}{RGB}{255,155,155}

%
\definecolor{headerRow1}{RGB}{220,228,248}   
\definecolor{headerRow2}{RGB}{248,238,225}   
\definecolor{feedbackCol}{RGB}{248,238,225}  

\begin{table*}[ht]
\centering
\caption{\textbf{LLaMA 3.2 1B Benefits Across All Feedback
Channels; Calculation Feedback Shows Largest Gains.}
Ablation on PhysicsQA across feedback channels and baselines;
green shading indicates error reduction under our framework,
red shading indicates error increase. Discussed in
\S\ref{sec:stpeslevelworksandfail}.}
\label{tab:component}
\fontsize{7pt}{9pt}\selectfont
\setlength{\tabcolsep}{3pt}
\renewcommand{\arraystretch}{1.35}
\resizebox{\textwidth}{!}{%
\begin{tabular}{l | cccc | cccc | cccc | cccc}
\toprule

\rowcolor{headerRow1}
\textbf{Ours $-$ Baseline (\%)} &
\multicolumn{4}{c|}{\textbf{Qwen 2.5 1.5B}} &
\multicolumn{4}{c|}{\textbf{LLaMA 3.2 1B}} &
\multicolumn{4}{c|}{\textbf{LLaMA 3.2 3B}} &
\multicolumn{4}{c}{\textbf{Phi 3.5 Mini 3.8B}} \\

\hline

\cellcolor{headerRow2}\textbf{Feedback Type} &
CoT & RAG & SFT & DPO &
CoT & RAG & SFT & DPO &
CoT & RAG & SFT & DPO &
CoT & RAG & SFT & DPO \\

\midrule
\cellcolor{feedbackCol}Problem Statement (MC)
& \cellcolor{cg2}$+$5.2
& \cellcolor{cg3}$+$10.9
& \cellcolor{cg1}$+$0.5
& \cellcolor{cg3}$+$12.0
& \cellcolor{cg2}$+$10.3
& \cellcolor{cg3}$+$22.1
& \cellcolor{cg3}$+$26.7
& \cellcolor{cg3}$+$13.3
& \cellcolor{cr1}$-$4.1
& \cellcolor{cr2}$-$6.1
& \cellcolor{cr2}$-$5.6
& \cellcolor{cr1}$-$2.9
& \cellcolor{cg1}$+$1.7
& \cellcolor{cr1}$-$0.4
& \cellcolor{cg1}$+$0.3
& \cellcolor{cg1}$+$1.9 \\

\cellcolor{feedbackCol}Concept (CM)
& \cellcolor{cr3}$-$27.5
& \cellcolor{cr1}$-$4.7
& \cellcolor{cr3}$-$16.6
& \cellcolor{cg2}$+$17.8
& \cellcolor{cg3}$+$21.0
& \cellcolor{cg2}$+$10.8
& \cellcolor{cg3}$+$24.3
& \cellcolor{cg3}$+$28.6
& \cellcolor{cg3}$+$31.1
& \cellcolor{cg1}$+$4.5
& \cellcolor{cg2}$+$13.1
& \cellcolor{cg3}$+$38.0
& \cellcolor{cr2}$-$5.9
& \cellcolor{cg1}$+$1.2
& \cellcolor{cr3}$-$12.1
& \cellcolor{cr1}$-$1.1 \\

\cellcolor{feedbackCol}Calculation (CE)
& \cellcolor{cg3}$+$20.7
& \cellcolor{cr2}$-$6.9
& \cellcolor{cg1}$+$3.6
& \cellcolor{cg2}$+$7.5
& \cellcolor{cg3}$+$27.5
& \cellcolor{cg3}$+$18.0
& \cellcolor{cg3}$+$22.1
& \cellcolor{cg1}$+$2.5
& \cellcolor{cr1}$-$0.7
& \cellcolor{cg2}$+$6.4
& \cellcolor{cg1}$+$1.9
& \cellcolor{cg1}$+$0.2
& \cellcolor{cg3}$+$33.4
& \cellcolor{cg1}$+$2.9
& \cellcolor{cg3}$+$39.0
& \cellcolor{cg1}$+$4.3 \\

\bottomrule
\end{tabular}}
\end{table*}

\subsection{Reasoning Error Distribution Across Conditions.}
\label{sec:errortypereduciton}

We analyze the distribution of Problem Miscomprehension
(MC), Conceptual Misapplication (CM), and Calculation
Error (CE) among incorrect predictions on PhysicsQA
across all five conditions
(Table~\ref{tab:error_distribution}). DPO produces the
largest CM increase of any condition: for Qwen 2.5 1.5B,
CM rises from 38.6\% under CoT to 83.9\% under DPO, a
45.3 percentage point increase; for LLaMA 3.2 1B, CM
rises from 89.7\% to 97.3\%, the highest proportion
observed across any model or condition in this table.
Under our framework, CE reduces consistently across
three of four models — from 52.1\% to 31.4\% for Qwen
2.5 1.5B, from 55.0\% to 27.5\% for LLaMA 3.2 1B,
and from 56.9\% to 23.5\% for Phi 3.5 Mini 3.8B —
while LLaMA 3.2 3B shows no meaningful change in CE
(37.8\% to 38.5\%). MC reduces for three of four
models under our framework, with the exception of
LLaMA 3.2 3B where MC increases from 7.9\% to 12.0\%.
LLaMA 3.2 1B is the only model where all three error
types reduce simultaneously, with MC dropping from
22.3\% to 12.0\%, CM from 89.7\% to 68.7\%, and CE
from 55.0\% to 27.5\%. CM remains the largest error
category across all models and conditions; under our
framework, CM stays above 42\% across all four models
and increases relative to CoT for Qwen 2.5 1.5B and
Phi 3.5 Mini 3.8B.


\section{Discussion}
\label{sec:discussion}

\subsection{Structured Error Feedback Improvement.}
\label{sec:stpeslevelworksandfail}

The reduction patterns in Table~\ref{tab:component} are
consistent across all four baseline comparisons, not only
relative to CoT. Calculation feedback is the most
consistent of the three channels, reducing Calculation
errors across three of four models regardless of which
baseline is compared. Concept feedback is most effective
where DPO caused the most damage — models with the highest
Conceptual Misapplication proportions under DPO
(Table~\ref{tab:error_distribution}) show the largest
reductions under our framework. Statement feedback is the
least consistent, reducing Problem Miscomprehension errors
for LLaMA 3.2 1B across all baselines but increasing them
for LLaMA 3.2 3B across all baselines. Concept feedback
increases Conceptual Misapplication errors for Qwen 2.5
1.5B relative to most baselines — the only model where
this channel consistently shows negative values. These
negative values indicate that feedback does not always
correct the identified error; when feedback does not
target the actual source of failure, the revised solution
$y_2$ may introduce new errors. LLaMA 3.2 1B, which
shows the highest baseline error rates, benefits from
all three channels across all baseline comparisons.

\subsection{Where Step-Level Reward and Feedback Fall Short.}
\label{sec:wherefails}

Despite consistent accuracy gains, three failure patterns emerge from the results. Conceptual Misapplication errors persist above 42\% across all models under our framework and increase relative to CoT for Qwen 2.5 1.5B and Phi
3.5 Mini 3.8B (Table~\ref{tab:error_distribution}); retrieval feedback supplies the correct governing principle but does not guarantee its correct application.
Problem Miscomprehension errors increase for LLaMA 3.2 3B across all baseline comparisons (Table~\ref{tab:component}), where Conceptual
Misapplication dominates at 73.8\% under CoT and the training loop's prioritization of conceptual correction
may alter problem interpretation as a secondary effect.
The step-level reward signal depends on the verifier
correctly classifying the error type; when classification
is incorrect, feedback is routed to the wrong channel
and $y_2$ is conditioned on feedback that does not
address the actual failure.


\section{Conclusion}

Step-level reward training for physics reasoning in
small language models identifies the first reasoning
error in a solution, generates targeted structured
feedback by error type, and trains the model via
policy gradient with KL regularization, without
access to correct solutions as generation targets
and without preference data construction. Across
four models and five physics benchmarks, the
framework delivers accuracy gains of 17--20\% over
CoT prompting and 10--16\% over all evaluated
baselines, reducing Calculation errors by up to
33.4\% and Problem Miscomprehension errors by up
to 10.3\%. Conceptual Misapplication errors persist
above 42\% across all conditions, remaining the
most consistent unresolved failure mode across all
evaluated settings.

\section{Limitations}
\label{sec:limitations}

Our evaluation is restricted to physics benchmarks in English;
whether the Problem Miscomprehension, Conceptual Misapplication,
and Calculation Error taxonomy and the step-level reward signal
generalize to other scientific domains or non-English corpora
remains untested. Results are reported for single training runs
of 2{,}494 samples over 60 epochs without multi-seed validation.
The framework depends on GPT-4o for both error verification and
structured feedback generation. This coupling means systematic
GPT-4o classification errors would corrupt both the reward signal
and feedback conditioning $y_2$, without being detectable from
accuracy metrics alone. Structured JSON output is required for
reward computation; parse failures default to minimal reward and
their frequency is not reported. The step-level reward depends on
regex-based step counting and LoRA adapters trained on seven-tag
structured solutions; non-standard formatting would produce
incorrect reward signals without detection. The CM and CE channels
invoke GPT-4o via external API while baselines operate without
equivalent resources, introducing a parity gap. Individual
channels were not ablated; contribution is inferred from
error-type reduction patterns. Per-iteration accuracy between
$y_1$ and $y_2$ was not logged. The proportion of training
samples skipped due to $r_1 > \tau$ was not logged during
training; reporting this rate would clarify how often the
early-stopping threshold is triggered and its effect on effective
training corpus size. While topic distribution analysis confirms
common domain proportions between the SFT corpus and evaluation
benchmarks (Figure~\ref{fig:domain_dist}), no exact string
matches were found between the SFT training corpus and JEEBench;
however, formal semantic deduplication has not been performed and
topical overlap cannot be fully ruled out.

\section{Ethical Considerations}

All datasets used for training and evaluation are publicly available and contain no personal data, sensitive information, or harmful content. The physics formula corpus is publicly available educational material. The 2{,}494 training samples
are sourced from existing publicly available JEE physics problem sets; no new human annotation was conducted in this work. GPT-4o (OpenAI API) is used as the external verifier at training time; all four models are accessed via Nebius API for CoT inference. Experiments were conducted on H100 GPUs. Evaluation is
restricted to English-language physics benchmarks; generalization to other languages or scientific domains isvnot claimed. Code, prompts, and training configurations are
provided in the Appendix (\S\ref{sec:appendixmain}).

\bibliography{custom}

\clearpage

\appendix

\onecolumn

\section*{Appendix}
\label{sec:appendixmain}
\begin{tcolorbox}[
    enhanced,
    colback=pastelBlue,
    colframe=pastelBlueFrame,
    colbacktitle=pastelBlueFrame,
    coltitle=white,
    fonttitle=\normalfont\bfseries\large,
    title={A \quad Training Dataset Samples},
    boxrule=0.8pt, arc=5pt,
    left=10pt, right=10pt, top=6pt, bottom=6pt,
]
\normalfont\small
The fine-tuning corpus comprises 2,494 high school-level physics problems
sourced from standard Indian JEE preparation materials. Each problem is paired
with a structured chain-of-thought solution formatted using seven XML tags:
\texttt{<problem\_analysis>}, \texttt{<principle>},
\texttt{<governing\_equation>}, \texttt{<value\_identification>},
\texttt{<substitution>}, \texttt{<calculation>}, and \texttt{<final\_answer>}.
This tag structure enforces explicit separation between physical reasoning and
algebraic execution, and is consistent with the inference prompt used at
evaluation time. Solutions were generated using an automated structuring prompt
(see Appendix~C) and verified for tag-level correctness before inclusion. Four
representative samples are shown below, spanning Thermodynamics,
Electromagnetism, Nuclear Physics, and Classical Mechanics.
\end{tcolorbox}
\label{sec:SFtsample}
\bigskip


\noindent
\colorbox{pastelBlueFrame}{%
  \small\textbf{\textcolor{white}{Sample 1 (ID: 2162)}}%
}

\smallskip

\noindent
\colorbox{pastelBlue}{%
  \parbox{\dimexpr\linewidth-2\fboxsep\relax}{%
    \small\raggedright\itshape
    \textcolor{pastelBlueFrame}{%
      Thermodynamics \textemdash{} Adiabatic compression followed by
      isochoric heating, multi-part symbolic derivation%
    }%
  }%
}
\medskip

\noindent\small
Two moles of an ideal monatomic gas, initially at pressure $p_1$ and volume
$V_1$, undergo adiabatic compression until its volume is $V_2$. The gas is then
given heat $Q$ at constant volume $V_2$. (a)~Sketch the complete process on a
$P$-$V$ diagram. (b)~Find the total work done by the gas, the total change in
its internal energy, and the final temperature of the gas. [Give answers in
terms of $p_1, V_1, V_2, Q$ and $R$.]

\medskip

\begin{tcolorbox}[
    enhanced, breakable,
    colback=pastelBlue,
    colframe=pastelBlueFrame,
    colbacktitle=pastelBlueFrame,
    coltitle=white,
    fonttitle=\normalfont\bfseries\small,
    title={Chain-of-Thought Solution \textemdash{} Sample 1},
    boxrule=0.7pt, arc=4pt,
    left=10pt, right=10pt, top=8pt, bottom=8pt,
]
\ttfamily\small
\textcolor{pastelBlueFrame}{\textbf{<problem\_analysis>}}\\
Two sequential processes: (1) adiabatic compression V1->V2,\\
(2) isochoric heating at V2 with heat Q.\\
Monatomic ideal gas, n=2 mol => gamma=5/3, Cv=3R/2.\\
Find: total W, total DU, final temperature T3.\\
\textcolor{pastelBlueFrame}{\textbf{</problem\_analysis>}}

\medskip
\textcolor{pastelBlueFrame}{\textbf{<principle>}}\\
First Law of Thermodynamics: DQ = DU + DW.\\
Adiabatic: DQ=0 => DU=-W. \quad Isochoric: DW=0 => DU=DQ.\\
\textcolor{pastelBlueFrame}{\textbf{</principle>}}

\medskip
\textcolor{pastelBlueFrame}{\textbf{<governing\_equation>}}\\
Adiabatic: p1*V1\^{}gamma = p2*V2\^{}gamma\\
W\_adiabatic = (p1*V1 - p2*V2) / (gamma - 1)\\
Isochoric: W2=0, DU2=Q\\
\textcolor{pastelBlueFrame}{\textbf{</governing\_equation>}}

\medskip
\textcolor{pastelBlueFrame}{\textbf{<value\_identification>}}\\
n=2 mol, gamma=5/3, Cv=(3/2)R\\
Initial: pressure p1, volume V1 \quad After compression: volume V2\\
Heat added at constant volume: Q\\
\textcolor{pastelBlueFrame}{\textbf{</value\_identification>}}

\medskip
\textcolor{pastelBlueFrame}{\textbf{<substitution>}}\\
Substitute p2=p1*(V1/V2)\^{}gamma into W1:\\
W1 = [p1*V1/(gamma-1)] * [1-(V1/V2)\^{}(gamma-1)]\\
\phantom{W1}= (3*p1*V1/2) * [1-(V1/V2)\^{}(2/3)] \quad [gamma=5/3]\\
W2=0 => W\_total=W1\\
DU1=-W1=(3*p1*V1/2)*[(V1/V2)\^{}(2/3)-1], \quad DU2=Q\\
\textcolor{pastelBlueFrame}{\textbf{</substitution>}}

\medskip
\textcolor{pastelBlueFrame}{\textbf{<calculation>}}\\
W  = (3*p1*V1/2)*[1-(V1/V2)\^{}(2/3)]\\
DU = (3*p1*V1/2)*[(V1/V2)\^{}(2/3)-1] + Q\\
From DU1=n*Cv*(T2-T1) and Q=n*Cv*(T3-T2), with n=2:\\
T3 = Q/(3R) + (p1*V1)/(2R)*(V1/V2)\^{}(2/3)\\
\textcolor{pastelBlueFrame}{\textbf{</calculation>}}

\medskip
\begin{tcolorbox}[
    enhanced,
    colback=pastelGreen,
    colframe=pastelGreenFrame,
    boxrule=0.5pt, arc=3pt,
    left=6pt, right=6pt, top=4pt, bottom=4pt,
]
\ttfamily\small
\textcolor{pastelGreenFrame}{\textbf{<final\_answer>}}\\
W  = (3p1V1/2)[1-(V1/V2)\^{}(2/3)]\\
DU = (3p1V1/2)[(V1/V2)\^{}(2/3)-1] + Q\\
T3 = Q/(3R) + (p1V1/2R)*(V1/V2)\^{}(2/3)\\
\textcolor{pastelGreenFrame}{\textbf{</final\_answer>}}
\end{tcolorbox}
\end{tcolorbox}

\bigskip
\noindent\rule{\linewidth}{0.4pt}
\bigskip


\noindent
\colorbox{pastelBlueFrame}{\small\textbf{\textcolor{white}{Sample 2 (ID: 166)}}}
\quad
\colorbox{pastelBlue}{\small\textit{\textcolor{pastelBlueFrame}{Electromagnetism \textemdash{} Electron in crossed $E$ and $B$ fields, coupled ODEs reducing to cyclotron SHM}}}

\medskip

\noindent\small
An electron is released from the origin where a uniform electric field $E$
exists along the negative $y$-axis and a uniform magnetic field $B$ exists
along the negative $z$-axis. Find the displacement of the electron along the
$y$-axis when its velocity becomes perpendicular to the electric field for
the first time.

\medskip

\begin{tcolorbox}[
    enhanced, breakable,
    colback=pastelBlue,
    colframe=pastelBlueFrame,
    colbacktitle=pastelBlueFrame,
    coltitle=white,
    fonttitle=\normalfont\bfseries\small,
    title={Chain-of-Thought Solution \textemdash{} Sample 2},
    boxrule=0.7pt, arc=4pt,
    left=10pt, right=10pt, top=8pt, bottom=8pt,
]
\ttfamily\small
\textcolor{pastelBlueFrame}{\textbf{<problem\_analysis>}}\\
Electron (charge -e, mass m) starts from rest at origin.\\
E=-E*j, B=-B*k. Velocity in xy-plane: v=vx*i+vy*j.\\
Condition: vy=0 (v perpendicular to E). Find y at first vy=0 after t=0.\\
\textcolor{pastelBlueFrame}{\textbf{</problem\_analysis>}}

\medskip
\textcolor{pastelBlueFrame}{\textbf{<principle>}}\\
Lorentz force: F=q(E+v x B).\\
Coupled ODEs in vx,vy reduce to SHM in vy (cyclotron motion).\\
\textcolor{pastelBlueFrame}{\textbf{</principle>}}

\medskip
\textcolor{pastelBlueFrame}{\textbf{<governing\_equation>}}\\
F = -e(E + v x B)\\
omega = eB/m \quad (cyclotron frequency)\\
vy(t) = (E/B)*sin(omega*t)\\
y(t)  = (E/B*omega)*(1-cos(omega*t))\\
\textcolor{pastelBlueFrame}{\textbf{</governing\_equation>}}

\medskip
\textcolor{pastelBlueFrame}{\textbf{<value\_identification>}}\\
E=electric field magnitude, B=magnetic field magnitude\\
e=electron charge, m=electron mass, omega=eB/m\\
Initial: vx(0)=vy(0)=0\\
\textcolor{pastelBlueFrame}{\textbf{</value\_identification>}}

\medskip
\textcolor{pastelBlueFrame}{\textbf{<substitution>}}\\
F = eE*j + eB*(vy*i - vx*j)\\
dvx/dt=(eB/m)*vy \quad ...(1)\\
dvy/dt=(e/m)*(E-vx*B) \quad ...(2)\\
Differentiate (2), substitute (1): d2vy/dt2 = -omega\^{}2*vy => SHM\\
With vy(0)=0, dvy/dt(0)=eE/m:\\
vy(t)=(E/B)*sin(omega*t), \quad y(t)=(E/B*omega)*(1-cos(omega*t))\\
\textcolor{pastelBlueFrame}{\textbf{</substitution>}}

\medskip
\textcolor{pastelBlueFrame}{\textbf{<calculation>}}\\
vy=0 first at: omega*t=pi => t*=pi*m/(eB)\\
y at t=t*: y=(E/B*omega)*(1-cos(pi))=2E/(B*omega)=2Em/(eB\^{}2)\\
\textcolor{pastelBlueFrame}{\textbf{</calculation>}}

\medskip
\begin{tcolorbox}[
    enhanced,
    colback=pastelGreen,
    colframe=pastelGreenFrame,
    boxrule=0.5pt, arc=3pt,
    left=6pt, right=6pt, top=4pt, bottom=4pt,
]
\ttfamily\small
\textcolor{pastelGreenFrame}{\textbf{<final\_answer>}}\\
y = 2Em/(eB\^{}2)\\
\textcolor{pastelGreenFrame}{\textbf{</final\_answer>}}
\end{tcolorbox}
\end{tcolorbox}

\bigskip
\noindent\rule{\linewidth}{0.4pt}
\bigskip


\noindent
\colorbox{pastelBlueFrame}{\small\textbf{\textcolor{white}{Sample 3 (ID: 889)}}}
\quad
\colorbox{pastelBlue}{\small\textit{\textcolor{pastelBlueFrame}{Nuclear Physics \textemdash{} Radioactive decay chain, rate equations, population maximization (JEE 2001)}}}

\medskip

\noindent\small
A radioactive nucleus $X$ decays to $Y$ with
$\lambda_X = 0.1~\mathrm{s}^{-1}$. $Y$ decays to stable $Z$ with
$\lambda_Y = \tfrac{1}{30}~\mathrm{s}^{-1}$. Initially only $X$ nuclei are
present, $N_0 = 10^{20}$. (a)~Set up rate equations for $N_X$, $N_Y$, $N_Z$.
(b)~Given $N_Y(t) = \frac{N_0\lambda_X}{\lambda_X - \lambda_Y}
[e^{-\lambda_Y t} - e^{-\lambda_X t}]$,
find $t$ at which $N_Y$ is maximum and determine $N_X$, $N_Z$ at that instant.

\medskip

\begin{tcolorbox}[
    enhanced, breakable,
    colback=pastelBlue,
    colframe=pastelBlueFrame,
    colbacktitle=pastelBlueFrame,
    coltitle=white,
    fonttitle=\normalfont\bfseries\small,
    title={Chain-of-Thought Solution \textemdash{} Sample 3},
    boxrule=0.7pt, arc=4pt,
    left=10pt, right=10pt, top=8pt, bottom=8pt,
]
\ttfamily\small
\textcolor{pastelBlueFrame}{\textbf{<problem\_analysis>}}\\
Decay chain X->Y->Z (Z stable). N\_X(0)=1e20, N\_Y(0)=N\_Z(0)=0.\\
Part (a): write coupled ODEs. Part (b): maximize N\_Y(t).\\
\textcolor{pastelBlueFrame}{\textbf{</problem\_analysis>}}

\medskip
\textcolor{pastelBlueFrame}{\textbf{<principle>}}\\
Radioactive decay rate proportional to current population.\\
N\_Y sourced by X decay, depleted by Y decay.\\
\textcolor{pastelBlueFrame}{\textbf{</principle>}}

\medskip
\textcolor{pastelBlueFrame}{\textbf{<governing\_equation>}}\\
dN\_X/dt = -lambda\_X * N\_X \hfill ...(i)\\
dN\_Y/dt =  lambda\_X*N\_X - lambda\_Y*N\_Y \hfill ...(ii)\\
dN\_Z/dt =  lambda\_Y * N\_Y \hfill ...(iii)\\
\textcolor{pastelBlueFrame}{\textbf{</governing\_equation>}}

\medskip
\textcolor{pastelBlueFrame}{\textbf{<value\_identification>}}\\
lambda\_X=0.1 s\^{}-1, \quad lambda\_Y=1/30 s\^{}-1, \quad N0=1e20\\
N\_X(t)=N0*exp(-lambda\_X*t)\\
\textcolor{pastelBlueFrame}{\textbf{</value\_identification>}}

\medskip
\textcolor{pastelBlueFrame}{\textbf{<substitution>}}\\
dN\_Y/dt=0 => lambda\_X*N\_X = lambda\_Y*N\_Y \hfill ...(iv)\\
Substitute N\_X, N\_Y into (iv) and simplify:\\
lambda\_X/lambda\_Y = exp((lambda\_X-lambda\_Y)*t)\\
t* = ln(lambda\_X/lambda\_Y) / (lambda\_X-lambda\_Y)\\
\textcolor{pastelBlueFrame}{\textbf{</substitution>}}

\medskip
\textcolor{pastelBlueFrame}{\textbf{<calculation>}}\\
t* = ln(3)/(2/30) = 15*ln(3) = 16.48 s\\
N\_X = 1e20*exp(-0.1*16.48) = 1.92e19\\
N\_Y = N\_X*lambda\_X/lambda\_Y = 1.92e19*3 = 5.76e19\\
N\_Z = N0 - N\_X - N\_Y = 2.32e19\\
\textcolor{pastelBlueFrame}{\textbf{</calculation>}}

\medskip
\begin{tcolorbox}[
    enhanced,
    colback=pastelGreen,
    colframe=pastelGreenFrame,
    boxrule=0.5pt, arc=3pt,
    left=6pt, right=6pt, top=4pt, bottom=4pt,
]
\ttfamily\small
\textcolor{pastelGreenFrame}{\textbf{<final\_answer>}}\\
t* = 15ln(3) = 16.48 s\\
N\_X = 1.92 x 10\^{}19, \quad N\_Z = 2.32 x 10\^{}19\\
\textcolor{pastelGreenFrame}{\textbf{</final\_answer>}}
\end{tcolorbox}
\end{tcolorbox}

\bigskip
\noindent\rule{\linewidth}{0.4pt}
\bigskip


\noindent
\colorbox{pastelBlueFrame}{\small\textbf{\textcolor{white}{Sample 4 (ID: 2248)}}}
\quad
\colorbox{pastelBlue}{\small\textit{\textcolor{pastelBlueFrame}{Classical Mechanics \textemdash{} Compound Atwood machine, constraint analysis, Newton's laws}}}

\medskip

\noindent\small
Three blocks of masses $m_1$, $m_2$ and $m_3$ are connected via a compound
pulley system (pulley $B$ is itself suspended from a string over a fixed
pulley). All surfaces are frictionless; strings and pulleys are massless.
Find the acceleration of $m_1$.

\medskip

\begin{tcolorbox}[
    enhanced, breakable,
    colback=pastelBlue,
    colframe=pastelBlueFrame,
    colbacktitle=pastelBlueFrame,
    coltitle=white,
    fonttitle=\normalfont\bfseries\small,
    title={Chain-of-Thought Solution \textemdash{} Sample 4},
    boxrule=0.7pt, arc=4pt,
    left=10pt, right=10pt, top=8pt, bottom=8pt,
]
\ttfamily\small
\textcolor{pastelBlueFrame}{\textbf{<problem\_analysis>}}\\
m1 on frictionless table, string over fixed pulley to movable pulley B.\\
m2 and m3 hang from B on either side.\\
a0=accel. of m1 (right)=downward accel. of B.\\
a=accel. of m2/m3 relative to B.\\
Ground-frame: m2->(a0-a) down, m3->(a0+a) down.\\
\textcolor{pastelBlueFrame}{\textbf{</problem\_analysis>}}

\medskip
\textcolor{pastelBlueFrame}{\textbf{<principle>}}\\
Newton's Second Law for each body.\\
Fixed string lengths impose kinematic constraints.\\
Massless pulley B: net force=0.\\
\textcolor{pastelBlueFrame}{\textbf{</principle>}}

\medskip
\textcolor{pastelBlueFrame}{\textbf{<governing\_equation>}}\\
Pulley B: 2T'=T => T'=T/2\\
m1: T=m1*a0 \hfill ...(ii)\\
m2: m2*g-T/2=m2*(a0-a) \hfill ...(iii)\\
m3: m3*g-T/2=m3*(a0+a) \hfill ...(iv)\\
\textcolor{pastelBlueFrame}{\textbf{</governing\_equation>}}

\medskip
\textcolor{pastelBlueFrame}{\textbf{<value\_identification>}}\\
m1,m2,m3=block masses\\
T=upper string tension, T'=T/2=lower string tension\\
a0=accel. of m1 and B, \quad a=relative accel. of m2/m3 w.r.t. B\\
\textcolor{pastelBlueFrame}{\textbf{</value\_identification>}}

\medskip
\textcolor{pastelBlueFrame}{\textbf{<substitution>}}\\
Substitute T=m1*a0 into (iii),(iv):\\
a0-a = g - m1*a0/(2*m2) \hfill ...(iii')\\
a0+a = g - m1*a0/(2*m3) \hfill ...(iv')\\
Add (iii') and (iv') to eliminate a:\\
2*a0 = 2g - (m1*a0/2)*(1/m2+1/m3)\\
\textcolor{pastelBlueFrame}{\textbf{</substitution>}}

\medskip
\textcolor{pastelBlueFrame}{\textbf{<calculation>}}\\
a0*[1+(m1/4)*(1/m2+1/m3)] = g\\
a0 = g / [1+(m1/4)*(1/m2+1/m3)]\\
\textcolor{pastelBlueFrame}{\textbf{</calculation>}}

\medskip
\begin{tcolorbox}[
    enhanced,
    colback=pastelGreen,
    colframe=pastelGreenFrame,
    boxrule=0.5pt, arc=3pt,
    left=6pt, right=6pt, top=4pt, bottom=4pt,
]
\ttfamily\small
\textcolor{pastelGreenFrame}{\textbf{<final\_answer>}}\\
a0 = g / [1 + (m1/4)*(1/m2 + 1/m3)]\\
\textcolor{pastelGreenFrame}{\textbf{</final\_answer>}}
\end{tcolorbox}
\end{tcolorbox}

\bigskip

\begin{tcolorbox}[
    enhanced,
    colback=pastelBlueDeep,
    colframe=pastelBlueFrame,
    boxrule=0.5pt, arc=3pt,
    left=8pt, right=8pt, top=5pt, bottom=5pt,
]
\small\normalfont
All 2,494 training samples follow this structure. The dataset spans six physics
domains: Mechanics, Electromagnetism, Thermodynamics, Waves, Optics, and Modern
Physics. Difficulty is restricted to high school and early undergraduate level,
consistent with the JEE source material.
\end{tcolorbox}

\twocolumn

\onecolumn


\section{Verifier Prompt}
\label{sec:appendix_config}

The external verifier (GPT-4o) receives the model-generated
solution $y_1$ and the ground truth solution $y^*$ and returns
a structured JSON object containing the first error step
$e_{\text{first}}$, error type $c_1 \in \{\text{MC, CM, CE}\}$,
and a natural language explanation $\text{expl}_1$.

\begin{lstlisting}[
  language=Python,
  basicstyle=\fontsize{7pt}{9pt}\ttfamily,
  breaklines=true,
  frame=single,
  framesep=4pt,
  xleftmargin=6pt,
  xrightmargin=6pt,
  caption={Verifier prompt used by GPT-4o to identify and
           classify the first reasoning error.},
  label={lst:verifier}
]
eval_prompt = """
You are a physics reasoning evaluator. Identify and explain
the first reasoning error (if any) in the model's solution.

Classify the error as one of:
- "Problem Miscomprehension": misreading given values or
  problem conditions before any physics is applied.
- "Conceptual Error": wrong physics law, formula, or
  principle applied.
- "Computational Error": arithmetic, calculus, or algebraic
  mistake.

Return JSON:
{
  "error_step": <first error step number>,
  "error_type": "<one of the three categories>",
  "error_explanation": "<plain text explanation>"
}

If no error:
{
  "error_step": 0,
  "error_type": "no_error",
  "error_explanation": "No error in your solution."
}

Question: {question}
Ground Truth CoT: {cot_solution}
LLM Generated Solution: {error_solution}
"""
\end{lstlisting}


\section{Step-Level Reward Computation}
\label{sec:appendix_reward}

The reward $r = e_{\text{first}} / (n + 1)$ is computed from
the verifier output. A solution with no error receives $r = 1.0$.
Solutions where the first error occurs later in the chain
receive higher reward.

\begin{lstlisting}[
  language=Python,
  basicstyle=\fontsize{7pt}{9pt}\selectfont\ttfamily,
  breaklines=true,
  frame=single,
  framesep=4pt,
  xleftmargin=6pt,
  xrightmargin=6pt,
  caption={Step-level reward computation from verifier output.},
  label={lst:reward}
]
def generate_reward(sample, eval_model):
    decoded = eval_model.invoke(
        eval_prompt.format(
            question=sample['question'],
            cot_solution=sample['cot_solution'],
            error_solution=sample['incorrect_solution']
        )
    ).content
    decoded = ast.literal_eval(decoded)
    error_step = decoded['error_step']

    if error_step == 0:
        reward = 1.0  # No error - full reward
    else:
        step_pattern = r"(?:#+\s*)?Step\s+\d+:"
        matches = re.findall(
            step_pattern,
            sample['incorrect_solution']
        )
        total_steps = len(matches) if matches else 1

        # Earlier failure means lower reward
        reward = error_step / (total_steps + 1)

    return reward, decoded
\end{lstlisting}

\newpage
\section{Structured Feedback Generation}
\label{sec:appendix_feedback}

Feedback is routed to one of three channels based on the
classified error type. MC errors receive statement-level
feedback directly from GPT-4o. CM errors invoke
\texttt{FeedbackRAGAgent} using LLaMA 3.2 3B via Nebius API
with ChromaDB retrieval over the NEET physics corpus.
CE errors invoke \texttt{CodeAgent} using LLaMA 3.2 3B via
Nebius API to execute corrected Python code.

\begin{lstlisting}[
  language=Python,
  basicstyle=\fontsize{7pt}{9pt}\selectfont\ttfamily,
  breaklines=true,
  breakatwhitespace=false,
  frame=single,
  framesep=4pt,
  xleftmargin=6pt,
  xrightmargin=6pt,
  caption={Feedback routing by error type. MC errors receive
           statement feedback; CM errors invoke retrieval-
           augmented feedback; CE errors invoke code-execution
           feedback.},
  label={lst:feedback}
]
def generate_feedback(decoded, question):
    if decoded['error_step'] == 0:
        return "There is no error in your solution."

    if decoded['error_type'] == 'Problem Miscomprehension':
        # Statement feedback - direct prompt
        feedback = feedback_prompt.format(
            error_step=decoded['error_step'],
            error_type=decoded['error_type'],
            error_explanation=decoded['error_explanation']
        )

    elif decoded['error_type'] == 'Conceptual Error':
        # Retrieval-augmented feedback via LLaMA 3.2 3B
        agent = FeedbackRAGAgent(
            pdf_path="physics-formulas-for-neet-2023.pdf",
            chroma_path="./chroma_db_feedback",
            embedding_model_name=
                "sentence-transformers/all-MiniLM-L6-v2",
            llm_model_name=
                "meta-llama/Llama-3.2-3B-Instruct",
            llm_api_base=
                "https://api.studio.nebius.com/v1/"
        )
        feedback = agent.run_feedback_cycle(
            question=question,
            error_step=decoded["error_step"],
            error_explanation=decoded["error_explanation"]
        )

    elif decoded['error_type'] == 'Computational Error':
        # Code-execution feedback via LLaMA 3.2 3B
        agent = CodeAgent(
            question,
            decoded['error_step'],
            decoded['error_explanation'],
            llm_model_name=
                "meta-llama/Llama-3.2-3B-Instruct",
            llm_api_base=
                "https://api.studio.nebius.com/v1/"
        )
        feedback = agent.run()

    return feedback
\end{lstlisting}
\clearpage

\section{Policy Gradient Training Objective}
\label{sec:appendix_training}

The policy is updated using REINFORCE with KL regularization.
The loss penalizes the difference between attempt 1 and
attempt 2 log-probabilities weighted by the step-level reward,
minus a KL divergence term from the reference policy.

\begin{lstlisting}[
  language=Python,
  basicstyle=\fontsize{7pt}{9pt}\ttfamily,
  breaklines=true,
  frame=single,
  framesep=4pt,
  xleftmargin=6pt,
  xrightmargin=6pt,
  caption={Policy gradient loss with KL regularization.
           $r_2$ is the step-level reward on the revised
           solution $y_2$; \texttt{beta2} controls the KL
           penalty weight.},
  label={lst:loss}
]
loss = -(
    (
        (att2_log_probs * attempt2_answer_mask[:, 1:]).sum(-1)
        / attempt2_answer_mask[:, 1:].sum(-1)
        -
        (att1_log_probs[:, 1:] * attempt1_answer_mask[:, 1:]).sum(-1)
        / attempt1_answer_mask[:, 1:].sum(-1)
    ) * reward2_tensor - beta2 * kl_div
).mean()
\end{lstlisting}


\section{Training Hyperparameters}
\label{sec:appendix_hyperparams}

\begin{table}[ht]
\centering
\fontsize{8pt}{10pt}\selectfont
\setlength{\tabcolsep}{20pt}
\renewcommand{\arraystretch}{1.4}
\begin{tabular}{l | r}
\toprule
\rowcolor{headerGroup}
\textbf{Hyperparameter} & \textbf{Value} \\
\midrule
Batch size              & 5 \\
Stage 1 epochs (SFT)    & 10 \\
Stage 2 epochs (RL)     & 60 \\
Learning rate           & $5 \times 10^{-6}$ \\
KL penalty $\beta$      & 0.1 \\
Early stopping $\tau$   & 0.9 \\
Max prompt length       & 512 tokens \\
Max attempt length      & 1,240 tokens \\
LoRA rank               & 16 \\
LoRA $\alpha$           & 32 \\
Temperature             & 1.0 \\
\bottomrule
\end{tabular}
\caption{Training hyperparameters used across all four models.
Discussed in \S\ref{sec:methodology}.}
\label{tab:hyperparams}
\end{table}
\twocolumn


\onecolumn

\begin{tcolorbox}[
    enhanced,
    colback=pastelTeal,
    colframe=pastelTealFrame,
    colbacktitle=pastelTealFrame,
    coltitle=white,
    fonttitle=\normalfont\bfseries\large,
    title={B \quad SFT Training Configuration},
    boxrule=0.8pt, arc=5pt,
    left=10pt, right=10pt, top=6pt, bottom=6pt,
]
\normalfont\small
Supervised fine-tuning is performed on all four ultra-small models using
LoRA~\cite{hu2022lora} adapters via the Hugging Face PEFT framework. The
warm-up stage trains each model on the 2,494-problem fine-tuning corpus for
a fixed number of epochs, after which the LoRA adapter is frozen and used as
the initialisation point for DPO training. All experiments use a fixed seed
of 42 throughout for full reproducibility. Effective batch size equals
\texttt{per\_device\_batch\_size} $\times$ \texttt{gradient\_accumulation\_steps}.
All models use cosine learning rate scheduling, \texttt{weight\_decay=0.01},
\texttt{bf16=True}, and are trained on a single H100 GPU.
\end{tcolorbox}
\label{sec:sftconfig}
\bigskip

\begin{tcolorbox}[
    enhanced,
    colback=pastelTeal,
    colframe=pastelTealFrame,
    colbacktitle=pastelTealFrame,
    coltitle=white,
    fonttitle=\normalfont\bfseries\small,
    title={Shared Configuration \textemdash{} All Models},
    boxrule=0.7pt, arc=4pt,
    left=10pt, right=10pt, top=8pt, bottom=8pt,
]
\ttfamily\small
\textcolor{pastelTealFrame}{\textbf{task\_type}}          = CAUSAL\_LM\\
\textcolor{pastelTealFrame}{\textbf{target\_modules}}     = q\_proj, k\_proj, v\_proj, o\_proj,\\
\phantom{\textcolor{pastelTealFrame}{\textbf{target\_modules}}     = }gate\_proj, up\_proj, down\_proj\\
\textcolor{pastelTealFrame}{\textbf{lora\_dropout}}       = 0.05\\
\textcolor{pastelTealFrame}{\textbf{bias}}                = none\\
\textcolor{pastelTealFrame}{\textbf{packing}}             = True\\
\textcolor{pastelTealFrame}{\textbf{lr\_scheduler\_type}} = cosine\\
\textcolor{pastelTealFrame}{\textbf{warmup\_ratio}}       = 0.05\\
\textcolor{pastelTealFrame}{\textbf{weight\_decay}}       = 0.01\\
\textcolor{pastelTealFrame}{\textbf{precision}}           = bf16\\
\textcolor{pastelTealFrame}{\textbf{hardware}}            = single H100 GPU\\
\textcolor{pastelTealFrame}{\textbf{seed}}                = 42\\
\textcolor{pastelTealFrame}{\textbf{alpha rule}}          = alpha = 2 $\times$ r \quad (consistent LoRA update magnitude)
\end{tcolorbox}

\bigskip

\begin{tcolorbox}[
    enhanced,
    colback=pastelTealDeep,
    colframe=pastelTealFrame,
    colbacktitle=pastelTealFrame,
    coltitle=white,
    fonttitle=\normalfont\bfseries\small,
    title={Per-Model Hyperparameters},
    boxrule=0.7pt, arc=4pt,
    left=6pt, right=6pt, top=8pt, bottom=8pt,
]
\centering\small
\renewcommand{\arraystretch}{1.3}
\setlength{\tabcolsep}{6pt}
\begin{tabular}{lcccccccc}
\toprule
\rowcolor{pastelTealFrame!80}
\textcolor{white}{\textbf{Model}} &
\textcolor{white}{\textbf{r}} &
\textcolor{white}{\textbf{alpha}} &
\textcolor{white}{\textbf{LR}} &
\textcolor{white}{\textbf{Epochs}} &
\textcolor{white}{\textbf{Batch}} &
\textcolor{white}{\textbf{Grad Accum}} &
\textcolor{white}{\textbf{Eff.\ Batch}} &
\textcolor{white}{\textbf{Seq Len}} \\
\midrule
\rowcolor{pastelTeal}
Qwen 2.5 1.5B     & 32 & 64  & 1e-4 & 3 & 4 & 8  & 32 & 4096 \\
\rowcolor{white}
LLaMA 3.2 1B      & 32 & 64  & 1e-4 & 3 & 4 & 8  & 32 & 4096 \\
\rowcolor{pastelTeal}
LLaMA 3.2 3B      & 32 & 64  & 1e-4 & 3 & 4 & 8  & 32 & 4096 \\
\rowcolor{white}
Phi 3.5 Mini 3.8B & 64 & 128 & 5e-5 & 3 & 2 & 16 & 32 & 4096 \\
\bottomrule
\end{tabular}

\vspace{0.5em}
\noindent\small\normalfont
All models share \texttt{max\_grad\_norm=1.0}.
\texttt{alpha = 2\,$\times$\,r} is maintained throughout for consistent
LoRA update magnitude scaling and cross-model comparability.
\end{tcolorbox}

\bigskip

\begin{tcolorbox}[
    enhanced,
    breakable,
    colback=pastelTeal,
    colframe=pastelTealFrame,
    colbacktitle=pastelTealFrame,
    coltitle=white,
    fonttitle=\normalfont\bfseries\small,
    title={Training Prompt Format \textemdash{}
           Inference-Consistent Instruction Template},
    boxrule=0.7pt,
    arc=4pt,
    left=10pt,
    right=10pt,
    top=8pt,
    bottom=8pt,
]

\ttfamily\small\raggedright
\setlength{\parindent}{0pt}
\setlength{\parskip}{2pt}

You are an expert physics assistant. You are given a physics problem.
Your task is to generate the complete solution using exactly the
following seven XML tags in order:

\medskip

\textcolor{pastelTealFrame}{%
    \textbf{<problem\_analysis>}%
}\par
\hspace*{1em}Restate the givens, unknowns, domain, and assumptions.

\smallskip

\textcolor{pastelTealFrame}{%
    \textbf{<principle>}%
}\par
\hspace*{1em}Name the physical law in plain English.

\smallskip

\textcolor{pastelTealFrame}{%
    \textbf{<governing\_equation>}%
}\par
\hspace*{1em}Give the symbolic formula only and define all symbols.

\smallskip

\textcolor{pastelTealFrame}{%
    \textbf{<value\_identification>}%
}\par
\hspace*{1em}List every given value explicitly with its units.

\smallskip

\textcolor{pastelTealFrame}{%
    \textbf{<substitution>}%
}\par
\hspace*{1em}Substitute the values line by line and show each step.

\smallskip

\textcolor{pastelTealFrame}{%
    \textbf{<calculation>}%
}\par
\hspace*{1em}Show the arithmetic step by step, carrying the units
throughout.

\smallskip

\textcolor{pastelTealFrame}{%
    \textbf{<final\_answer>}%
}\par
\hspace*{1em}Give the answer with units and use
\textbackslash boxed\{\}.

\medskip

\textbf{Critical rules:}

\smallskip

\hangindent=1.2em
\hangafter=1
-- Never skip a tag and never add extra tags.\par

\hangindent=1.2em
\hangafter=1
-- The
\textcolor{pastelTealFrame}{%
    \textbf{<governing\_equation>}%
}
tag must use symbolic form only, with no numerical values.\par

\hangindent=1.2em
\hangafter=1
-- The
\textcolor{pastelTealFrame}{%
    \textbf{<substitution>}%
}
tag must reference the values listed in
\textcolor{pastelTealFrame}{%
    \textbf{<value\_identification>}%
}
exactly.\par

\hangindent=1.2em
\hangafter=1
-- If the solution has multiple parts, repeat the complete tag
structure for each part.

\medskip

\textbf{Question:}\par
\{question\}

\end{tcolorbox}

\bigskip
\begin{tcolorbox}[
    enhanced,
    breakable,
    colback=pastelTeal,
    colframe=pastelTealFrame,
    colbacktitle=pastelTealFrame,
    coltitle=white,
    fonttitle=\normalfont\bfseries\small,
    title={SFT Solution Structuring Prompt \textemdash{}
           Automated XML Conversion},
    boxrule=0.7pt,
    arc=4pt,
    left=10pt,
    right=10pt,
    top=8pt,
    bottom=8pt,
]

\ttfamily\small\raggedright
\setlength{\parindent}{0pt}
\setlength{\parskip}{2pt}

You are converting physics solutions into a structured educational
format. Your output must use exactly these seven XML tags in exactly
this order. Never skip a tag. Never add extra tags.

\medskip

\textcolor{pastelTealFrame}{%
    \textbf{<problem\_analysis>}%
}\par
\hspace*{1em}Restate the givens, unknowns, domain, and assumptions.

\smallskip

\textcolor{pastelTealFrame}{%
    \textbf{<principle>}%
}\par
\hspace*{1em}Name the physical law in plain English.

\smallskip

\textcolor{pastelTealFrame}{%
    \textbf{<governing\_equation>}%
}\par
\hspace*{1em}Give the symbolic formula only and define all symbols.

\smallskip

\textcolor{pastelTealFrame}{%
    \textbf{<value\_identification>}%
}\par
\hspace*{1em}List every given value explicitly with its units.

\smallskip

\textcolor{pastelTealFrame}{%
    \textbf{<substitution>}%
}\par
\hspace*{1em}Substitute the values line by line and show each step.

\smallskip

\textcolor{pastelTealFrame}{%
    \textbf{<calculation>}%
}\par
\hspace*{1em}Show the arithmetic step by step, carrying the units
throughout.

\smallskip

\textcolor{pastelTealFrame}{%
    \textbf{<final\_answer>}%
}\par
\hspace*{1em}Give the answer with units and use
\textbackslash boxed\{\}.

\medskip

\textbf{Critical rules:}

\smallskip

\hangindent=1.2em
\hangafter=1
-- The
\textcolor{pastelTealFrame}{%
    \textbf{<value\_identification>}%
}
tag must explicitly list EVERY numerical value from the problem
statement.\par

\hangindent=1.2em
\hangafter=1
-- The
\textcolor{pastelTealFrame}{%
    \textbf{<governing\_equation>}%
}
tag must contain symbolic expressions only, with no numerical
values.\par

\hangindent=1.2em
\hangafter=1
-- The
\textcolor{pastelTealFrame}{%
    \textbf{<substitution>}%
}
tag must reference the values listed in
\textcolor{pastelTealFrame}{%
    \textbf{<value\_identification>}%
}
exactly.\par

\hangindent=1.2em
\hangafter=1
-- Preserve all LaTeX mathematical notation.\par

\hangindent=1.2em
\hangafter=1
-- If the solution has multiple parts, repeat the complete tag
structure for each part.

\medskip

\textbf{Problem:}\par
\{question\}

\smallskip

\textbf{Original Solution:}\par
\{solution\}

\end{tcolorbox}

\bigskip

\begin{tcolorbox}[
    enhanced,
    colback=pastelTealDeep,
    colframe=pastelTealFrame,
    boxrule=0.5pt, arc=3pt,
    left=8pt, right=8pt, top=5pt, bottom=5pt,
]
\small\normalfont
Training time per model ranged from 45 to 60 minutes on a single H100 GPU.
The resulting LoRA adapter is saved as a checkpoint and used directly as the
starting point for DPO training, with no intermediate merging into the base
model weights.
\end{tcolorbox}

\twocolumn


\onecolumn

\begin{tcolorbox}[
    enhanced,
    colback=pastelPurple,
    colframe=pastelPurpleFrame,
    colbacktitle=pastelPurpleFrame,
    coltitle=white,
    fonttitle=\normalfont\bfseries\large,
    title={C \quad DPO Dataset Samples and Configuration},
    boxrule=0.8pt, arc=5pt,
    left=10pt, right=10pt, top=6pt, bottom=6pt,
]
\normalfont\small
Direct Preference Optimization is applied on top of the SFT LoRA checkpoint.
Each DPO training pair consists of a \textit{chosen} solution (fully correct,
all seven tags structurally and physically consistent) and a \textit{rejected}
solution (structurally identical but with one critical error injected at a
single tag). The error propagates as a domino through all subsequent tags,
producing a fluent, complete, but physically incorrect solution. The
\texttt{<problem\_analysis>} and \texttt{<principle>} tags are held identical
across chosen and rejected in every pair, isolating the preference signal to
the reasoning execution rather than problem interpretation.
\end{tcolorbox}
\label{sec:dpoconfig}
\bigskip

\begin{tcolorbox}[
    enhanced,
    breakable,
    colback=pastelPurple,
    colframe=pastelPurpleFrame,
    colbacktitle=pastelPurpleFrame,
    coltitle=white,
    fonttitle=\normalfont\bfseries\small,
    title={DPO Rejected Sample Generation Prompt},
    boxrule=0.7pt,
    arc=4pt,
    left=10pt,
    right=10pt,
    top=8pt,
    bottom=8pt,
]

\ttfamily\small\raggedright
\setlength{\parindent}{0pt}
\setlength{\parskip}{2pt}

You are a physics expert tasked with generating a REJECTED (incorrect)
response for a physics problem. This is meant to be used as a negative
example in Direct Preference Optimization (DPO).

\medskip

Your task is to generate a rejected solution that:

\smallskip

\hangindent=1.2em
\hangafter=1
-- closely follows the structure and reasoning style of the correct
solution\par

\hangindent=1.2em
\hangafter=1
-- introduces a subtle but critical mistake, such as a physics
misconception, mathematical error, or incorrect substitution, midway
through the steps\par

\hangindent=1.2em
\hangafter=1
-- clearly results in an INCORRECT final answer\par

\hangindent=1.2em
\hangafter=1
-- maintains the same level of detail and formatting as the correct
solution\par

\hangindent=1.2em
\hangafter=1
-- makes sure the LENGTH is similar to the accepted solution, with a
similar token count\par

\hangindent=1.2em
\hangafter=1
-- uses exactly the same XML tags as the correct solution:

\smallskip

\hspace*{1.2em}%
\textcolor{pastelPurpleFrame}{%
    \textbf{<problem\_analysis>}%
}\par

\hspace*{1.2em}%
\textcolor{pastelPurpleFrame}{%
    \textbf{<principle>}%
}\par

\hspace*{1.2em}%
\textcolor{pastelPurpleFrame}{%
    \textbf{<governing\_equation>}%
}\par

\hspace*{1.2em}%
\textcolor{pastelPurpleFrame}{%
    \textbf{<value\_identification>}%
}\par

\hspace*{1.2em}%
\textcolor{pastelPurpleFrame}{%
    \textbf{<substitution>}%
}\par

\hspace*{1.2em}%
\textcolor{pastelPurpleFrame}{%
    \textbf{<calculation>}%
}\par

\hspace*{1.2em}%
\textcolor{pastelPurpleFrame}{%
    \textbf{<final\_answer>}%
}

\medskip

DO NOT include any meta-commentary. Output ONLY the flawed solution.

\medskip

\textbf{Problem:}\par
\{question\}

\smallskip

\textbf{Correct Solution (for reference):}\par
\{solution\}

\medskip

Now generate ONLY the flawed (rejected) solution:

\end{tcolorbox}

\bigskip
\noindent\rule{\linewidth}{0.4pt}
\bigskip


\noindent
\colorbox{pastelPurpleFrame}{\small\textbf{\textcolor{white}{Pair 1 (ID: 2162)}}}
\quad
\colorbox{pastelPurple}{\small\textit{\textcolor{pastelPurpleFrame}{Thermodynamics \textemdash{} Adiabatic compression + isochoric heating}}}

\smallskip
\noindent\small\textit{Error injected at \texttt{<governing\_equation>}: $\gamma$ set to $7/5$ (diatomic) instead of $5/3$ (monatomic). Domino propagates through substitution, calculation, and final answer.}

\medskip

\noindent
\colorbox{pastelGreenFrame}{\small\textbf{\textcolor{white}{Chosen \textemdash{} Correct}}}

\medskip

\begin{tcolorbox}[
    enhanced, breakable,
    colback=pastelGreen,
    colframe=pastelGreenFrame,
    boxrule=0.6pt, arc=4pt,
    left=10pt, right=10pt, top=6pt, bottom=6pt,
]
\ttfamily\small
\textcolor{pastelGreenFrame}{\textbf{<problem\_analysis>}}\\
Two sequential processes: (1) adiabatic compression V1->V2,\\
(2) isochoric heating at V2 with heat Q.\\
Monatomic ideal gas, n=2 mol => gamma=5/3, Cv=3R/2.\\
\textcolor{pastelGreenFrame}{\textbf{</problem\_analysis>}}

\medskip
\textcolor{pastelGreenFrame}{\textbf{<principle>}}\\
First Law of Thermodynamics: DQ=DU+DW.\\
Adiabatic: DQ=0 => DU=-W. Isochoric: DW=0 => DU=DQ.\\
\textcolor{pastelGreenFrame}{\textbf{</principle>}}

\medskip
\textcolor{pastelGreenFrame}{\textbf{<governing\_equation>}}\\
Adiabatic: p1*V1\^{}gamma = p2*V2\^{}gamma \quad [gamma=5/3, monatomic]\\
W\_adiabatic = (p1*V1 - p2*V2)/(gamma-1)\\
\textcolor{pastelGreenFrame}{\textbf{</governing\_equation>}}

\medskip
\textcolor{pastelGreenFrame}{\textbf{<value\_identification>}}\\
n=2 mol, gamma=5/3, Cv=(3/2)R\\
Initial: p1, V1. After compression: V2. Heat added: Q.\\
\textcolor{pastelGreenFrame}{\textbf{</value\_identification>}}

\medskip
\textcolor{pastelGreenFrame}{\textbf{<substitution>}}\\
W1 = (3*p1*V1/2)*[1-(V1/V2)\^{}(2/3)] \quad [from gamma=5/3]\\
DU1=-W1=(3*p1*V1/2)*[(V1/V2)\^{}(2/3)-1], DU2=Q\\
\textcolor{pastelGreenFrame}{\textbf{</substitution>}}

\medskip
\textcolor{pastelGreenFrame}{\textbf{<calculation>}}\\
W  = (3*p1*V1/2)*[1-(V1/V2)\^{}(2/3)]\\
DU = (3*p1*V1/2)*[(V1/V2)\^{}(2/3)-1] + Q\\
T3 = Q/(3R) + (p1*V1)/(2R)*(V1/V2)\^{}(2/3)\\
\textcolor{pastelGreenFrame}{\textbf{</calculation>}}

\medskip
\textcolor{pastelGreenFrame}{\textbf{<final\_answer>}}\\
W=(3p1V1/2)[1-(V1/V2)\^{}(2/3)], DU=(3p1V1/2)[(V1/V2)\^{}(2/3)-1]+Q\\
T3=Q/(3R)+(p1V1/2R)*(V1/V2)\^{}(2/3)\\
\textcolor{pastelGreenFrame}{\textbf{</final\_answer>}}
\end{tcolorbox}

\medskip

\noindent
\colorbox{pastelRedFrame}{\small\textbf{\textcolor{white}{Rejected \textemdash{} Error at \texttt{<governing\_equation>}: $\gamma = 7/5$ used instead of $5/3$}}}

\medskip

\begin{tcolorbox}[
    enhanced, breakable,
    colback=pastelRed,
    colframe=pastelRedFrame,
    boxrule=0.6pt, arc=4pt,
    left=10pt, right=10pt, top=6pt, bottom=6pt,
]
\ttfamily\small
\textcolor{gray}{\textbf{<problem\_analysis>}} [Identical to chosen] \textcolor{gray}{\textbf{</problem\_analysis>}}\\
\textcolor{gray}{\textbf{<principle>}} [Identical to chosen] \textcolor{gray}{\textbf{</principle>}}

\medskip
\textcolor{pastelRedFrame}{\textbf{<governing\_equation>}}\\
Adiabatic: p1*V1\^{}gamma = p2*V2\^{}gamma \quad
\textcolor{pastelRedFrame}{\textbf{[ERROR: gamma=7/5, diatomic value]}}\\
W\_adiabatic = (p1*V1 - p2*V2)/(gamma-1)\\
\textcolor{pastelRedFrame}{\textbf{</governing\_equation>}}

\medskip
\textcolor{pastelRedFrame}{\textbf{<value\_identification>}}\\
n=2 mol, gamma=7/5, Cv=(5/2)R \quad
\textcolor{pastelRedFrame}{\textbf{[ERROR: wrong Cv for monatomic gas]}}\\
Initial: p1, V1. After compression: V2. Heat added: Q.\\
\textcolor{pastelRedFrame}{\textbf{</value\_identification>}}

\medskip
\textcolor{pastelRedFrame}{\textbf{<substitution>}}\\
W1 = (5*p1*V1/2)*[1-(V1/V2)\^{}(2/5)] \quad [wrong exponent from gamma=7/5]\\
DU1=-W1=(5*p1*V1/2)*[(V1/V2)\^{}(2/5)-1], DU2=Q\\
\textcolor{pastelRedFrame}{\textbf{</substitution>}}

\medskip
\textcolor{pastelRedFrame}{\textbf{<calculation>}}\\
W  = (5*p1*V1/2)*[1-(V1/V2)\^{}(2/5)] \quad [WRONG]\\
DU = (5*p1*V1/2)*[(V1/V2)\^{}(2/5)-1] + Q \quad [WRONG]\\
T3 = Q/(5R) + (p1*V1)/(2R)*(V1/V2)\^{}(2/5) \quad [WRONG]\\
\textcolor{pastelRedFrame}{\textbf{</calculation>}}

\medskip
\textcolor{pastelRedFrame}{\textbf{<final\_answer>}}\\
W=(5p1V1/2)[1-(V1/V2)\^{}(2/5)], DU=(5p1V1/2)[(V1/V2)\^{}(2/5)-1]+Q\\
T3=Q/(5R)+(p1V1/2R)*(V1/V2)\^{}(2/5) \quad [INCORRECT]\\
\textcolor{pastelRedFrame}{\textbf{</final\_answer>}}
\end{tcolorbox}

\bigskip
\noindent\rule{\linewidth}{0.4pt}
\bigskip


\noindent
\colorbox{pastelPurpleFrame}{\small\textbf{\textcolor{white}{Pair 2 (ID: 166)}}}
\quad
\colorbox{pastelPurple}{\small\textit{\textcolor{pastelPurpleFrame}{Electromagnetism \textemdash{} Electron in crossed $E$ and $B$ fields}}}

\smallskip
\noindent\small\textit{Error injected at \texttt{<substitution>}: force sign flipped --- electron treated as positive charge. Domino propagates through velocity, displacement, and final answer.}

\medskip

\noindent
\colorbox{pastelGreenFrame}{\small\textbf{\textcolor{white}{Chosen \textemdash{} Correct}}}

\medskip

\begin{tcolorbox}[
    enhanced, breakable,
    colback=pastelGreen,
    colframe=pastelGreenFrame,
    boxrule=0.6pt, arc=4pt,
    left=10pt, right=10pt, top=6pt, bottom=6pt,
]
\ttfamily\small
\textcolor{pastelGreenFrame}{\textbf{<problem\_analysis>}}\\
Electron (charge -e, mass m) from rest at origin.\\
E=-E*j, B=-B*k. Find y when vy=0 first time after t=0.\\
\textcolor{pastelGreenFrame}{\textbf{</problem\_analysis>}}

\medskip
\textcolor{pastelGreenFrame}{\textbf{<principle>}}\\
Lorentz force: F=q(E+v x B). Coupled ODEs reduce to SHM in vy.\\
\textcolor{pastelGreenFrame}{\textbf{</principle>}}

\medskip
\textcolor{pastelGreenFrame}{\textbf{<governing\_equation>}}\\
F=-e(E+v x B), omega=eB/m\\
vy(t)=(E/B)*sin(omega*t), y(t)=(E/B*omega)*(1-cos(omega*t))\\
\textcolor{pastelGreenFrame}{\textbf{</governing\_equation>}}

\medskip
\textcolor{pastelGreenFrame}{\textbf{<value\_identification>}}\\
E,B=field magnitudes, e=electron charge, m=electron mass\\
omega=eB/m, Initial: vx(0)=vy(0)=0\\
\textcolor{pastelGreenFrame}{\textbf{</value\_identification>}}

\medskip
\textcolor{pastelGreenFrame}{\textbf{<substitution>}}\\
F=-e(-E*j+v x (-B*k))=eE*j+eB*(vy*i-vx*j)\\
dvx/dt=(eB/m)*vy ...(1), dvy/dt=(e/m)*(E-vx*B) ...(2)\\
d2vy/dt2=-omega\^{}2*vy => vy=(E/B)*sin(omega*t)\\
\textcolor{pastelGreenFrame}{\textbf{</substitution>}}

\medskip
\textcolor{pastelGreenFrame}{\textbf{<calculation>}}\\
vy=0 at omega*t=pi => t*=pi*m/(eB)\\
y=(E/B*omega)*(1-cos(pi))=2Em/(eB\^{}2)\\
\textcolor{pastelGreenFrame}{\textbf{</calculation>}}

\medskip
\textcolor{pastelGreenFrame}{\textbf{<final\_answer>}}\\
y = 2Em/(eB\^{}2)\\
\textcolor{pastelGreenFrame}{\textbf{</final\_answer>}}
\end{tcolorbox}

\medskip

\noindent
\colorbox{pastelRedFrame}{\small\textbf{\textcolor{white}{Rejected \textemdash{} Error at \texttt{<substitution>}: charge sign flipped, $q = +e$}}}

\medskip

\begin{tcolorbox}[
    enhanced, breakable,
    colback=pastelRed,
    colframe=pastelRedFrame,
    boxrule=0.6pt, arc=4pt,
    left=10pt, right=10pt, top=6pt, bottom=6pt,
]
\ttfamily\small
\textcolor{gray}{\textbf{<problem\_analysis>}} [Identical to chosen] \textcolor{gray}{\textbf{</problem\_analysis>}}\\
\textcolor{gray}{\textbf{<principle>}} [Identical to chosen] \textcolor{gray}{\textbf{</principle>}}\\
\textcolor{gray}{\textbf{<governing\_equation>}} [Identical to chosen] \textcolor{gray}{\textbf{</governing\_equation>}}\\
\textcolor{gray}{\textbf{<value\_identification>}} [Identical to chosen] \textcolor{gray}{\textbf{</value\_identification>}}

\medskip
\textcolor{pastelRedFrame}{\textbf{<substitution>}}\\
F=+e(-E*j+v x (-B*k)) \quad
\textcolor{pastelRedFrame}{\textbf{[ERROR: sign flipped, q=+e used]}}\\
=-eE*j+eB*(...)\\
dvx/dt=-(eB/m)*vy ...(1), dvy/dt=(e/m)*(vx*B-E) ...(2)\\
d2vy/dt2=-omega\^{}2*vy => vy=-(E/B)*sin(omega*t) \quad [wrong sign]\\
\textcolor{pastelRedFrame}{\textbf{</substitution>}}

\medskip
\textcolor{pastelRedFrame}{\textbf{<calculation>}}\\
vy=0 at omega*t=pi => t*=pi*m/(eB) \quad [same time, wrong trajectory]\\
y=-(E/B*omega)*(1-cos(pi))=-2Em/(eB\^{}2) \quad [WRONG: negative displacement]\\
\textcolor{pastelRedFrame}{\textbf{</calculation>}}

\medskip
\textcolor{pastelRedFrame}{\textbf{<final\_answer>}}\\
y = -2Em/(eB\^{}2) \quad [INCORRECT: electron displaced in wrong direction]\\
\textcolor{pastelRedFrame}{\textbf{</final\_answer>}}
\end{tcolorbox}

\bigskip

\begin{tcolorbox}[
    enhanced,
    colback=pastelPurple,
    colframe=pastelPurpleFrame,
    boxrule=0.5pt, arc=3pt,
    left=8pt, right=8pt, top=5pt, bottom=5pt,
]
\small\normalfont
All DPO pairs follow this structure. The \texttt{<problem\_analysis>} and
\texttt{<principle>} tags are identical across chosen and rejected in every
pair. The error is introduced at exactly one tag and propagates consistently
through all subsequent tags, producing a fluent, structurally complete, but
physically incorrect rejected solution. This controlled design ensures the
preference signal targets reasoning quality rather than structural formatting.
\end{tcolorbox}

\bigskip
\noindent\rule{\linewidth}{0.4pt}
\bigskip


\begin{tcolorbox}[
    enhanced,
    colback=pastelPurple,
    colframe=pastelPurpleFrame,
    colbacktitle=pastelPurpleFrame,
    coltitle=white,
    fonttitle=\normalfont\bfseries\small,
    title={DPO Training Configuration \textemdash{} Shared (All Models)},
    boxrule=0.7pt, arc=4pt,
    left=10pt, right=10pt, top=8pt, bottom=8pt,
]
\ttfamily\small
\textcolor{pastelPurpleFrame}{\textbf{loss\_type}}              = ipo\\
\textcolor{pastelPurpleFrame}{\textbf{label\_smoothing}}        = 0.05\\
\textcolor{pastelPurpleFrame}{\textbf{max\_prompt\_length}}     = 256\\
\textcolor{pastelPurpleFrame}{\textbf{lr\_scheduler\_type}}     = cosine\\
\textcolor{pastelPurpleFrame}{\textbf{warmup\_ratio}}           = 0.1\\
\textcolor{pastelPurpleFrame}{\textbf{gradient\_checkpointing}} = True\\
\textcolor{pastelPurpleFrame}{\textbf{reference\_free}}         = False\\
\textcolor{pastelPurpleFrame}{\textbf{optimizer}}               = AdamW\\
\textcolor{pastelPurpleFrame}{\textbf{epochs}}                  = 1\\
\textcolor{pastelPurpleFrame}{\textbf{seed}}                    = 42
\end{tcolorbox}

\bigskip

\begin{tcolorbox}[
    enhanced,
    colback=pastelPurpleDeep,
    colframe=pastelPurpleFrame,
    colbacktitle=pastelPurpleFrame,
    coltitle=white,
    fonttitle=\normalfont\bfseries\small,
    title={Per-Model Hyperparameters},
    boxrule=0.7pt, arc=4pt,
    left=6pt, right=6pt, top=8pt, bottom=8pt,
]
\centering\small
\renewcommand{\arraystretch}{1.3}
\setlength{\tabcolsep}{6pt}
\begin{tabular}{lccccccc}
\toprule
\rowcolor{pastelPurpleFrame!80}
\textcolor{white}{\textbf{Model}} &
\textcolor{white}{\textbf{beta}} &
\textcolor{white}{\textbf{LR}} &
\textcolor{white}{\textbf{Batch}} &
\textcolor{white}{\textbf{Grad Accum}} &
\textcolor{white}{\textbf{Eff.\ Batch}} &
\textcolor{white}{\textbf{max\_grad\_norm}} &
\textcolor{white}{\textbf{max\_length}} \\
\midrule
\rowcolor{pastelPurple}
Qwen 2.5 1.5B     & 0.10 & 2e-6 & 2 & 16 & 32 & 1.0 & 4096 \\
\rowcolor{white}
LLaMA 3.2 1B      & 0.10 & 2e-6 & 2 & 16 & 32 & 1.0 & 4096 \\
\rowcolor{pastelPurple}
LLaMA 3.2 3B      & 0.10 & 2e-6 & 2 & 16 & 32 & 1.0 & 4096 \\
\rowcolor{white}
Phi 3.5 Mini 3.8B & 0.05 & 1e-6 & 1 & 16 & 16 & 0.5 & 4096 \\
\bottomrule
\end{tabular}

\vspace{0.5em}
\noindent\small\normalfont
\texttt{beta} is set lower for Phi 3.5 Mini to preserve pretrained physics
knowledge acquired during SFT. Learning rate decreases with model size as
larger models are more sensitive to weight updates post-LoRA. Each DPO
training run takes approximately 1.5--2 hours on a single H100 GPU.
Early stopping is applied based on validation loss to prevent over-optimisation.
\end{tcolorbox}

\twocolumn


\onecolumn

\begin{tcolorbox}[
    enhanced,
    colback=pastelOrange,
    colframe=pastelOrangeFrame,
    colbacktitle=pastelOrangeFrame,
    coltitle=white,
    fonttitle=\normalfont\bfseries\large,
    title={E \quad Retrieval-Augmented Generation (RAG) Setup},
    boxrule=0.8pt, arc=5pt,
    left=10pt, right=10pt, top=6pt, bottom=6pt,
]
\normalfont\small
RAG is used as a strong baseline that provides models with external physics
knowledge at inference time, without any parameter updates. The design
intentionally constrains what is retrievable to isolate the model's reasoning
ability from its recall ability — the system can retrieve the correct formula
but must still apply it correctly.
\end{tcolorbox}
\label{sec:ragconfig}
\bigskip

\begin{figure}[h!]
    \centering
    \includegraphics[width=0.95\textwidth, page=1]{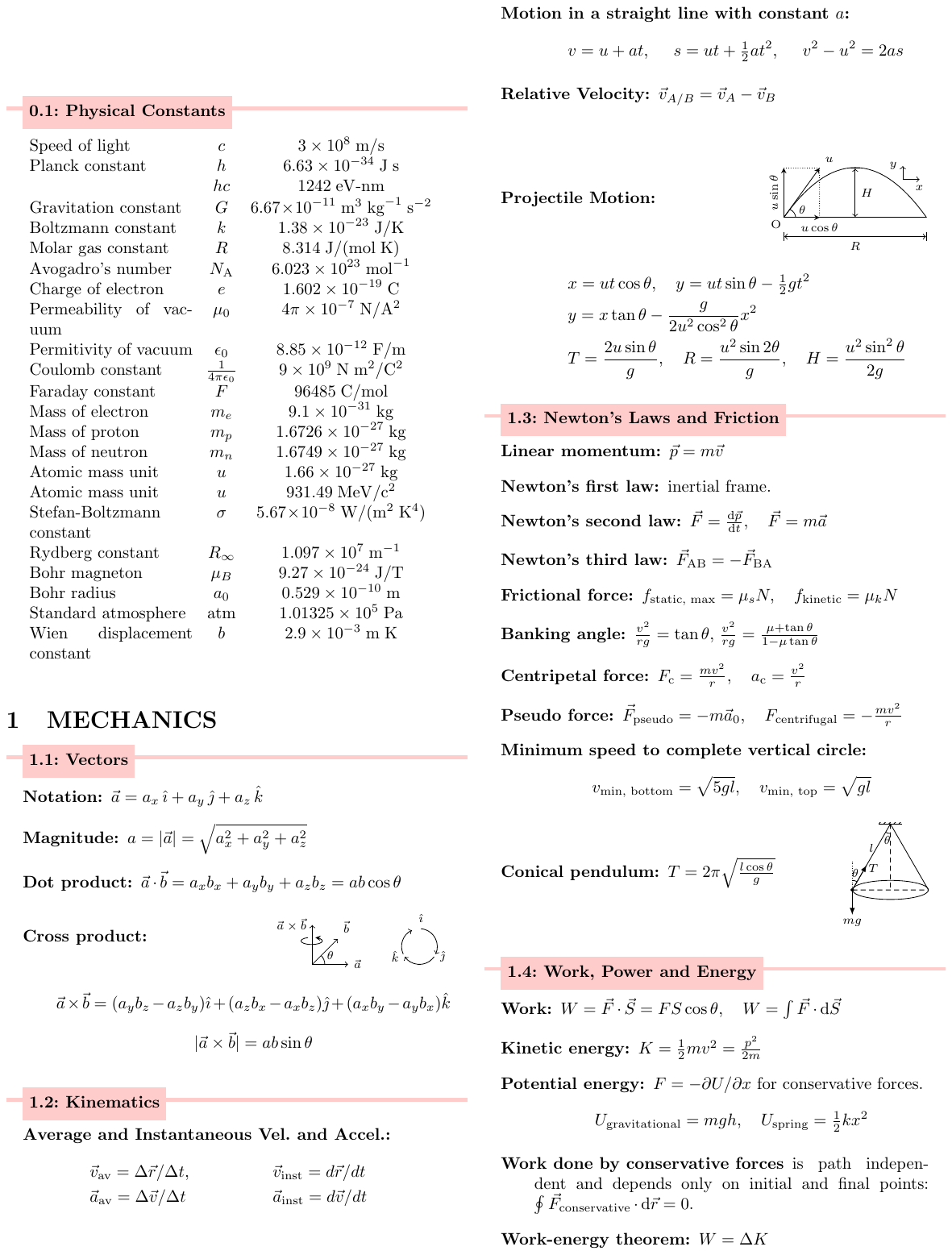}
    \caption{The physics formula sheet used as the RAG knowledge base.
             The corpus is strictly limited to formulas, definitions, and
             physical constants. It explicitly lacks narrative explanations,
             problem-solving strategies, or worked examples, ensuring the
             RAG system acts as a formula reference rather than a solution
             retriever.}
    \label{fig:rag_formula_sheet_p1}
\end{figure}

\begin{tcolorbox}[
    enhanced,
    colback=pastelOrange,
    colframe=pastelOrangeFrame,
    colbacktitle=pastelOrangeFrame,
    coltitle=white,
    fonttitle=\normalfont\bfseries\small,
    title={Knowledge Base},
    boxrule=0.7pt, arc=4pt,
    left=10pt, right=10pt, top=8pt, bottom=8pt,
]
\normalfont\small
The retrieval corpus is a concise physics formula sheet covering six domains:
Mechanics, Waves, Optics, Heat and Thermodynamics, Electricity and Magnetism,
and Modern Physics, together with a table of standard physical constants. The
corpus contains \textit{only} symbolic formulas, definitions, and physical
constants --- it explicitly excludes narrative explanations, worked examples,
and problem-solving strategies. This design ensures the model cannot locate and
reproduce a pre-existing solution. The model must still perform all critical
reasoning steps: identifying the relevant formula from context, performing
correct substitution, and executing the calculation. The full formula sheet
used as the RAG knowledge base is shown in
Figure~\ref{fig:rag_formula_sheet_p1}.
\end{tcolorbox}

\bigskip

\begin{tcolorbox}[
    enhanced,
    colback=pastelOrange,
    colframe=pastelOrangeFrame,
    colbacktitle=pastelOrangeFrame,
    coltitle=white,
    fonttitle=\normalfont\bfseries\small,
    title={Chunking and Embedding Configuration},
    boxrule=0.7pt, arc=4pt,
    left=10pt, right=10pt, top=8pt, bottom=8pt,
]
\normalfont\small
The formula sheet is chunked into segments of 500 characters with a
50-character overlap to preserve context across formula boundaries. Chunks
are embedded using OpenAI's \texttt{text-embedding-ada-002} model and stored
in a vector store built with LangChain.

\medskip

\ttfamily\small
\textcolor{pastelOrangeFrame}{\textbf{Chunking strategy}} \;=\; 500-character segments, 50-character overlap\\
\textcolor{pastelOrangeFrame}{\textbf{Embedding model}}   \;=\; text-embedding-ada-002 (OpenAI)\\
\textcolor{pastelOrangeFrame}{\textbf{Vector store}}      \;=\; LangChain FAISS\\
\textcolor{pastelOrangeFrame}{\textbf{Retrieval}}         \;=\; top-$k$ = 3 most similar chunks\\
\textcolor{pastelOrangeFrame}{\textbf{Similarity metric}} \;=\; cosine similarity
\end{tcolorbox}

\bigskip

\begin{tcolorbox}[
    enhanced,
    breakable,
    colback=pastelOrange,
    colframe=pastelOrangeFrame,
    colbacktitle=pastelOrangeFrame,
    coltitle=white,
    fonttitle=\normalfont\bfseries\small,
    title={Inference Prompt \textemdash{} RAG Instruction Template},
    boxrule=0.7pt,
    arc=4pt,
    left=10pt,
    right=10pt,
    top=8pt,
    bottom=8pt,
]

\ttfamily\small\raggedright
\setlength{\parindent}{0pt}
\setlength{\parskip}{2pt}

You are an expert physics assistant. Use the provided context to solve
the physics problem below. Think step by step.

\medskip

\textcolor{pastelOrangeFrame}{%
    \textbf{Context (retrieved physics formulas):}%
}\par
\{retrieved\_chunks\}

\medskip

\textcolor{pastelOrangeFrame}{\textbf{Question:}}\par
\{question\}

\medskip

Using the context above, provide a complete step-by-step solution.
Your final answer should be clearly stated at the end.

\end{tcolorbox}

\bigskip

\begin{tcolorbox}[
    enhanced,
    colback=pastelOrangeDeep,
    colframe=pastelOrangeFrame,
    boxrule=0.5pt, arc=3pt,
    left=8pt, right=8pt, top=5pt, bottom=5pt,
]
\small\normalfont
The RAG baseline receives identical prompting structure to all other conditions.
The only difference is the injected context block. No parameter updates are
performed during RAG inference --- the knowledge base is static across all
models and benchmarks.
\end{tcolorbox}

%
%

\begin{figure}[h!]
    \centering
    \includegraphics[width=1.0\textwidth, page=2]{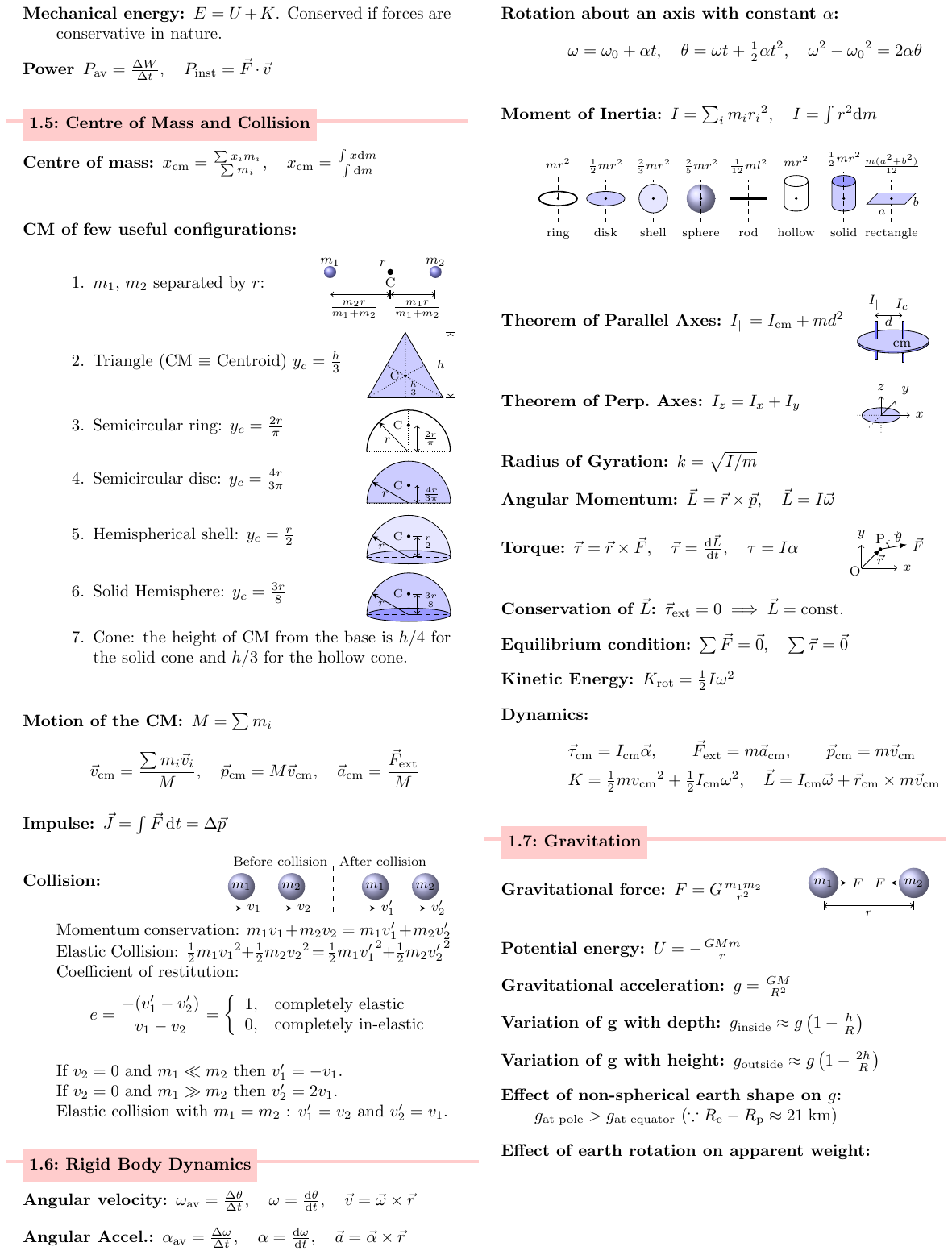}
    \caption*{Figure~\ref{fig:rag_formula_sheet_p1} continued (Page 2).}
\end{figure}


\begin{figure}[h!]
    \centering
    \includegraphics[width=1.0\textwidth, page=4]{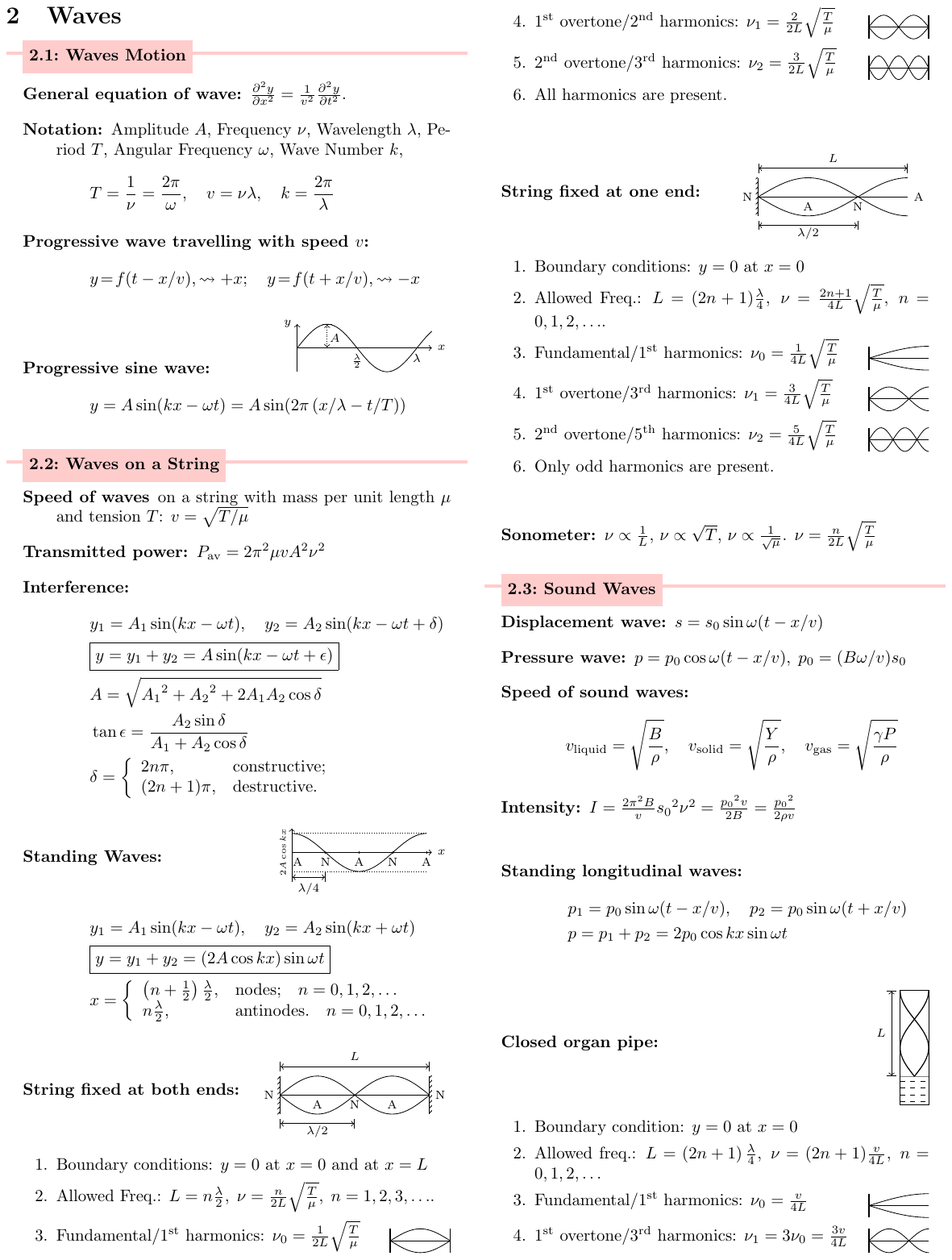}
    \caption*{Figure~\ref{fig:rag_formula_sheet_p1} continued (Page 4).}
\end{figure}


\begin{figure}[h!]
    \centering
    \includegraphics[width=1.0\textwidth, page=6]{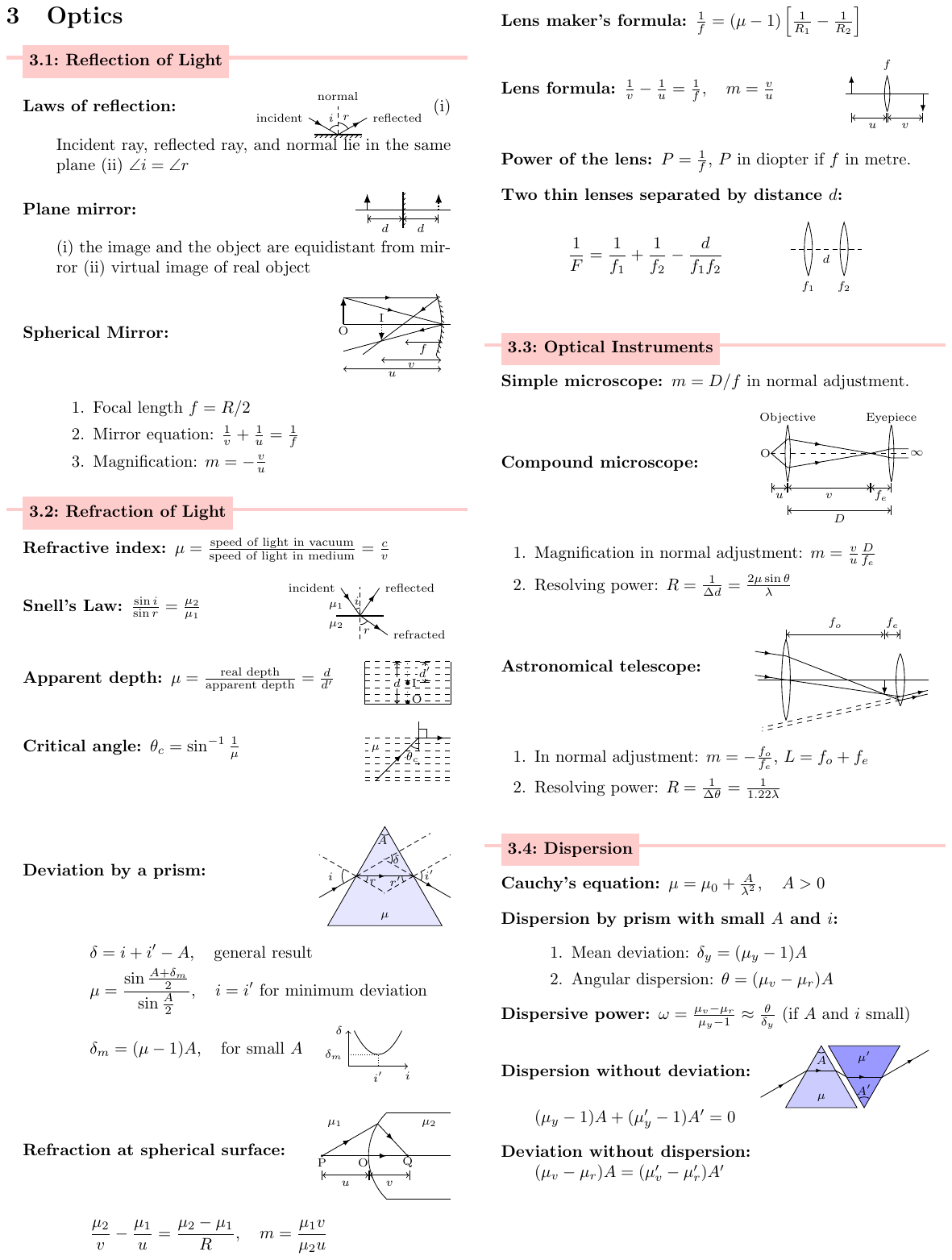}
    \caption*{Figure~\ref{fig:rag_formula_sheet_p1} continued (Page 6).}
\end{figure}

\begin{figure}[h!]
    \centering
    \includegraphics[width=1.0\textwidth, page=7]{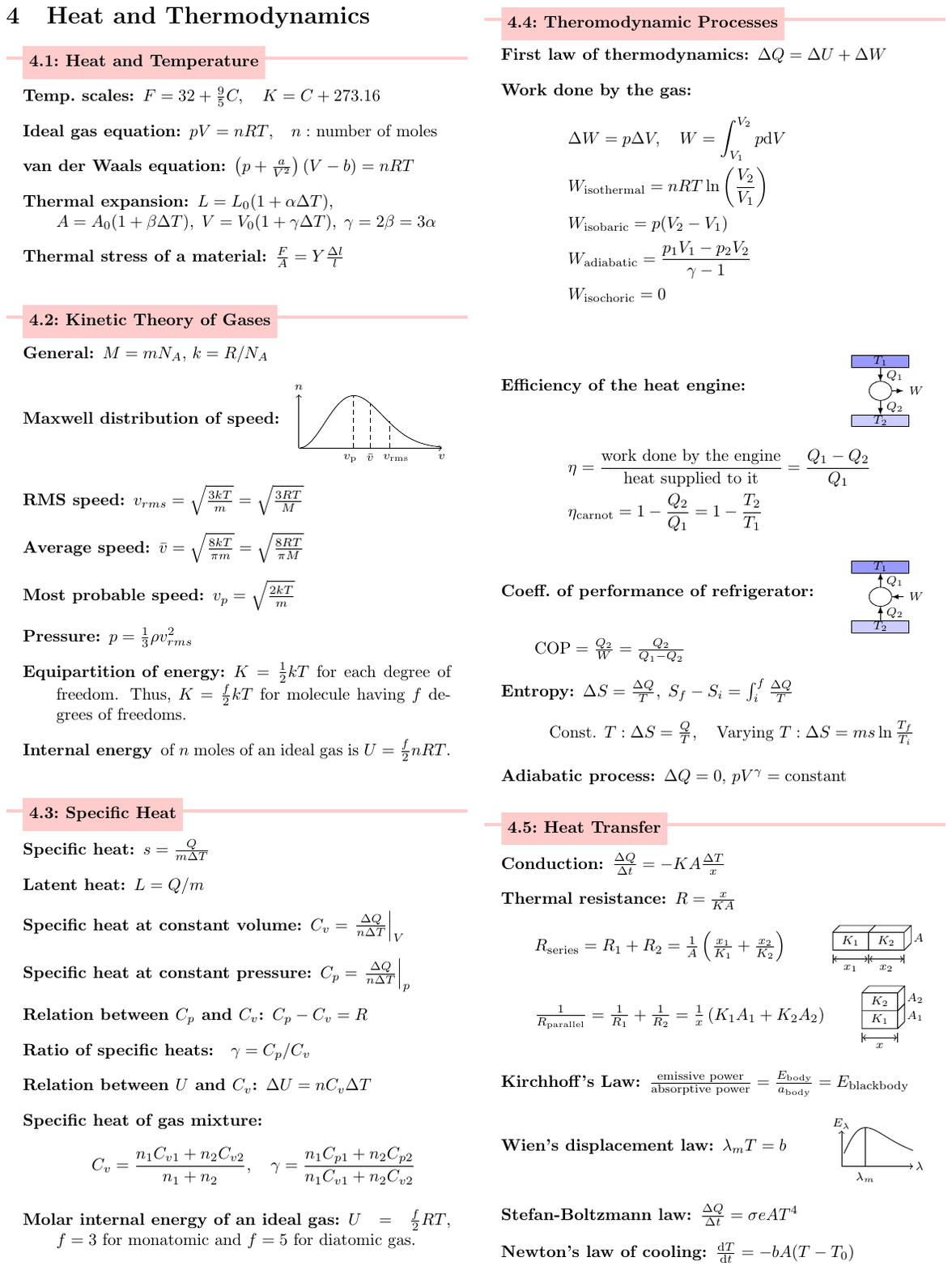}
    \caption*{Figure~\ref{fig:rag_formula_sheet_p1} continued (Page 7).}
\end{figure}

\begin{figure}[h!]
    \centering
    \includegraphics[width=1.0\textwidth, page=8]{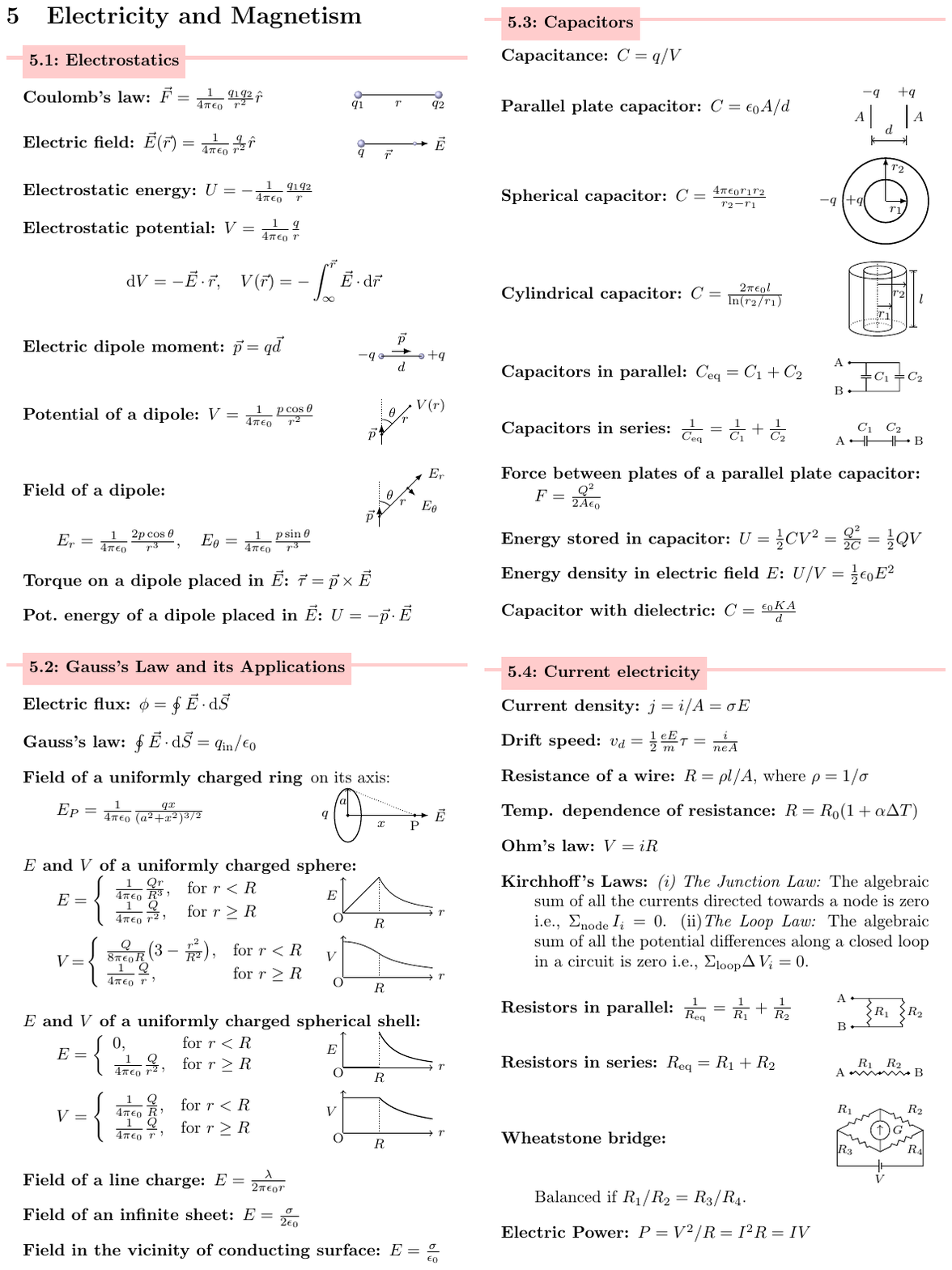}
    \caption*{Figure~\ref{fig:rag_formula_sheet_p1} continued (Page 8).}
\end{figure}



\begin{figure}[h!]
    \centering
    \includegraphics[width=1.0\textwidth, page=11]{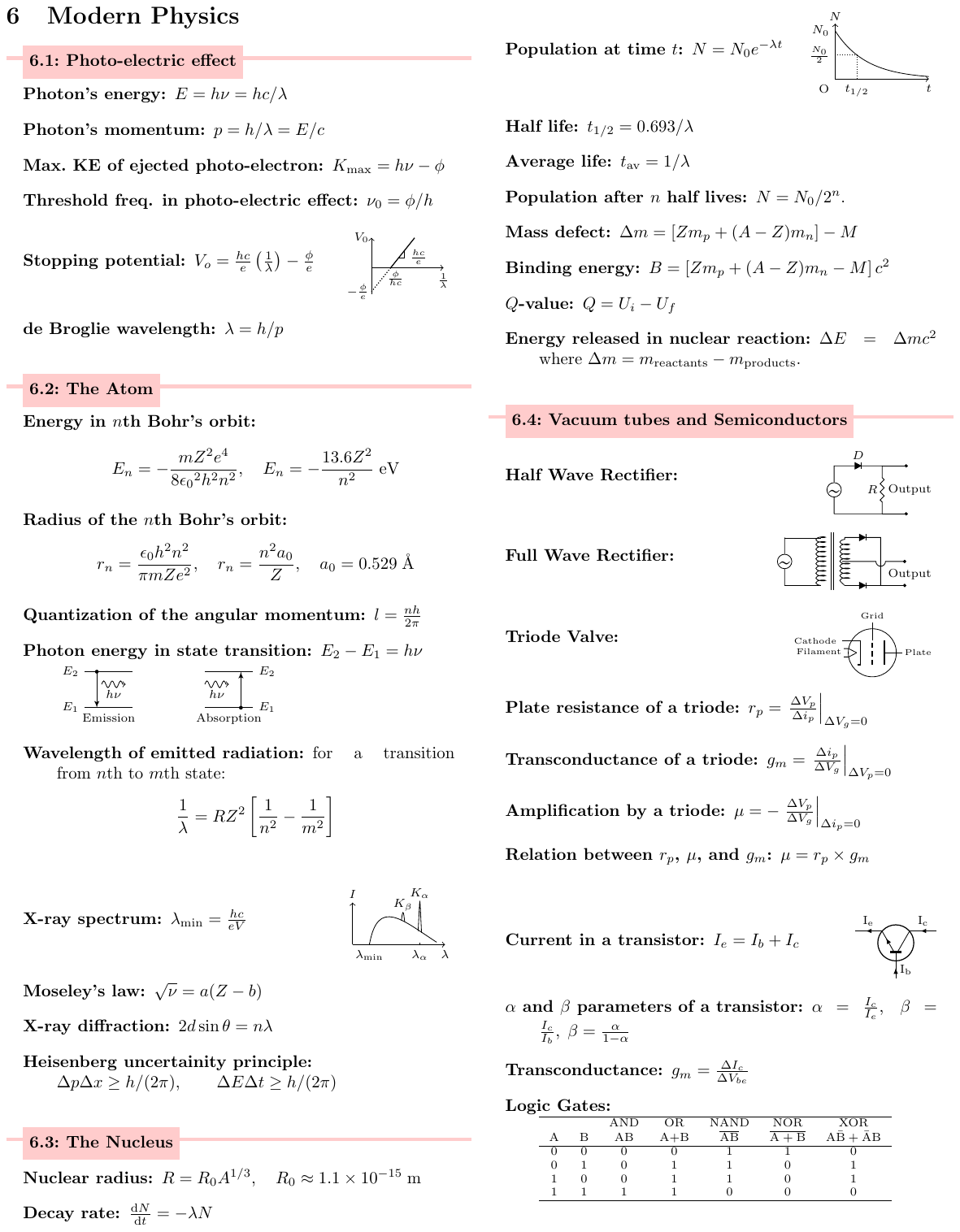}
    \caption*{Figure~\ref{fig:rag_formula_sheet_p1} continued (Page 11).}
\end{figure}

%
\twocolumn


\onecolumn

\begin{tcolorbox}[
    enhanced,
    colback=pastelBlue,
    colframe=pastelBlueFrame,
    colbacktitle=pastelBlueFrame,
    coltitle=white,
    fonttitle=\normalfont\bfseries\large,
    title={F \quad Evaluation Pipeline},
    boxrule=0.8pt, arc=5pt,
    left=10pt, right=10pt, top=6pt, bottom=6pt,
]
\normalfont\small
We evaluate model outputs using a three-step shadow evaluation protocol
that combines automated verification with LLM-based error analysis. Each
sample produces a shadow list $[S_1, S_2, S_3]$ where each entry is
\texttt{No} (correct / no error) or \texttt{Yes} (incorrect / error
present).
\end{tcolorbox}
\label{sec:evalpipeline}
\bigskip

\begin{tcolorbox}[
    enhanced,
    colback=pastelBlue,
    colframe=pastelBlueFrame,
    colbacktitle=pastelBlueFrame,
    coltitle=white,
    fonttitle=\normalfont\bfseries\small,
    title={Shadow List Classification Logic},
    boxrule=0.7pt, arc=4pt,
    left=6pt, right=6pt, top=8pt, bottom=8pt,
]
\centering\small
\renewcommand{\arraystretch}{1.4}
\setlength{\tabcolsep}{8pt}
\begin{tabular}{ccc p{7cm}}
\toprule
\rowcolor{pastelBlueFrame!85}
\textcolor{white}{\textbf{S1 (Answer)}} &
\textcolor{white}{\textbf{S2 (Tags)}} &
\textcolor{white}{\textbf{S3 (Errors)}} &
\textcolor{white}{\textbf{Classification}} \\
\midrule
\rowcolor{pastelGreen}
No  & No  & No  & \textbf{True Positive} --- fully correct \\
\rowcolor{pastelRed}
Yes & Yes & Yes & \textbf{True Negative} --- all errors present \\
\rowcolor{pastelRed}
Yes & No  & Yes & \textbf{True Negative} --- needs human verification \\
\rowcolor{pastelBlue}
Yes & No  & No  & \textbf{Ambiguous} --- wrong answer, no detectable error \\
\bottomrule
\end{tabular}

\vspace{0.4em}
\noindent\small\normalfont
S1 = answer correctness, \; S2 = tag compliance, \; S3 = step-level error presence.
\end{tcolorbox}

\bigskip
\noindent\rule{\linewidth}{0.3pt}
\bigskip

\begin{tcolorbox}[
    enhanced,
    colback=pastelBlue,
    colframe=pastelBlueFrame,
    colbacktitle=pastelBlueFrame,
    coltitle=white,
    fonttitle=\normalfont\bfseries\small,
    title={F.1 \quad Three-Step Protocol},
    boxrule=0.7pt, arc=4pt,
    left=10pt, right=10pt, top=8pt, bottom=8pt,
]
\normalfont\small

\textbf{Step 1 --- Answer Verification (automated).}
The \texttt{\textbackslash boxed\{\}} content is extracted from the
\texttt{<final\_answer>} tag and normalised (whitespace, \LaTeX{} formatting,
case). Returns \texttt{No} (correct) or \texttt{Yes} (wrong / no boxed answer
found). No LLM is involved.

\medskip
\textbf{Step 2 --- Tag Compliance (automated).}
All seven required XML tags are checked for presence and closure:
\texttt{<problem\_analysis>}, \texttt{<principle>},
\texttt{<governing\_equation>}, \texttt{<value\_identification>},
\texttt{<substitution>}, \texttt{<calculation>}, \texttt{<final\_answer>}.
Returns \texttt{No} (all present) or \texttt{Yes} (any tag missing).
Step 3 is skipped if Step 2 returns \texttt{Yes}.

\medskip
\textbf{Step 3 --- Error Analysis (LLM-based).}
An LLM judge evaluates the model's solution across three independent
dimensions: Problem Miscomprehension (MC), Conceptual Misapplication (CM),
and Calculation Error (CE). 
\end{tcolorbox}

\bigskip

\begin{tcolorbox}[
    enhanced,
    breakable,
    colback=pastelTeal,
    colframe=pastelTealFrame,
    colbacktitle=pastelTealFrame,
    coltitle=white,
    fonttitle=\normalfont\bfseries\small,
    title={F.2 \quad Step 3 Error Analysis Prompt},
    boxrule=0.7pt,
    arc=4pt,
    left=10pt,
    right=10pt,
    top=8pt,
    bottom=8pt,
]

\ttfamily\small\raggedright
\setlength{\parindent}{0pt}
\setlength{\parskip}{2pt}

\textbf{Question:} \{question\}\par
\textbf{Generated Solution:} \{solution\}\par
\textbf{Original Answer:} \{answer\}\par
\textbf{Problem ID:} \{problem\_id\}

\medskip

\textcolor{pastelTealFrame}{%
    \textbf{1. Problem Comprehension:}%
}\par
1. Does the solution attempt to address the correct objective asked in
the question?\par
2. Are the correct values, variables, and notations from the question
being used?

\medskip

\textcolor{pastelTealFrame}{%
    \textbf{2. Concept Application:}%
}\par
1. Check the solution against the relevant physics concepts and formulas
required.\par
2. Verify whether the correct physics concepts and formulas are applied.

\smallskip

\textbf{Important: Do not verify mathematical reasoning or
calculations.}\par
\textbf{Focus only on whether the correct physics concepts are applied.}

\medskip

\textcolor{pastelTealFrame}{%
    \textbf{3. Calculation Accuracy:}%
}\par
Check each step and verify all mathematical calculations, including
arithmetic, algebraic manipulation, substitutions, integration,
differentiation, fractions, exponents, and numerical approximations.

\medskip

\textcolor{pastelTealFrame}{\textbf{Output format}}
(strictly follow; no extra text):\par
\textbf{Problem Miscomprehension Flag: [Yes/No]}\par
\textbf{Concept Error Flag: [Yes/No]}\par
\textbf{Calculation Error Flag: [Yes/No]}\par
\textbf{Judgement Key: [Correct/Incorrect]}

\end{tcolorbox}

\bigskip
\noindent\rule{\linewidth}{0.3pt}
\bigskip


\begin{tcolorbox}[
    enhanced,
    colback=pastelBlue,
    colframe=pastelBlueFrame,
    colbacktitle=pastelBlueFrame,
    coltitle=white,
    fonttitle=\normalfont\bfseries\small,
    title={F.3 \quad Erroneous Solution Examples},
    boxrule=0.7pt, arc=4pt,
    left=10pt, right=10pt, top=5pt, bottom=5pt,
]
\normalfont\small
Three samples from PhysicsQA, one per error type. The erroneous tag is
highlighted in each rejected solution to show exactly what the Step 3
judge detects.
\end{tcolorbox}

\bigskip

\noindent
\colorbox{pastelBlueFrame}{\small\textbf{\textcolor{white}{Example 1}}}
\quad
\colorbox{pastelBlue}{\small\textit{\textcolor{pastelBlueFrame}{Problem Miscomprehension (MC) \textemdash{} model swaps given quantity}}}

\smallskip
\noindent\small
A gas can be taken from A to B via two processes ACB and ADB.
Path ACB: $Q=60~\mathrm{J}$, $W=30~\mathrm{J}$. Path ADB: $W=10~\mathrm{J}$.
Find heat flow in path ADB. \quad \textbf{Ground Truth: (c) 40 J}

\medskip
\noindent\colorbox{pastelGreenFrame}{\small\textbf{\textcolor{white}{Correct Solution}}}

\medskip

\begin{tcolorbox}[enhanced, breakable, colback=pastelGreen, colframe=pastelGreenFrame,
    boxrule=0.6pt, arc=3pt, left=10pt, right=10pt, top=6pt, bottom=6pt]
\ttfamily\small
\textcolor{pastelGreenFrame}{\textbf{<value\_identification>}}
$Q_\text{ACB}=60$ J, $W_\text{ACB}=30$ J; $W_\text{ADB}=10$ J, $Q_\text{ADB}=?$
\textcolor{pastelGreenFrame}{\textbf{</value\_identification>}}

\smallskip
\textcolor{pastelGreenFrame}{\textbf{<substitution>}}
$\Delta U=60-30=30$ J; \quad $30=Q_\text{ADB}-10$
\textcolor{pastelGreenFrame}{\textbf{</substitution>}}

\smallskip
\textcolor{pastelGreenFrame}{\textbf{<calculation>}}
$Q_\text{ADB}=40$ J
\textcolor{pastelGreenFrame}{\textbf{</calculation>}}

\smallskip
\textcolor{pastelGreenFrame}{\textbf{<final\_answer>}}
$\boxed{(c)~40~\mathrm{J}}$
\textcolor{pastelGreenFrame}{\textbf{</final\_answer>}}
\end{tcolorbox}

\medskip
\noindent\colorbox{pastelRedFrame}{\small\textbf{\textcolor{white}{Erroneous \textemdash{} MC: $W=10$ J misread as $Q=10$ J}}}

\medskip

\begin{tcolorbox}[enhanced, breakable, colback=pastelRed, colframe=pastelRedFrame,
    boxrule=0.6pt, arc=3pt, left=10pt, right=10pt, top=6pt, bottom=6pt]
\ttfamily\small
\textcolor{pastelRedFrame}{\textbf{<value\_identification>}}
\textcolor{pastelRedFrame}{\textbf{Path ADB: Q=10 J [ERROR: W=10 J misread as Q=10 J]}}
\textcolor{pastelRedFrame}{\textbf{</value\_identification>}}

\smallskip
\textcolor{pastelRedFrame}{\textbf{<substitution>}}
$\Delta U=30$ J; \quad \textcolor{pastelRedFrame}{\textbf{$30=10-W_\text{ADB}$}}
\textcolor{pastelRedFrame}{\textbf{</substitution>}}

\smallskip
\textcolor{pastelRedFrame}{\textbf{<final\_answer>}}
\textcolor{pastelRedFrame}{\textbf{$\boxed{(d)~20~\mathrm{J}}$ [INCORRECT]}}
\textcolor{pastelRedFrame}{\textbf{</final\_answer>}}
\end{tcolorbox}

\medskip
\begin{tcolorbox}[enhanced, colback=pastelBlueDeep, colframe=pastelBlueFrame,
    boxrule=0.4pt, arc=2pt, left=8pt, right=8pt, top=4pt, bottom=4pt]
\ttfamily\small
\textbf{Problem Miscomprehension Flag: \textcolor{pastelRedFrame}{Yes}} \quad
Concept Error Flag: No \quad Calculation Error Flag: No \quad
Judgement Key: Incorrect
\end{tcolorbox}

\bigskip
\noindent\rule{\linewidth}{0.3pt}
\bigskip

\noindent
\colorbox{pastelPurpleFrame}{\small\textbf{\textcolor{white}{Example 2}}}
\quad
\colorbox{pastelPurple}{\small\textit{\textcolor{pastelPurpleFrame}{Conceptual Misapplication (CM) \textemdash{} monatomic $C_p$ applied to diatomic gas}}}

\smallskip
\noindent\small
Diatomic rigid gas does $W=10~\mathrm{J}$ at constant pressure. Find heat absorbed.
\quad \textbf{Ground Truth: (c) 35 J}

\medskip
\noindent\colorbox{pastelGreenFrame}{\small\textbf{\textcolor{white}{Correct Solution}}}

\medskip

\begin{tcolorbox}[enhanced, breakable, colback=pastelGreen, colframe=pastelGreenFrame,
    boxrule=0.6pt, arc=3pt, left=10pt, right=10pt, top=6pt, bottom=6pt]
\ttfamily\small
\textcolor{pastelGreenFrame}{\textbf{<governing\_equation>}}
rigid diatomic: $C_v=5R/2$, $C_p=7R/2$; \quad $\Delta Q/\Delta W=C_p/R=7/2$
\textcolor{pastelGreenFrame}{\textbf{</governing\_equation>}}

\smallskip
\textcolor{pastelGreenFrame}{\textbf{<substitution>}}
$\Delta Q = 10 \times (7/2) = 35$ J
\textcolor{pastelGreenFrame}{\textbf{</substitution>}}

\smallskip
\textcolor{pastelGreenFrame}{\textbf{<final\_answer>}}
$\boxed{(c)~35~\mathrm{J}}$
\textcolor{pastelGreenFrame}{\textbf{</final\_answer>}}
\end{tcolorbox}

\medskip
\noindent\colorbox{pastelRedFrame}{\small\textbf{\textcolor{white}{Erroneous \textemdash{} CM: monatomic $C_p=5R/2$ used instead of diatomic $7R/2$}}}

\medskip

\begin{tcolorbox}[enhanced, breakable, colback=pastelRed, colframe=pastelRedFrame,
    boxrule=0.6pt, arc=3pt, left=10pt, right=10pt, top=6pt, bottom=6pt]
\ttfamily\small
\textcolor{pastelRedFrame}{\textbf{<governing\_equation>}}
\textcolor{pastelRedFrame}{\textbf{$C_v=3R/2$, $C_p=5R/2$ [ERROR: monatomic values for diatomic gas]}}
\textcolor{pastelRedFrame}{\textbf{</governing\_equation>}}

\smallskip
\textcolor{pastelRedFrame}{\textbf{<substitution>}}
\textcolor{pastelRedFrame}{\textbf{$\Delta Q=10\times(5/2)=25$ J [WRONG]}}
\textcolor{pastelRedFrame}{\textbf{</substitution>}}

\smallskip
\textcolor{pastelRedFrame}{\textbf{<final\_answer>}}
\textcolor{pastelRedFrame}{\textbf{$\boxed{(d)~20~\mathrm{J}}$ [INCORRECT]}}
\textcolor{pastelRedFrame}{\textbf{</final\_answer>}}
\end{tcolorbox}

\medskip
\begin{tcolorbox}[enhanced, colback=pastelBlueDeep, colframe=pastelBlueFrame,
    boxrule=0.4pt, arc=2pt, left=8pt, right=8pt, top=4pt, bottom=4pt]
\ttfamily\small
Problem Miscomprehension Flag: No \quad
\textbf{Concept Error Flag: \textcolor{pastelRedFrame}{Yes}} \quad
Calculation Error Flag: No \quad Judgement Key: Incorrect
\end{tcolorbox}

\bigskip
\noindent\rule{\linewidth}{0.3pt}
\bigskip

\noindent
\colorbox{pastelTealFrame}{\small\textbf{\textcolor{white}{Example 3}}}
\quad
\colorbox{pastelTeal}{\small\textit{\textcolor{pastelTealFrame}{Calculation Error (CE) \textemdash{} correct concept, arithmetic error in final multiplication}}}

\smallskip
\noindent\small
Helium in fixed $67.2~\mathrm{L}$ cylinder at STP. Heat for $\Delta T=20~\mathrm{K}$?
[$R=8.31~\mathrm{J\,mol^{-1}\,K^{-1}}$]
\quad \textbf{Ground Truth: (c) 748 J}

\medskip
\noindent\colorbox{pastelGreenFrame}{\small\textbf{\textcolor{white}{Correct Solution}}}

\medskip

\begin{tcolorbox}[enhanced, breakable, colback=pastelGreen, colframe=pastelGreenFrame,
    boxrule=0.6pt, arc=3pt, left=10pt, right=10pt, top=6pt, bottom=6pt]
\ttfamily\small
\textcolor{pastelGreenFrame}{\textbf{<calculation>}}
$\Delta Q = 3 \times 1.5 \times 8.31 \times 20 = 747.9 \approx 748$ J
\textcolor{pastelGreenFrame}{\textbf{</calculation>}}

\smallskip
\textcolor{pastelGreenFrame}{\textbf{<final\_answer>}}
$\boxed{(c)~748~\mathrm{J}}$
\textcolor{pastelGreenFrame}{\textbf{</final\_answer>}}
\end{tcolorbox}

\medskip
\noindent\colorbox{pastelRedFrame}{\small\textbf{\textcolor{white}{Erroneous \textemdash{} CE: dropped factor of 2 in multiplication}}}

\medskip

\begin{tcolorbox}[enhanced, breakable, colback=pastelRed, colframe=pastelRedFrame,
    boxrule=0.6pt, arc=3pt, left=10pt, right=10pt, top=6pt, bottom=6pt]

\ttfamily\small\raggedright
\textcolor{gray}{\textbf{<problem\_analysis>}} [Identical]
\textcolor{gray}{\textbf{</problem\_analysis>}}\\
\textcolor{gray}{\textbf{<principle>}} [Identical]
\textcolor{gray}{\textbf{</principle>}}\\
\textcolor{gray}{\textbf{<governing\_equation>}} [Identical]
\textcolor{gray}{\textbf{</governing\_equation>}}\\
\textcolor{gray}{\textbf{<value\_identification>}} [Identical]
\textcolor{gray}{\textbf{</value\_identification>}}\\
\textcolor{gray}{\textbf{<substitution>}} [Identical]
\textcolor{gray}{\textbf{</substitution>}}
\smallskip

\textcolor{pastelRedFrame}{\textbf{<calculation>}}
$= 4.5 \times 8.31 \times 20$\\
\textcolor{pastelRedFrame}{\textbf{$= 37.395 \times 20 = 374$ J [WRONG --- dropped factor of 2]}}
\textcolor{pastelRedFrame}{\textbf{</calculation>}}

\smallskip
\textcolor{pastelRedFrame}{\textbf{<final\_answer>}}
\textcolor{pastelRedFrame}{\textbf{$\boxed{(d)~374~\mathrm{J}}$ [INCORRECT]}}
\textcolor{pastelRedFrame}{\textbf{</final\_answer>}}
\end{tcolorbox}

\medskip
\begin{tcolorbox}[enhanced, colback=pastelBlueDeep, colframe=pastelBlueFrame,
    boxrule=0.4pt, arc=2pt, left=8pt, right=8pt, top=4pt, bottom=4pt]
\ttfamily\small
Problem Miscomprehension Flag: No \quad Concept Error Flag: No\\
\textbf{Calculation Error Flag: \textcolor{pastelRedFrame}{Yes}} \quad
Judgement Key: Incorrect
\end{tcolorbox}

\bigskip

\begin{tcolorbox}[
    enhanced,
    colback=pastelBlueDeep,
    colframe=pastelBlueFrame,
    boxrule=0.5pt, arc=3pt,
    left=8pt, right=8pt, top=5pt, bottom=5pt,
]
\small\normalfont
All three examples demonstrate the independence of the three error flags.
The Step 3 judge evaluates each dimension separately --- a solution can
fail on MC without CM or CE, and vice versa. This independence enables
the error distribution analysis to attribute performance gaps to specific
failure modes rather than conflating them into a single accuracy metric.
\end{tcolorbox}

\twocolumn


\onecolumn

\begin{tcolorbox}[
    enhanced,
    colback=pastelBlue,
    colframe=pastelBlueFrame,
    colbacktitle=pastelBlueFrame,
    coltitle=white,
    fonttitle=\normalfont\bfseries\large,
    title={D \quad Benchmark Dataset Samples},
    boxrule=0.8pt, arc=5pt,
    left=10pt, right=10pt, top=6pt, bottom=6pt,
]
\normalfont\small
We evaluate on five benchmarks spanning foundational to advanced physics
reasoning. Two to three representative samples per benchmark are shown below,
reproduced in full. All samples are drawn directly from the evaluation sets
used in our experiments.
\end{tcolorbox}
\label{sec:benchmarkss}
\bigskip


\begin{tcolorbox}[
    enhanced,
    colback=pastelGreen,
    colframe=pastelGreenFrame,
    colbacktitle=pastelGreenFrame,
    coltitle=white,
    fonttitle=\normalfont\bfseries\normalsize,
    title={D.1 \quad PhysicsQA \quad
           \normalfont\small $N = 370$ $\mid$
           JEE-sourced MCQ with CoT solutions $\mid$
           Intermediate difficulty},
    boxrule=0.7pt, arc=4pt,
    left=8pt, right=8pt, top=5pt, bottom=5pt,
]
\normalfont\small
PhysicsQA comprises 370 intermediate-level high school physics problems sourced
from Indian JEE preparation materials (2000--2010). Each problem is a four-option
MCQ accompanied by a verified chain-of-thought solution, enabling step-level
evaluation beyond final answer accuracy.
\end{tcolorbox}

\medskip

\noindent
\colorbox{pastelGreenFrame}{\small\textbf{\textcolor{white}{Sample 1}}}
\quad
\colorbox{pastelGreen}{\small\textit{\textcolor{pastelGreenFrame}{Heat Transfer \textemdash{} Path-independent internal energy, First Law}}}

\medskip

\begin{tcolorbox}[
    enhanced, breakable,
    colback=pastelGreen,
    colframe=pastelGreenFrame,
    boxrule=0.6pt, arc=3pt,
    left=10pt, right=10pt, top=6pt, bottom=6pt,
]
\small\normalfont
A gas can be taken from A to B via two different processes ACB and ADB. When
path ACB is used, $60~\mathrm{J}$ of heat flows into the system and
$30~\mathrm{J}$ of work is done by the system. If path ADB is used, work done
by the system is $10~\mathrm{J}$. The heat flow into the system in path ADB is:

\medskip
\textbf{Options:} (a) 100 J \quad (b) 80 J \quad (c) 40 J \quad (d) 20 J

\medskip
\textbf{Solution:}\\
$\Delta Q = \Delta U + \Delta W \;\Rightarrow\; \Delta U = \Delta Q - \Delta W$\\
Internal energy is a state function: $(\Delta U)_\text{ACB} = (\Delta U)_\text{ADB}$\\
$60 - 30 = \Delta Q_\text{ADB} - 10 \;\Rightarrow\; \Delta Q_\text{ADB} = 40~\mathrm{J}$

\medskip
\begin{tcolorbox}[enhanced, colback=pastelGreenFrame!15, colframe=pastelGreenFrame,
    boxrule=0.4pt, arc=2pt, left=6pt, right=6pt, top=3pt, bottom=3pt]
\small\textbf{Answer: (c) 40 J}
\end{tcolorbox}
\end{tcolorbox}

\medskip

\noindent
\colorbox{pastelGreenFrame}{\small\textbf{\textcolor{white}{Sample 2}}}
\quad
\colorbox{pastelGreen}{\small\textit{\textcolor{pastelGreenFrame}{Thermodynamics \textemdash{} Polytropic process, molar heat capacity}}}

\medskip

\begin{tcolorbox}[
    enhanced, breakable,
    colback=pastelGreen,
    colframe=pastelGreenFrame,
    boxrule=0.6pt, arc=3pt,
    left=10pt, right=10pt, top=6pt, bottom=6pt,
]
\small\normalfont
In a process, temperature and volume of one mole of an ideal monatomic gas vary
according to $VT = K$ (constant). The temperature increases by $\Delta T$.
The heat absorbed by the gas is ($R$ = gas constant):

\medskip
\textbf{Options:} (a) $(2K/3)\Delta T$ \quad (b) $\frac{1}{2}R\Delta T$ \quad
(c) $\frac{3}{2}R\Delta T$ \quad (d) $\frac{1}{2}KR\Delta T$

\medskip
\textbf{Solution:}\\
$VT = K$ and $PV = RT \Rightarrow PV^2 = \text{const.}$ (polytropic, $n=2$)\\
Molar heat capacity: $C = \frac{R}{1-n} + C_v = \frac{R}{1-2} + \frac{3R}{2} = \frac{R}{2}$\\
$\Delta Q = nC\Delta T = \frac{1}{2}R\Delta T$

\medskip
\begin{tcolorbox}[enhanced, colback=pastelGreenFrame!15, colframe=pastelGreenFrame,
    boxrule=0.4pt, arc=2pt, left=6pt, right=6pt, top=3pt, bottom=3pt]
\small\textbf{Answer: (b) $\frac{1}{2}R\Delta T$}
\end{tcolorbox}
\end{tcolorbox}

\medskip

\noindent
\colorbox{pastelGreenFrame}{\small\textbf{\textcolor{white}{Sample 3}}}
\quad
\colorbox{pastelGreen}{\small\textit{\textcolor{pastelGreenFrame}{Thermodynamics \textemdash{} Isochoric heating of helium, molar heat capacity}}}

\medskip

\begin{tcolorbox}[
    enhanced, breakable,
    colback=pastelGreen,
    colframe=pastelGreenFrame,
    boxrule=0.6pt, arc=3pt,
    left=10pt, right=10pt, top=6pt, bottom=6pt,
]
\small\normalfont
A cylinder with fixed capacity $67.2~\mathrm{L}$ contains helium gas at STP.
The heat needed to raise the temperature by $20^\circ\mathrm{C}$ is
[$R = 8.31~\mathrm{J\,mol^{-1}\,K^{-1}}$]:

\medskip
\textbf{Options:} (a) 350 J \quad (b) 700 J \quad (c) 748 J \quad (d) 374 J

\medskip
\textbf{Solution:}\\
$n = 67.2 / 22.4 = 3~\mathrm{mol}$\\
$\Delta Q = nC_v\Delta T = 3 \times \tfrac{3}{2}R \times 20
          = 3 \times 1.5 \times 8.31 \times 20 = 747.9 \approx 748~\mathrm{J}$

\medskip
\begin{tcolorbox}[enhanced, colback=pastelGreenFrame!15, colframe=pastelGreenFrame,
    boxrule=0.4pt, arc=2pt, left=6pt, right=6pt, top=3pt, bottom=3pt]
\small\textbf{Answer: (c) 748 J}
\end{tcolorbox}
\end{tcolorbox}

\bigskip
\noindent\rule{\linewidth}{0.4pt}
\bigskip


\begin{tcolorbox}[
    enhanced,
    colback=pastelBlue,
    colframe=pastelBlueFrame,
    colbacktitle=pastelBlueFrame,
    coltitle=white,
    fonttitle=\normalfont\bfseries\normalsize,
    title={D.2 \quad SciEval-Static Physics \quad
           \normalfont\small $N = 164$ $\mid$
           Conceptual MCQ $\mid$
           Introductory--Intermediate},
    boxrule=0.7pt, arc=4pt,
    left=8pt, right=8pt, top=5pt, bottom=5pt,
]
\normalfont\small
SciEval-Static is the static physics subset of the SciEval benchmark, comprising
164 multiple-choice questions across Fluid Mechanics, Forces, and Waves. Each
item has a single correct answer label; no reference solution is provided.
Ability tags classify questions as Base Knowledge or Scientific Calculation.
\end{tcolorbox}

\medskip

\noindent
\colorbox{pastelBlueFrame}{\small\textbf{\textcolor{white}{Sample 1}}}
\quad
\colorbox{pastelBlue}{\small\textit{\textcolor{pastelBlueFrame}{Fluid Mechanics \textemdash{} Buoyancy (Base Knowledge)}}}

\medskip

\begin{tcolorbox}[enhanced, breakable, colback=pastelBlue, colframe=pastelBlueFrame,
    boxrule=0.6pt, arc=3pt, left=10pt, right=10pt, top=6pt, bottom=6pt]
\small\normalfont
How do buoyant forces occur?

\medskip
A. Buoyant forces occur when an object is exposed to air, causing air pressure to create a lifting force.\\
B. Buoyant forces occur when an object is partially or completely submerged in a liquid, equal to the weight of liquid displaced.\\
C. Buoyant forces occur when an object is heated, causing it to expand and displace liquid.\\
D. Buoyant forces occur when an object is compressed, reducing its volume and displacing liquid.

\medskip
\begin{tcolorbox}[enhanced, colback=pastelBlueFrame!15, colframe=pastelBlueFrame,
    boxrule=0.4pt, arc=2pt, left=6pt, right=6pt, top=3pt, bottom=3pt]
\small\textbf{Answer: B} \quad Topic: Fluid Mechanics \quad Ability: Base Knowledge
\end{tcolorbox}
\end{tcolorbox}

\medskip

\noindent
\colorbox{pastelBlueFrame}{\small\textbf{\textcolor{white}{Sample 2}}}
\quad
\colorbox{pastelBlue}{\small\textit{\textcolor{pastelBlueFrame}{Fluid Mechanics \textemdash{} Bernoulli's Principle (Scientific Calculation)}}}

\medskip

\begin{tcolorbox}[enhanced, breakable, colback=pastelBlue, colframe=pastelBlueFrame,
    boxrule=0.6pt, arc=3pt, left=10pt, right=10pt, top=6pt, bottom=6pt]
\small\normalfont
How does changing the speed of a fluid affect its pressure?

\medskip
A. Fluid pressure is constant regardless of velocity.\\
B. Direct proportionality between fluid pressure and velocity.\\
C. No relationship between fluid pressure and velocity.\\
D. Inverse proportionality between fluid pressure and velocity.

\medskip
\begin{tcolorbox}[enhanced, colback=pastelBlueFrame!15, colframe=pastelBlueFrame,
    boxrule=0.4pt, arc=2pt, left=6pt, right=6pt, top=3pt, bottom=3pt]
\small\textbf{Answer: D} \quad Topic: Fluid Mechanics \quad Ability: Scientific Calculation
\end{tcolorbox}
\end{tcolorbox}

\medskip

\noindent
\colorbox{pastelBlueFrame}{\small\textbf{\textcolor{white}{Sample 3}}}
\quad
\colorbox{pastelBlue}{\small\textit{\textcolor{pastelBlueFrame}{Forces \textemdash{} Vector resultant acceleration (Scientific Calculation)}}}

\medskip

\begin{tcolorbox}[enhanced, breakable, colback=pastelBlue, colframe=pastelBlueFrame,
    boxrule=0.6pt, arc=3pt, left=10pt, right=10pt, top=6pt, bottom=6pt]
\small\normalfont
An object with mass $2~\mathrm{kg}$ is acted on by two forces.
$F_1 = \langle -9~\mathrm{N},\ 8~\mathrm{N} \rangle$ and
$F_2 = \langle -7~\mathrm{N},\ -4~\mathrm{N} \rangle$.
What is the object's rate and direction of acceleration?

\medskip
A. $10.3~\mathrm{m/s^2}$, $124^\circ$ clockwise from $x$-axis\\
B. $7.8~\mathrm{m/s^2}$, $95^\circ$ clockwise from $x$-axis\\
C. $8.25~\mathrm{m/s^2}$, $104^\circ$ clockwise from $x$-axis\\
D. $6.5~\mathrm{m/s^2}$, $86^\circ$ clockwise from $x$-axis

\medskip
\begin{tcolorbox}[enhanced, colback=pastelBlueFrame!15, colframe=pastelBlueFrame,
    boxrule=0.4pt, arc=2pt, left=6pt, right=6pt, top=3pt, bottom=3pt]
\small\textbf{Answer: C} \quad Topic: Forces and Newton's Laws \quad Ability: Scientific Calculation
\end{tcolorbox}
\end{tcolorbox}

\bigskip
\noindent\rule{\linewidth}{0.4pt}
\bigskip


\begin{tcolorbox}[
    enhanced,
    colback=pastelPurple,
    colframe=pastelPurpleFrame,
    colbacktitle=pastelPurpleFrame,
    coltitle=white,
    fonttitle=\normalfont\bfseries\normalsize,
    title={D.3 \quad MMLU High School Physics \quad
           \normalfont\small $N = 170$ $\mid$
           Conceptual MCQ $\mid$ High School},
    boxrule=0.7pt, arc=4pt,
    left=8pt, right=8pt, top=5pt, bottom=5pt,
]
\normalfont\small
The MMLU High School Physics subset comprises 170 four-option multiple-choice
questions testing foundational physics knowledge. No reference solutions are
provided; evaluation is based on final answer accuracy only.
\end{tcolorbox}

\medskip

\noindent
\colorbox{pastelPurpleFrame}{\small\textbf{\textcolor{white}{Sample 1}}}
\quad
\colorbox{pastelPurple}{\small\textit{\textcolor{pastelPurpleFrame}{Electromagnetism \textemdash{} Capacitor charge calculation}}}

\medskip

\begin{tcolorbox}[enhanced, breakable, colback=pastelPurple, colframe=pastelPurpleFrame,
    boxrule=0.6pt, arc=3pt, left=10pt, right=10pt, top=6pt, bottom=6pt]
\small\normalfont
The plates of a capacitor are charged to a potential difference of 5 V.
If the capacitance is 2 mF, what is the charge on the positive plate?

\medskip
A. 0.005 C \quad B. 0.01 C \quad C. 0.02 C \quad D. 0.5 C

\medskip
\begin{tcolorbox}[enhanced, colback=pastelPurpleFrame!15, colframe=pastelPurpleFrame,
    boxrule=0.4pt, arc=2pt, left=6pt, right=6pt, top=3pt, bottom=3pt]
\small\textbf{Answer: B} \quad Topic: Electromagnetism
\end{tcolorbox}
\end{tcolorbox}

\medskip

\noindent
\colorbox{pastelPurpleFrame}{\small\textbf{\textcolor{white}{Sample 2}}}
\quad
\colorbox{pastelPurple}{\small\textit{\textcolor{pastelPurpleFrame}{Electromagnetism \textemdash{} Field decay with distance}}}

\medskip

\begin{tcolorbox}[enhanced, breakable, colback=pastelPurple, colframe=pastelPurpleFrame,
    boxrule=0.6pt, arc=3pt, left=10pt, right=10pt, top=6pt, bottom=6pt]
\small\normalfont
Which of these quantities decreases as the inverse square of distance for
distances far from the objects producing the fields?

\medskip
A. The electric field produced by a finite-length charged rod\\
B. The electric field produced by an infinitely long charged cylinder\\
C. The electric field produced by an infinite plane of charge\\
D. The magnetic field produced by an infinitely long, straight current wire

\medskip
\begin{tcolorbox}[enhanced, colback=pastelPurpleFrame!15, colframe=pastelPurpleFrame,
    boxrule=0.4pt, arc=2pt, left=6pt, right=6pt, top=3pt, bottom=3pt]
\small\textbf{Answer: A} \quad Topic: Electromagnetism
\end{tcolorbox}
\end{tcolorbox}

\medskip

\noindent
\colorbox{pastelPurpleFrame}{\small\textbf{\textcolor{white}{Sample 3}}}
\quad
\colorbox{pastelPurple}{\small\textit{\textcolor{pastelPurpleFrame}{Electromagnetism \textemdash{} Field near a non-uniformly charged conductor}}}

\medskip

\begin{tcolorbox}[enhanced, breakable, colback=pastelPurple, colframe=pastelPurpleFrame,
    boxrule=0.6pt, arc=3pt, left=10pt, right=10pt, top=6pt, bottom=6pt]
\small\normalfont
A solid metal object is isolated from other charges and has non-uniform charge
distribution on its surface. It may be correctly concluded that the:

\medskip
A. electric field outside the object is zero\\
B. electric field outside equals the field inside the object\\
C. external field is directly proportional to distance from the centre of mass\\
D. external field very close to the surface equals the surface charge density
   at any location divided by the permittivity of free space

\medskip
\begin{tcolorbox}[enhanced, colback=pastelPurpleFrame!15, colframe=pastelPurpleFrame,
    boxrule=0.4pt, arc=2pt, left=6pt, right=6pt, top=3pt, bottom=3pt]
\small\textbf{Answer: D} \quad Topic: Electromagnetism
\end{tcolorbox}
\end{tcolorbox}

\bigskip
\noindent\rule{\linewidth}{0.4pt}
\bigskip


\begin{tcolorbox}[
    enhanced,
    colback=pastelOrange,
    colframe=pastelOrangeFrame,
    colbacktitle=pastelOrangeFrame,
    coltitle=white,
    fonttitle=\normalfont\bfseries\normalsize,
    title={D.4 \quad MMLU College Physics \quad
           \normalfont\small $N = 118$ $\mid$
           Conceptual MCQ $\mid$ Undergraduate},
    boxrule=0.7pt, arc=4pt,
    left=8pt, right=8pt, top=5pt, bottom=5pt,
]
\normalfont\small
The MMLU College Physics subset comprises 118 questions at undergraduate
difficulty, covering Modern Physics, Optics, Thermodynamics, and Mechanics.
Problems frequently require synthesis of multiple concepts and dimensional
reasoning.
\end{tcolorbox}

\medskip

\noindent
\colorbox{pastelOrangeFrame}{\small\textbf{\textcolor{white}{Sample 1}}}
\quad
\colorbox{pastelOrange}{\small\textit{\textcolor{pastelOrangeFrame}{Modern Physics \textemdash{} Quantum efficiency and Poisson statistics}}}

\medskip

\begin{tcolorbox}[enhanced, breakable, colback=pastelOrange, colframe=pastelOrangeFrame,
    boxrule=0.6pt, arc=3pt, left=10pt, right=10pt, top=6pt, bottom=6pt]
\small\normalfont
The quantum efficiency of a photon detector is 0.1. If 100 photons are sent
into the detector one after the other, the detector will detect photon:

\medskip
A. an average of 10 times, with an rms deviation of about 4\\
B. an average of 10 times, with an rms deviation of about 3\\
C. an average of 10 times, with an rms deviation of about 1\\
D. an average of 10 times, with an rms deviation of about 0.1

\medskip
\begin{tcolorbox}[enhanced, colback=pastelOrangeFrame!15, colframe=pastelOrangeFrame,
    boxrule=0.4pt, arc=2pt, left=6pt, right=6pt, top=3pt, bottom=3pt]
\small\textbf{Answer: B} \quad Topic: Modern Physics \quad
Note: Binomial, $n=100$, $p=0.1$; $\sigma = \sqrt{np(1-p)} = \sqrt{9} = 3$
\end{tcolorbox}
\end{tcolorbox}

\medskip

\noindent
\colorbox{pastelOrangeFrame}{\small\textbf{\textcolor{white}{Sample 2}}}
\quad
\colorbox{pastelOrange}{\small\textit{\textcolor{pastelOrangeFrame}{Optics \textemdash{} Thin film interference, reflected wavelength}}}

\medskip

\begin{tcolorbox}[enhanced, breakable, colback=pastelOrange, colframe=pastelOrangeFrame,
    boxrule=0.6pt, arc=3pt, left=10pt, right=10pt, top=6pt, bottom=6pt]
\small\normalfont
White light is normally incident on a puddle of water (refractive index 1.33).
A thin 500 nm layer of oil (refractive index 1.5) floats on the surface.
Of the following, the most strongly reflected wavelength is:

\medskip
A. 500 nm \quad B. 550 nm \quad C. 600 nm \quad D. 650 nm

\medskip
\begin{tcolorbox}[enhanced, colback=pastelOrangeFrame!15, colframe=pastelOrangeFrame,
    boxrule=0.4pt, arc=2pt, left=6pt, right=6pt, top=3pt, bottom=3pt]
\small\textbf{Answer: C} \quad Topic: Optics
\end{tcolorbox}
\end{tcolorbox}

\medskip

\noindent
\colorbox{pastelOrangeFrame}{\small\textbf{\textcolor{white}{Sample 3}}}
\quad
\colorbox{pastelOrange}{\small\textit{\textcolor{pastelOrangeFrame}{Thermodynamics \textemdash{} Reversible process and entropy}}}

\medskip

\begin{tcolorbox}[enhanced, breakable, colback=pastelOrange, colframe=pastelOrangeFrame,
    boxrule=0.6pt, arc=3pt, left=10pt, right=10pt, top=6pt, bottom=6pt]
\small\normalfont
Which of the following is true about any system that undergoes a reversible
thermodynamic process?

\medskip
A. There are no changes in the internal energy of the system.\\
B. The temperature of the system remains constant during the process.\\
C. The entropy of the system and its environment remains unchanged.\\
D. The entropy of the system and its environment must increase.

\medskip
\begin{tcolorbox}[enhanced, colback=pastelOrangeFrame!15, colframe=pastelOrangeFrame,
    boxrule=0.4pt, arc=2pt, left=6pt, right=6pt, top=3pt, bottom=3pt]
\small\textbf{Answer: C} \quad Topic: Thermodynamics
\end{tcolorbox}
\end{tcolorbox}

\bigskip
\noindent\rule{\linewidth}{0.4pt}
\bigskip


\begin{tcolorbox}[
    enhanced,
    colback=pastelRed,
    colframe=pastelRedFrame,
    colbacktitle=pastelRedFrame,
    coltitle=white,
    fonttitle=\normalfont\bfseries\normalsize,
    title={D.5 \quad JEEBench \quad
           \normalfont\small $N = 123$ $\mid$
           JEE Advanced MCQ $\mid$ Highest difficulty},
    boxrule=0.7pt, arc=4pt,
    left=8pt, right=8pt, top=5pt, bottom=5pt,
]
\normalfont\small
JEEBench comprises 123 physics problems drawn from JEE Advanced examination
papers (2016--2023). Problems require multi-concept synthesis, mathematical
fluency, and derivation under examination conditions. Ground truth is the
correct answer option only; no reference solutions are provided.
\end{tcolorbox}

\medskip

\noindent
\colorbox{pastelRedFrame}{\small\textbf{\textcolor{white}{Sample 1 (JEE Adv 2016, Paper 1)}}}
\quad
\colorbox{pastelRed}{\small\textit{\textcolor{pastelRedFrame}{Modern Physics \textemdash{} Planck's constant from photoelectric stopping potential data}}}

\medskip

\begin{tcolorbox}[enhanced, breakable, colback=pastelRed, colframe=pastelRedFrame,
    boxrule=0.6pt, arc=3pt, left=10pt, right=10pt, top=6pt, bottom=6pt]
\small\normalfont
In a historical experiment to determine Planck's constant, a metal surface was
irradiated with light of different wavelengths. The stopping potentials measured
are:

\medskip
\begin{center}
\small
\begin{tabular}{cc}
\toprule
$\lambda~(\mu\mathrm{m})$ & $V_0~(\mathrm{V})$ \\
\midrule
0.3 & 2.0 \\
0.4 & 1.0 \\
0.5 & 0.4 \\
\bottomrule
\end{tabular}
\end{center}

\medskip
Given $c = 3\times10^8~\mathrm{m\,s^{-1}}$, $e = 1.6\times10^{-19}~\mathrm{C}$.
Planck's constant found from this experiment is:

\medskip
(A) $6.0\times10^{-34}$ J\,s \quad
(B) $6.4\times10^{-34}$ J\,s \quad
(C) $6.6\times10^{-34}$ J\,s \quad
(D) $6.8\times10^{-34}$ J\,s

\medskip
\begin{tcolorbox}[enhanced, colback=pastelRedFrame!15, colframe=pastelRedFrame,
    boxrule=0.4pt, arc=2pt, left=6pt, right=6pt, top=3pt, bottom=3pt]
\small\textbf{Gold Answer: B}
\end{tcolorbox}
\end{tcolorbox}

\medskip

\noindent
\colorbox{pastelRedFrame}{\small\textbf{\textcolor{white}{Sample 2 (JEE Adv 2016, Paper 1)}}}
\quad
\colorbox{pastelRed}{\small\textit{\textcolor{pastelRedFrame}{Mechanics \textemdash{} Stick on inclined wall, torque and friction equilibrium}}}

\medskip

\begin{tcolorbox}[enhanced, breakable, colback=pastelRed, colframe=pastelRedFrame,
    boxrule=0.6pt, arc=3pt, left=10pt, right=10pt, top=6pt, bottom=6pt]
\small\normalfont
A uniform wooden stick of mass $1.6~\mathrm{kg}$ and length $l$ rests inclined
on a smooth vertical wall of height $h < l$ such that a small portion extends
beyond. The wall reaction is perpendicular to the stick; the stick makes
$30^\circ$ with the wall; the bottom rests on a rough floor. The wall reaction
equals the floor reaction in magnitude. ($g = 10~\mathrm{m\,s^{-2}}$)

\medskip
(A) $h/l = \sqrt{3}/16$, \quad $f = 16\sqrt{3}/3~\mathrm{N}$\\
(B) $h/l = 3/16$, \quad $f = 16\sqrt{3}/3~\mathrm{N}$\\
(C) $h/l = 3\sqrt{3}/16$, \quad $f = 8\sqrt{3}/3~\mathrm{N}$\\
(D) $h/l = 3\sqrt{3}/16$, \quad $f = 16\sqrt{3}/3~\mathrm{N}$

\medskip
\begin{tcolorbox}[enhanced, colback=pastelRedFrame!15, colframe=pastelRedFrame,
    boxrule=0.4pt, arc=2pt, left=6pt, right=6pt, top=3pt, bottom=3pt]
\small\textbf{Gold Answer: D}
\end{tcolorbox}
\end{tcolorbox}

\medskip

\noindent
\colorbox{pastelRedFrame}{%
  \small\textbf{\textcolor{white}{Sample 3 (JEE Adv 2016, Paper 2)}}%
}

\vspace{2pt}

\noindent
\colorbox{pastelRed}{%
  \parbox{\dimexpr\linewidth-2\fboxsep\relax}{%
    \small\raggedright\itshape
    \textcolor{pastelRedFrame}{%
      Nuclear Physics \textemdash{} Electrostatic energy and nuclear
      radius from binding energy difference%
    }%
  }%
}
\medskip

\begin{tcolorbox}[enhanced, breakable, colback=pastelRed, colframe=pastelRedFrame,
    boxrule=0.6pt, arc=3pt, left=10pt, right=10pt, top=6pt, bottom=6pt]
\small\normalfont
The electrostatic energy of $Z$ protons uniformly distributed in a spherical
nucleus of radius $R$ is
$E = \frac{3}{5}\frac{Z(Z-1)e^2}{4\pi\varepsilon_0 R}$.
Measured masses: neutron $= 1.008665~\mathrm{u}$,
$^1_1\mathrm{H} = 1.007825~\mathrm{u}$,
$^{15}_7\mathrm{N} = 15.000109~\mathrm{u}$,
$^{15}_8\mathrm{O} = 15.003065~\mathrm{u}$.
Radii of $^{15}\mathrm{N}$ and $^{15}\mathrm{O}$ are equal.
[$1~\mathrm{u} = 931.5~\mathrm{MeV}/c^2$,
$e^2/(4\pi\varepsilon_0) = 1.44~\mathrm{MeV\,fm}$,
$1~\mathrm{fm} = 10^{-15}~\mathrm{m}$]

\medskip
Assuming the binding energy difference between $^{15}$N and $^{15}$O is purely
electrostatic, the radius of either nucleus is:

\medskip
(A) 2.85 fm \quad (B) 3.03 fm \quad (C) 3.42 fm \quad (D) 3.80 fm

\medskip
\begin{tcolorbox}[enhanced, colback=pastelRedFrame!15, colframe=pastelRedFrame,
    boxrule=0.4pt, arc=2pt, left=6pt, right=6pt, top=3pt, bottom=3pt]
\small\textbf{Gold Answer: C}
\end{tcolorbox}
\end{tcolorbox}

\bigskip

\begin{tcolorbox}[
    enhanced,
    colback=pastelBlueDeep,
    colframe=pastelBlueFrame,
    boxrule=0.5pt, arc=3pt,
    left=8pt, right=8pt, top=5pt, bottom=5pt,
]
\small\normalfont
All benchmark samples are used as-is from their respective datasets with no
modification. Models are prompted to produce a structured seven-tag solution
before selecting a final answer, enabling step-level error analysis alongside
final answer accuracy evaluation.
\end{tcolorbox}

\twocolumn


\end{document}